\documentclass{article}
\usepackage[preprint]{neurips_2026}
\usepackage{booktabs}
\usepackage{amsmath}
\usepackage[T1]{fontenc}
\usepackage{makecell}
\usepackage{tablefootnote}
\usepackage{longtable}
\usepackage{pdflscape}
\usepackage{lastpage}
\usepackage{hyperref}
\usepackage{amssymb}
\usepackage{graphicx}
\usepackage[section]{placeins}
\newcommand{\monster}{\textsc{Monster}}

\begin{document}

\title{The Multiverse of Time Series Machine Learning: an Archive for Multivariate Time Series Classification}
\author{
  Matthew Middlehurst \\
  University of Bradford \\
  \texttt{m.b.middlehurst@bradford.ac.uk} \\
  \And
  Aiden Rushbrooke \\
  University of East Anglia \\
  \texttt{aiden.rushbrooke@uea.ac.uk} \\
  \And
  Ali Ismail-Fawaz, Maxime Devanne, Germain Forestier \\
  Universit\'{e} de Haute-Alsace \\
  \texttt{\{ali-el-hadi.ismail-fawaz, maxime.devanne,
  germain.forestier\}@uha.fr} \\
  \And
  Angus Dempster, Geoffrey I.\ Webb \\
  Monash University \\
  \texttt{\{angus.dempster, geoff.webb\}@monash.edu} \\
  \And
  Christopher Holder, Anthony Bagnall \\
  University of Southampton \\
  \texttt{\{c.l.holder, a.j.bagnall\}@soton.ac.uk}
}
\maketitle

\begin{abstract}
  Time series machine learning (TSML) is a growing research field that spans a wide range
  of tasks. The popularity of established tasks such as classification, clustering, and
  extrinsic regression has, in part, been driven by the availability of benchmark datasets.
  An archive of 30 multivariate time series classification datasets, introduced in 2018 and
  commonly known as the UEA archive, has since become an essential resource cited in hundreds
  of publications. We present a substantial expansion of this archive that more than quadruples
  its size, from 30 to 133 classification problems. We also release preprocessed versions of
  datasets containing missing values or unequal length series, bringing the total number of
  datasets to 147. Reflecting the growth of the archive and the broader community, we rebrand
  it as the Multiverse archive to capture its diversity of domains. The Multiverse archive
  includes datasets from multiple sources, consolidating other collections and standalone
  datasets into a single, unified repository. Recognising that running experiments across the
  full archive is computationally demanding, we recommend a subset of the full archive called
  Multiverse-core (MV-core) for initial exploration. To support researchers in using the new
  archive, we provide detailed guidance and a baseline evaluation of established and recent
  classification algorithms, establishing performance benchmarks for future research. We have
  created a dedicated repository for the Multiverse archive that provides a common \textit{aeon}
  and \textit{scikit-learn} compatible framework for reproducibility, an extensive record of
  published results, and an interactive interface to explore the results.
\end{abstract}

\section{Introduction}
\label{sec:introduction}

Time series data underpin a wide range of modern applications, from wearable and medical sensors, through industrial monitoring and robotics, to satellite imaging, finance, and web analytics. The rapid growth in the collection of such data has led to a corresponding increase in interest in time series machine learning (TSML), a field concerned with learning from ordered, possibly multivariate, sequences of observations. TSML spans a range of tasks, including forecasting, anomaly detection, and segmentation. Perhaps the most widely studied of these is time series classification (TSC), where collections of time series are used to learn to predict associated class labels. The popularity of TSC research has been driven by expanding application areas, the availability of open-source implementations of state-of-the-art algorithms, and benchmark datasets that support reproducible evaluation.

The University of California, Riverside (UCR) archive~\cite{dau18archive}, first introduced early this century, has played a central role in advancing univariate TSC. Many real-world problems, however, are multivariate, involving measurements taken simultaneously from multiple channels. To address this gap, the University of East Anglia (UEA) Multivariate TSC (MTSC) archive was first released in 2018~\cite{bagnall18mtsc} with 30 datasets.
In the conclusions of the original release~\cite{bagnall18mtsc}, the authors state: {\em ``This is our first attempt at a multivariate archive, and it should be considered a work in progress. We hope to release an expanded version in 2019''}. While the arXiv paper was subsequently extended to include a bake off of the state-of-the-art at the time~\cite{ruiz21mtsc}, the archive itself has not been expanded. The deadline may have slipped, but we now fulfil the hope with an extended MTSC archive.

This expansion increases the total number of released datasets from 30 to 147. Some of these are preprocessed variations of the same problem. There are 133 unique datasets. This is more than four times the size of the original archive and comparable in scale to the UCR 2019 release. Some of the new datasets come from other recently released archives, some are standalone datasets collected since the original UEA release, and others have been donated through the TSC website\footnote{\url{https://timeseriesclassification.com/}}. This work reflects contributions from researchers across multiple institutions, as well as datasets obtained through industrial partners and public organisations. Given this broader community ownership and the changing affiliations of the archive maintainers, we relaunch the collection under the name \textit{Multiverse} rather than tie it to a single institution. We have created an associated repository containing examples on downloading and using the archive, and a record of published and recreated results\footnote{\url{https://github.com/aeon-toolkit/multiverse}}.

Although the archive is much larger, it still presents practical challenges for routine algorithm development. Some datasets are very large in terms of the number of cases, the number of time points, or the number of channels. Several contain millions of cases, tens of thousands of time points, or more than a thousand channels. While such datasets are essential for assessing scalability, they are difficult to use in broad resample-based benchmark studies involving many algorithms. This can encourage researchers to evaluate on bespoke subsets of the archive, especially when computational resources are limited. The archive is also biased towards certain application domains, particularly EEG and human activity recognition, reflecting both contributor interests and the availability of public multivariate data. To address these practical issues, we propose a curated subset of the full archive, Multiverse-core (MV-core), intended to support initial algorithm development and routine benchmarking.

To support use of the new archive, we provide an initial benchmark using the best-performing classifiers from a 2021 MTSC comparison~\cite{ruiz21mtsc} together with the multivariate-capable methods from a recent univariate classification evaluation~\cite{middlehurst24bakeoff}. This benchmark is restricted to the default train-test splits supplied with the archive and should be viewed as a starting point rather than a comprehensive bake off. We also briefly review recent MTSC research, not to provide a full survey, but to place the archive in context and to motivate the need for a broader and more diverse benchmark collection. More broadly, our aim is to provide a framework for future bake offs and clearer guidance on how new algorithms should be assessed using the archive.

The remainder of this paper is structured as follows. Section 2 reviews existing TSML data archives. Section 3 summarises related research on MTSC. Section 4 describes the expanded Multiverse archive and its constituent datasets. Section 5 introduces and justifies Multiverse-core. Section 6 presents benchmark results and analysis. Section 7 concludes with a discussion of current limitations and future directions.

\section{Existing TSML Archives}
\label{sec:archives}

Curated benchmark archives are a cornerstone of reproducible research in machine learning. They provide the raw material for developing and evaluating new methods. Over time, they support the measurement of progress and enable subsequent users to have greater confidence that reported performance will translate to their own applications. Time series archives are available for a range of problem types.

\sloppy A univariate time series dataset for supervised machine learning is defined as $\mathcal{D}=\{(\mathbf{x}_i, y_i)\}_{i=1}^{n}$, where each of the $n$ cases consists of a single ordered sequence $\mathbf{x}_i = (x_{i,1}, x_{i,2}, \dots, x_{i,m})$ observed over $m$ time points and an associated target $y_i$. The ordering of the observations is fundamental, since unlike a standard tabular feature vector, neighbouring values in a time series are temporally dependent and informative patterns may arise in a variety of forms. For multivariate time series, each case consists of $d$ ordered dimensions, $\mathbf{X}_i = [\mathbf{x}_i^{(1)}, \ldots, \mathbf{x}_i^{(d)}]$, where each $\mathbf{x}_i^{(j)}$ is a univariate series. When all series are of equal length, the dataset may be represented as an array in $\mathbb{R}^{n \times d \times m}$, with univariate data as the special case $\mathbb{R}^{n \times 1 \times m}$.

The UCR archive has played a pivotal role in the development of univariate time series classification (UTSC) algorithms. It was motivated by a review of poor experimental practices in the field~\cite{keogh03benchmarks} and was first released in 2002 by Eamonn Keogh's group as a collection of 16 UTSC datasets. It then expanded incrementally to 45 datasets through ad hoc contributions, before further batch expansions to 85 datasets in 2015 and to 128 in 2018~\cite{dau18archive}. This archive has been cited in thousands of publications and has underpinned numerous comparative studies in both classification and clustering. It has helped clarify the relative strengths of different algorithms and improved experimental evaluation.

Research on multivariate time series classification (MTSC) in the 2010s suffered issues similar to UTSC work in the 2000s: studies were often conducted on small numbers of datasets selected without clear motivation, and reproducibility was difficult. The UEA archive~\cite{bagnall18mtsc} of 30 MTSC datasets was released in 2018 to provide a common benchmark and support progress in the field. A bake off of the state-of-the-art at the time was subsequently published~\cite{ruiz21mtsc}. The UEA archive has been cited in hundreds of publications, but it has not fully served its intended purpose as a centralised resource for reproducible MTSC research. Many studies using the archive evaluate only a narrow subset of benchmarks or select bespoke subsets, and the smaller number of datasets relative to the UCR archive makes it harder to draw robust conclusions about overall comparative performance. We review the usage of the UEA archive in more detail in Section~\ref{sec:uea}.

A range of specialised UTSC and MTSC archives have subsequently been released:
\begin{enumerate}
\item An archive of 19 time series extrinsic regression (TSER) datasets was introduced in~\cite{tan21regression}, 15 of which are multivariate. This was recently expanded to a collection of 63 datasets containing both univariate and multivariate problems~\cite{guijo2024unsupervised}.
\item An archive of 30 ordinal univariate and multivariate datasets was used in~\cite{guijo20ordinal}. This collection was partly derived from UCR/UEA, partly consisting of new problems.
\item Thirty new UTSC datasets were used in~\cite{middlehurst24bakeoff} to decrease reliance on the UCR archive.
\item An imbalanced-class variant of the UCR archive was introduced in 2024~\cite{qiu25esmote}. This contains 76 of the UCR problems resampled to simulate two class problems with a minority class of at most 10\% of the cases.
\item The Electroencephalography (EEG) time series archive contains 30 EEG classification problems from medical, psychology and BCI research~\cite{rushbrooke2026eeg}.
\item The MONSTER (Monash Scalable Time Series Evaluation) repository~\cite{dempster2025monster} provides a benchmark suite of large-scale univariate and multivariate time series classification datasets, intended to complement UCR/UEA and to encourage evaluation of scalability and computational trade-offs on substantially larger problems.
\item The Rehab-Pile archive~\cite{fawaz2025rehab} aggregates existing skeleton-based human motion rehabilitation datasets into a unified benchmark, comprising 39 classification datasets and 21 extrinsic regression datasets, with a standard evaluation protocol and publicly released datasets and implementations to support reproducible comparison.
\end{enumerate}

There are many other time series archives for tasks such as forecasting, anomaly detection, and segmentation that are beyond our scope. Collectively, the archives listed above form part of a broader TSML benchmark ecosystem spanning classic machine learning tasks such as classification,  clustering, and extrinsic regression (similar to traditional machine learning regression, but named as such to avoid confusion with forecasting regression~\cite{tan21regression}). Some adapt the UCR or UEA archives, some use datasets donated through the TSC website that were not included in earlier UCR or UEA releases, some are derived from open-source research data, and some have been assembled directly by archive maintainers. All address the general problem of learning from collections of time series. Our aim is to make the relationships between these datasets more explicit and to provide a more centralised resource for reproducible MTSC research. To this end, we relaunch the MTSC collection under the name \emph{Multiverse}.

\section{MTSC Related Research}
\label{sec:tsc}

Research into new algorithms for MTSC has grown rapidly in recent years. There is a diverse literature, spanning feature-based machine learning algorithms, new deep learning architectures, adaptations of established time series classifiers, and a wide range of application-specific studies.

We do not aim to provide a full survey of recent multivariate time series classification (MTSC) research here, nor do we intend the experiments presented in Section~\ref{sec:results} to constitute a new comprehensive bake off. Instead, we briefly highlight a few representative trends to motivate the need for a broader and more diverse archive. Following the earlier UEA archive and large-scale comparative studies~\cite{bagnall18mtsc,ruiz21mtsc}, recent MTSC research has expanded rapidly, particularly in deep learning.

\subsection{Machine Learning for MTSC}

Before the recent popularity of deep learning approaches, MTSC research was driven largely by methods adapted from univariate time series classification and by new algorithms designed to exploit multivariate structure more explicitly. Following the taxonomy of~\cite{middlehurst24bakeoff}, we organise algorithms by representation type, including distance-based, convolution-based, shapelet-based, feature-based, interval-based, dictionary-based, and hybrid ensemble approaches.

A traditional reference point for both univariate and multivariate time series classification is Dynamic Time Warping (DTW) used with a one-nearest-neighbour classifier. DTW is an elastic distance measure that aligns two series by allowing local temporal warping. In the multivariate setting, variants differ in how the dimensions are coupled. Dimension-dependent DTW (DTW\_D) enforces a single warping path shared across all dimensions, whereas dimension-independent DTW (DTW\_I) aligns each dimension separately, and adaptive DTW (DTW\_A) interpolates between these extremes~\cite{shokoohi2017generalizing}. These methods remain useful as simple and interpretable benchmarks for MTSC.

Another line of work uses random convolutional kernels in a transformation pipeline. ROCKET~\cite{dempster20rocket} generates a large number of randomly parameterised one-dimensional convolutional kernels and summarises each transformed series using pooled statistics of maximum value and the proportion of positive values. These features are then classified with a linear model, usually a ridge classifier. Multivariate extensions adapt this idea by allowing kernels to operate across selected dimensions. ROCKET led to several extensions, including Arsenal~\cite{middlehurst21hc2}, which ensembles multiple ROCKET classifiers, and MultiROCKET-Hydra, which combines the ideas of MultiROCKET~\cite{tan22multirocket} and Hydra~\cite{dempster23hydra}.

Shapelet-based methods take a different approach by representing a series through its similarity to informative subsequences. The Shapelet Transform Classifier (STC)~\cite{bostrom17binary} searches for discriminative subseries and transforms each series into a feature vector of distances from these shapelets, before applying a conventional classifier such as rotation forest~\cite{rodriguez06rotf}. More recent work has reduced the cost of shapelet discovery by replacing exhaustive search with random sampling. The Random Dilated Shapelet Transform (RDST)~\cite{guillaume22rdst}, for example, samples many shapelets from the training set and introduces dilation, allowing the sampled points within a shapelet to be spaced apart so that matching can occur at multiple temporal resolutions. In the multivariate setting, shapelets are typically extracted from individual channels rather than jointly across all dimensions.

A further strand of work represents each series through engineered summary features. Catch22~\cite{lubba19catch22} provides a compact set of 22 features distilled from the much larger hctsa library~\cite{fulcher17hctsa}, with the aim of capturing a broad range of time series properties in a computationally efficient form. Other approaches use much larger feature libraries. FreshPRINCE~\cite{middlehurst22freshprince}, for example, builds on TSFresh~\cite{christ2018time} and combines extracted features with an ensemble classifier. For multivariate data, such feature-based approaches are usually extended by applying the feature set to each channel and concatenating the resulting representations.

Interval-based classifiers form another family of MTSC methods. Rather than analysing the whole series directly, they sample subintervals and compute summary statistics or derived features on each one. The Canonical Interval Forest (CIF)~\cite{middlehurst20canonical} builds an ensemble of trees using attributes extracted from randomly selected intervals, including basic summary statistics and Catch22-based features. Diverse Representation CIF (DrCIF)~\cite{middlehurst21hc2} extends this idea by extracting interval features not only from the raw series but also from transformed representations such as periodograms and first-order differences. QUANT~\cite{dempster23quant} is a more recent interval-based method that summarises fixed dyadic intervals using quantiles arranged in a pyramid structure, again across multiple series representations. These approaches have been among the strongest traditional classifiers on large benchmark studies.

Dictionary-based methods convert time series into symbolic words and represent each case as a histogram over these words, analogous to bag-of-words models in text and computer vision. A  multivariate example is WEASEL+MUSE~\cite{Schafer2017WEASELMUSE}, which extends symbolic Fourier methods to multivariate data. The Temporal Dictionary Ensemble (TDE)~\cite{middlehurst20temporal} is a later dictionary-based approach that ensembles nearest-neighbour classifiers built on Symbolic Fourier Approximation (SFA) representations~\cite{schaefer12sfa}, with parameter selection optimised over a search space that also includes additional features such as spatial pyramid words. Dictionary methods are particularly notable because they transform time series into an explicitly symbolic representation while remaining highly competitive.

Finally, hybrid approaches attempt to combine complementary representations within a single classifier. HIVE-COTE 2.0~\cite{middlehurst21hc2} is the most prominent example, combining interval, dictionary, convolutional-kernel, and shapelet-based components, specifically DrCIF, TDE, Arsenal, and STC, within a probabilistic weighted ensemble. Another example is the Randomised Interval-Shapelet Transformation (RIST)~\cite{middlehurst23rist}, which deliberately combines interval and shapelet representations in a simpler pipeline. These hybrid methods reflect a broader lesson from MTSC research: no single representation is uniformly best, and strong performance often comes from combining approaches that capture different aspects of the data.

\subsection{Deep Learning for MTSC}

A large part of this recent work has focused on neural architectures that model temporal dependencies across channels more effectively. Early deep approaches included recurrent and convolutional models such as TapNet~\cite{Zhang2020TapNet}, XCM~\cite{Fauvel2021XCM}, and Disjoint-CNN~\cite{Foumani2021DisjointCNN}. More recent work has increasingly explored transformer-based and backbone-style architectures, including TST~\cite{Zerveas2021TST}, FormerTime~\cite{Cheng2023FormerTime}, ConvTran~\cite{Foumani2024ConvTran}, and ShapeFormer~\cite{Le2024ShapeFormer}. General time series backbones such as TimesNet~\cite{wu2023timesnet}, and more recently ModernTCN~\cite{Luo2024ModernTCN}, have also influenced MTSC research, even when proposed for broader time series analysis settings.

Alongside this, there has been growing interest in multiscale, time-frequency, and patch-based representations \cite{Lei2024TFNet,Wei2025PatchMTSC,Yuan2025Contrastive}, graph-based approaches for modelling inter-variable structure \cite{Younis2024MTS2Graph,Hong2024MagNet}, and self-supervised or contrastive representation learning \cite{Yue2022TS2Vec,Foumani2024Series2Vec,Liu2024TimesURL}. There is also a continuing thread of work on interpretability and efficiency, for example through tokenised shape abstractions, simpler interpretable models, or lightweight architectures \cite{Lee2024ZTime,Wen2024VQShape,IsmailFawaz2025LITETime}. Most recently, foundation-model and LLM-inspired approaches have begun to appear, although these remain comparatively early and their role in MTSC is not yet clear~\cite{Zhou2023OneFitsAll,Kamarthi2024LPTM,Chen2025LLMFew,Ilbert2025Foundation}.  Overall, the recent literature shows a wide range of architectural ideas. All of these papers use some or all of the UEA archive in their evaluation. However, there is significant variation in methodology and varying degrees of reproducibility.

\subsection{The use of the UEA MTSC Archive in Research}
\label{sec:uea}

The UEA archive has seen significant uptake in the last four years, particularly in deep learning research. We have screened papers citing the two archive papers to gain insight into research trends. Table~\ref{tab:uea} summarises our findings. The first two columns show the number of citations per year for each paper, based on a Google Scholar snapshot. Some papers cite both archive papers, while others cite only one, but the arXiv paper continues to receive more citations.

    \begin{table}[]
        \caption{Counts of papers roughly classified into one of the categories ML (machine learning), Conv (primarily CNN based for classification), Graph (learn structure on a graph representation), Transf (transformer based architecture using, e.g. attention or masking), foundation (model proposed as a backbone for multiple tasks which may be pretrained).} \label{tab:uea}
        \centering \small
        \begin{tabular}{lrrr|rrrrrrr}
        \toprule
        Year & arXiv & Journal & Total & ML & Conv & RNN & Graph & Transf & Found & Deep  \\
        \midrule
        2019 & 12 & 0 & 3 & 0 & 2 & 0 & 0 & 1 & 0 & 0 \\
        2020 & 33 & 0 & 8 & 4 & 1 & 0 & 0 & 3 & 0 & 0 \\
        2021 & 72 & 28 & 12 & 6 & 3 & 2 & 0 & 0 & 1 & 0 \\
        2022 & 99 & 100 & 18 & 4 & 2 & 1 & 2 & 7 & 2 & 0 \\
        2023 & 138 & 173 & 15 & 3 & 3 & 0 & 1 & 6 & 1 & 1 \\
        2024 & 197 & 174 & 39 & 6 & 4 & 1 & 5 & 12 & 7 & 4 \\
        2025 & 253 & 139 & 54 & 7 & 7 & 1 & 5 & 16 & 12 & 6 \\
        2026 & -- & -- & 6 & 0 & 2 & 0 & 0 & 3 & 1 & 0 \\
        \bottomrule
        \end{tabular}
    \end{table}

The archive is referenced both by papers proposing new algorithms and by work applying MTSC to a wide range of tasks such as anomaly detection, adversarial augmentation, class imbalance, handling missing values or imputation, noisy labelling, early classification, semi-supervised learning. It is used in the development of diverse applications including oil pump leakage detection, health data, manufacturing, medical signal analysis, ship motion analysis, classifying high-altitude flying objects, chemical sensors, solar flare prediction, rumour classification, and astronomy. We assessed papers citing the archive to identify those proposing new algorithms for the core task of MTSC. The resulting counts are shown in the `Total' column of Table~\ref{tab:uea}. These statistics come with caveats. We have probably missed some papers, we only included peer-reviewed publications (excluding arXiv-only preprints) and some papers use the archive without citation. Nevertheless, Table~\ref{tab:uea} demonstrates rapid growth in algorithm proposals for MTSC. In 2025 alone, 54 distinct approaches were proposed, and well over 100 algorithms have used evaluation on the UEA archive to directly support their contribution.

To characterise this trend, we categorised algorithms by type, using a simple priority order to handle hybrids. Each paper is assigned a single label, defined as follows:
\begin{enumerate}
    \item \textbf{Foundation (Found)} if it explicitly involves pretraining or masked modelling, uses an LLM or in-context learning, or is presented as a backbone for multiple tasks or general time-series analysis.
    \item \textbf{Transformer (Transf)} if it uses attention or patching within a transformer-style architecture.
    \item \textbf{Graph} if the core encoder uses an explicit graph construction and GNN-style propagation, for example GCN, GAT, hypergraph methods, or visibility graphs.
    \item \textbf{Convolutional (Conv)} if it is primarily CNN or TCN based, including FCN and ResNet-style variants.
    \item \textbf{Recurrent neural network (RNN)} for LSTM, GRU, and other recurrent architectures.
    \item \textbf{Deep other (Deep)} for deep generative or self-supervised encoders not captured above, for example VAE, GAN, diffusion, autoencoder, or contrastive methods.
    \item \textbf{Machine learning (ML)} for classical, non-deep feature or distance-based approaches, for example shapelet, dictionary, or feature-engineering baselines.
\end{enumerate}
This categorisation is approximate and intended to identify broad trends, since some papers span multiple modelling choices. Table~\ref{tab:uea} shows the breakdown by type and year. There is a steady stream of ML and Conv papers, but the growth in the last two years is driven primarily by Transformer and Foundation papers. There is also an increase in Graph methods. There is substantial variation in how papers use the UEA archive, both in which datasets are selected and in how assessment is performed. Many papers do not use all datasets, some truncate or pad unequal length problems, and some merge the original train and test splits and then resample. Furthermore, as highlighted in a recent review~\cite{Foumani2024DeepSurvey}, many do not release source code. This makes comparison between papers difficult. Table~\ref{tab:data} summarises code availability, dataset coverage, and the number of comparison algorithms. Approximately half of the papers provide a clear indication of source code. The number of datasets varies considerably, and papers in the Foundation category often evaluate on a fixed set of 10 datasets. This makes it difficult to detect small but consistent improvements, particularly when comparisons are made to many baselines. In our survey, 25 papers compare more algorithms than datasets, which exacerbates multiple-comparison effects and increases the risk of over-interpreting chance differences~\cite{Demsar2006comparisons}.

    \begin{table}[]
        \caption{Estimates of the number of datasets and the number of algorithms used in papers using the UEA archive to support new MTSC algorithms. }
        \label{tab:data}
        \centering \small
        \begin{tabular}{lrr|rr|rr}
        \toprule
        Year & No. papers & No. with code & No. datasets & Average & Algorithms & Average \\
        \midrule
        2019 & 3 & 2 & 26--30 & 28.5 & 8--8 & 8.0 \\
        2020 & 8 & 2 & 6--30 & 15.1 & 2--16 & 6.6 \\
        2021 & 12 & 7 & 1--30 & 19.2 & 1--15 & 7.5 \\
        2022 & 18 & 15 & 4--30 & 19.0 & 1--14 & 7.9 \\
        2023 & 15 & 6 & 3--30 & 17.3 & 3--12 & 7.1 \\
        2024 & 39 & 19 & 3--30 & 21.4 & 3--19 & 10.3 \\
        2025 & 54 & 20 & 4--30 & 19.2 & 1--27 & 10.3 \\
        2026 & 6 & 2 & 6--29 & 16.8 & 3--17 & 8.0 \\
        \bottomrule
        \end{tabular}
    \end{table}

Across the surveyed MTSC-algorithm papers, 133/155 explicitly mention a train/test evaluation, 26/155 report resampling (multiple resamples, repeated splits or train/test/validation splits), and 9/155 use some form of cross-validation. An increasingly common practice is reporting averages over multiple random seeds or runs on the same split (17 papers, concentrated in 2025). The experimental protocol is unclear or not stated in 16 papers.

    \begin{table}[]
        \caption{Published accuracy on the default test split of six classifiers proposed in the last five years on the 25 UEA datasets common to all papers. Classifiers are: ML algorithm HIVE-COTEv2 (HC2)~\cite{middlehurst21hc2}; Foundation backbone TimesNet~\cite{wu2023timesnet}, Transformer approach ConvTran~\cite{Foumani2024ConvTran}; Convolutional approaches TapNet~\cite{Zhang2020TapNet} and LiteMVTime~\cite{ismail23lite}; and deep learning algorithm ShapeNet~\cite{Li2021ShapeNet}. Best results in bold.}
        \label{tab:uea_acc}
        \centering \small
        \setlength{\tabcolsep}{5pt}
        \renewcommand{\arraystretch}{1.05}
        \begin{tabular}{lrrrrrr}
        \toprule
        Dataset & HC2 & TimesNet & ConvTran & TapNet & LiteMVTime & ShapeNet \\
        \midrule
        ArticularyWordRecognition & \textbf{0.993} & 0.973 & 0.983 & 0.987 & 0.973 & 0.987 \\
        AtrialFibrillation & 0.267 & 0.333 & \textbf{0.4} & 0.333 & 0.133 & \textbf{0.4} \\
        BasicMotions & \textbf{1} & 0.973 & \textbf{1} & \textbf{1} & \textbf{1} & \textbf{1} \\
        Cricket & \textbf{1} & 0.903 & \textbf{1} & 0.958 & 0.986 & 0.986 \\
        DuckDuckGeese & 0.56 & 0.58 & 0.62 & 0.575 & 0.18 & \textbf{0.725} \\
        Epilepsy & \textbf{1} & 0.877 & 0.985 & 0.971 & 0.993 & 0.987 \\
        EthanolConcentration & \textbf{0.772} & 0.357 & 0.361 & 0.323 & 0.692 & 0.312 \\
        ERing & \textbf{0.989} & 0.937 & 0.962 & 0.904 & 0.844 & 0.133 \\
        FaceDetection & 0.66 & \textbf{0.677} & 0.672 & 0.556 & 0.61 & 0.602 \\
        FingerMovements & 0.53 & 0.53 & 0.56 & 0.53 & 0.56 & \textbf{0.58} \\
        HandMovementDirection & 0.473 & \textbf{0.595} & 0.405 & 0.378 & 0.351 & 0.338 \\
        Handwriting & \textbf{0.548} & 0.311 & 0.375 & 0.357 & 0.4 & 0.451 \\
        Heartbeat & 0.732 & 0.78 & \textbf{0.785} & 0.751 & 0.615 & 0.756 \\
        Libras & \textbf{0.933} & 0.761 & 0.927 & 0.85 & 0.894 & 0.856 \\
        LSST & 0.643 & 0.382 & 0.615 & 0.568 & \textbf{0.664} & 0.59 \\
        MotorImagery & 0.54 & \textbf{0.61} & 0.56 & 0.59 & 0.53 & \textbf{0.61} \\
        NATOPS & 0.894 & 0.833 & \textbf{0.944} & 0.939 & 0.9 & 0.883 \\
        PenDigits & 0.979 & 0.984 & 0.987 & 0.98 & \textbf{0.989} & 0.977 \\
        PEMS-SF & \textbf{1} & 0.844 & 0.828 & 0.751 & 0.792 & 0.751 \\
        PhonemeSpectra & 0.29 & 0.146 & \textbf{0.306} & 0.175 & 0.158 & 0.298 \\
        RacketSports & \textbf{0.908} & 0.855 & 0.861 & 0.868 & 0.737 & 0.882 \\
        SelfRegulationSCP1 & 0.891 & 0.908 & \textbf{0.918} & 0.739 & 0.73 & 0.782 \\
        SelfRegulationSCP2 & 0.5 & 0.572 & \textbf{0.583} & 0.55 & 0.55 & 0.578 \\
        StandWalkJump & 0.467 & 0.533 & 0.333 & 0.4 & \textbf{0.667} & 0.533 \\
        UWaveGestureLibrary & \textbf{0.928} & 0.863 & 0.89 & 0.894 & 0.847 & 0.906 \\ \bottomrule \end{tabular}
    \end{table}

The selection of different datasets, the use of too many benchmark algorithms relative to the number of datasets, and differing experimental regimes make it hard to judge whether there has been progress. However, a more fundamental question is whether the archive is appropriate for comparison. Table~\ref{tab:uea_acc} presents published results for a subset of algorithms. This is not meant to be a comprehensive comparison, but it demonstrates that, if there has been progress in general-purpose MTSC, then the UEA archive may not be comprehensive enough to detect it reliably. ConvTran has the best average rank (2.62), but HC2, a six year old algorithm, wins on the most datasets, and there is no significant difference between the two under a paired Wilcoxon signed-rank test. We believe that an extended MTSC archive will make it easier to detect significant differences, and that an associated repository with code to reproduce experiments will make it easier to compare algorithms.

\section{The Multiverse: 2026}
\label{sec:datasets}

In this section, we give an overview of the datasets in the Multiverse archive, split into subsections denoting the different data sources that make up the archive. For the original UEA archive datasets and those donated from other published archives, we give a brief summary of the datasets present and provide references to the original archive for further details. A full table of all datasets in the archive is available in Table~\ref{tab:full_archive} in the appendix. The same summary information is discussed throughout this section. We also provide Table~\ref{tab:archive_classes}, which contains information on the label distribution of each dataset.

We loosely categorise each archive dataset by its application of origin into one of the following categories:
\begin{itemize}
\item The \textbf{Audio} category includes raw audio signals and derived features such as Mel-frequency cepstral coefficients (MFCCs). There are six audio datasets from the 2018 archive release. No new audio datasets are included in this extension.
\item \textbf{Biosignal} data consists of time series collected from medical sensors, such as EEG, magnetoencephalography (MEG), and photoplethysmography (PPG). Typically, these datasets aim to classify subjects performing tasks or physiological states such as heart and respiratory rates. This category has grown from eight datasets in the original archive to 37 in this update.
\item The \textbf{Digital} category contains series from digital sources rather than those generated by sensors. This includes data such as reactions on social media and web traffic over time. The archive expansion adds two datasets of this type.
\item \textbf{Motion}-related datasets cover various activities and movements, such as HAR, rehabilitation, vehicle dynamics, and animal behaviour. These data are typically recorded through sensors such as accelerometers and gyroscopes. The UEA archive includes 13 motion datasets. This category has grown to 65 datasets in the extended archive.
\item The \textbf{Image} category involves time series data extracted from images, including pixel intensities or object outlines. There are seven datasets in this category.
\item \textbf{Power} problems address the task of classifying electricity usage patterns, primarily gathered from household appliances. Most problems of this type are univariate, but we add three multivariate problems of this kind.
\item \textbf{Sensor}-based datasets are typically recordings from industrial or environmental sources such as air-quality and temperature sensors that do not neatly fit into other categories. Starting with two datasets, this update adds nine more.
\item \textbf{Spectro} datasets are spectrograph-based and feature spectral analyses of items such as food, alcoholic beverages, and vegetation. This category previously contained a single dataset and now contains two.
\end{itemize}
These are broad categories, intended to help guide users looking for specific types of data such as EEG or HAR. While it would be possible to categorise by more specific data sources, this would split the archive into dozens of categories, which would be difficult to maintain as more data is added.

Table~\ref{tab:archive_stats} shows summary information about the 133 archive datasets. We create bins for different dataset characteristics and display the number of datasets in each. The archive has a much wider range of dataset sizes, but it still skews towards smaller datasets on average. A large value in one characteristic is not necessarily mirrored in others. For example, the largest dataset in terms of train and test size, with millions of cases, has a series length of only 24 and ten channels. The number of classes in the archive is split almost evenly between binary and multi-class problems. While more datasets of this kind have been introduced in this iteration, there remains a lack of datasets containing unequal length series and missing values. While some problems naturally have equal length series and no missing values, others were pre-processed before we received them. Lastly, while the archive has grown in diversity in terms of dataset size, it has become less diverse in terms of data sources relative to the size of the archive. A researcher using the full archive in their experiments would find that over 75\% of the datasets are motion- or biosignal-based. Looking more closely at problem types, most datasets are EEG, HAR, or rehabilitation related. The over-representation of these problem types is one reason why we propose an archive subset for general use in Section~\ref{sec:subset}.
\begin{table}[t]
    \caption{Summary statistics for the 133 multivariate archive datasets, excluding missing value and unequal length variants. Provides statistics for the train set sizes (a), test set sizes (b), number of classes (c), number of time points (d) and number of dimensions (e) of the archive datasets. Also includes counts for series cleanliness before processing (f) and dataset type (g).}
    \label{tab:archive_stats}
    \footnotesize
    \centering

    \begin{tabular}{c c c}
		\begin{tabular}{l|c}
			\hline
            $\leq$200 & 41 (30.83\%) \\
            201-2000 & 51 (38.35\%) \\
            2001-20000 & 23 (17.29\%) \\
            $>$20000 & 18 (13.53\%) \\
            Minimum & 12  \\
            Maximum & 47,328,752 \\  \hline
		\end{tabular}
        &
		\begin{tabular}{l|c}
			\hline
            $\leq$200 & 60 (45.11\%) \\
            201-2000 & 37 (27.82\%) \\
            2001-20000 & 23 (17.29\%) \\
            $>$20000 & 13 (9.77\%) \\
            Minimum & 4 \\
            Maximum & 11,940,071 \\  \hline
		\end{tabular}
		&
        \begin{tabular}{l|c}
			\hline
            2 & 67 (50.38\%) \\
            3-5 & 38 (28.57\%) \\
            6-15 & 19 (14.29\%) \\
            $>$15 & 9 (6.77\%) \\
            Minimum & 2 \\
            Maximum & 39  \\  \hline
		\end{tabular} \\[1cm]
		(a) Train Size & (b) Test Size & (c) No. Classes \\[0.25cm]

        \multicolumn{3}{c}{
        \begin{tabular}{c c}
    	\begin{tabular}{l|c}
    		\hline
            $\leq$50 & 16 (12.03\%) \\
            51-500 & 76 (57.14\%) \\
            501-5000 & 37 (27.82\%) \\
            $>$5000 & 4 (3.01\%) \\
            Minimum & 8 \\
            Maximum & 17,984 \\  \hline
    	\end{tabular}
        &
        \begin{tabular}{l|c}
    		\hline
            $\leq$5 & 28 (21.05\%) \\
            6-20 & 43 (32.33\%) \\
            21-100 & 53 (39.85\%) \\
            $>$100 & 9 (6.77\%) \\
            Minimum & 2  \\
            Maximum & 1345  \\  \hline
    	\end{tabular} \\[1cm]
        (d) No. Time points & (e) No. Channels
        \end{tabular}
        } \\[1.25cm]

        \multicolumn{3}{c}{
        \begin{tabular}{c c}
        \begin{tabular}{l|c}
			\hline
            Clean & 119 (89.47\%) \\
            Unequal Length & 9 (6.77\%) \\
            Missing Values & 5 (3.76\%) \\ \hline
		\end{tabular}
        &
        \begin{tabular}{l|c}
    		\hline
            Audio & 6 (4.51\%) \\
            Biosignal & 37 (27.82\%) \\
            Digital & 2 (1.50\%) \\
            Motion & 65 (48.87\%) \\
            Image & 7 (5.26\%) \\
            Power & 3 (2.26\%) \\
            Sensor & 11 (8.27\%) \\
            Spectro & 2 (1.50\%) \\ \hline
    	\end{tabular} \\[1.25cm] 	
        (f) Series Cleanliness  & (g) Dataset Types
        \end{tabular}
		}
    \end{tabular}
\end{table}

\subsection{Original 30 UEA MTSC Datasets}

The original UEA multivariate archive consists of 30 datasets spanning a wide range of problem domains, sensors and data characteristics. These datasets were first released in 2018 with the explicit aim of providing a common benchmark for multivariate time series classification, mirroring the role of the UCR archive for univariate problems. Table~\ref{tab:old_datasets} provides the train and test split sizes, number of channels and series length among other summary information for each of the original datasets. Of the 30 datasets four contain unequal length series. A 26 dataset equal length subset of this collection has since underpinned large comparative studies such as the multivariate bake off~\cite{ruiz21mtsc}. In the following we provide a brief explanation of the 30 UEA datasets. We guide readers who want further details on these problems to the original publications~\cite{bagnall18mtsc,ruiz21mtsc}.

    \begin{table}
        \caption{The original 30 MTSC archive datasets. For datasets with unequal length series, both the minimum and maximum series lengths are shown.}
        \label{tab:old_datasets}
        \footnotesize
        \centering
        \begin{tabular}{ | p{4cm} | p{1.2cm} | p{1.2cm} | p{1.3cm} | p{1.25cm} | p{1cm} | p{1.3cm} | }
            \hline
            Dataset name & Train & Test & Channels & Length & Classes & Type \\
            \hline
            ArticularyWordRecognition & 275 & 300 & 9 & 144 & 25 & Motion \\ \hline
            AtrialFibrillation & 15 & 15 & 2 & 640 & 3 & Biosignal \\ \hline
            BasicMotions & 40 & 40 & 6 & 100 & 4 & Motion  \\ \hline
            CharacterTrajectories & 1422 & 1436 & 3 & 60-182 & 20 & Motion \\ \hline
            Cricket & 108 & 72 & 6 & 1197 & 12 & Motion \\ \hline
            DuckDuckGeese & 50 & 50 & 1345 & 270 & 5 & Audio \\ \hline
            ERing & 30 & 270 & 4 & 65 & 6 & Motion \\ \hline
            EigenWorms & 128 & 131 & 6 & 17984 & 5 & Motion \\ \hline
            Epilepsy & 137 & 138 & 3 & 206 & 4 & Motion \\ \hline
            EthanolConcentration & 261 & 263 & 3 & 1751 & 4 & Spectro \\ \hline
            FaceDetection & 5890 & 3524 & 144 & 62 & 2 & Biosignal \\ \hline
            FingerMovements & 316 & 100 & 28 & 50 & 2 & Biosignal \\ \hline
            HandMovementDirection & 160 & 74 & 10 & 400 & 4 & Biosignal \\ \hline
            Handwriting & 150 & 850 & 3 & 152 & 26 & Motion \\ \hline
            Heartbeat & 204 & 205 & 61 & 405 & 2 & Audio \\ \hline
            InsectWingbeat & 25000 & 25000 & 200 & 2-22 & 10 & Audio \\ \hline
            JapaneseVowels & 270 & 370 & 12 & 7-29 & 9 & Audio \\ \hline
            LSST & 2459 & 2466 & 6 & 36 & 14 & Sensor \\ \hline
            Libras & 180 & 180 & 2 & 45 & 15 & Motion \\ \hline
            MotorImagery & 278 & 100 & 64 & 3000 & 2 & Biosignal \\ \hline
            NATOPS & 180 & 180 & 24 & 51 & 6 & Motion \\ \hline
            PEMS-SF & 267 & 173 & 963 & 144 & 7 & Sensor \\ \hline
            PenDigits & 7494 & 3498 & 2 & 8 & 10 & Motion \\ \hline
            PhonemeSpectra & 3315 & 3353 & 11 & 217 & 39 & Audio \\ \hline
            RacketSports & 151 & 152 & 6 & 30 & 4 & Motion \\ \hline
            SelfRegulationSCP1 & 268 & 293 & 6 & 896 & 2 & Biosignal \\ \hline
            SelfRegulationSCP2 & 200 & 180 & 7 & 1152 & 2 & Biosignal \\ \hline
            SpokenArabicDigits & 6599 & 2199 & 13 & 4-93 & 10 & Audio \\ \hline
            StandWalkJump & 12 & 15 & 4 & 2500 & 3 & Biosignal \\ \hline
            UWaveGestureLibrary & 120 & 320 & 3 & 315 & 8 & Motion \\ \hline
        \end{tabular}
    \end{table}

A substantial fraction of the original archive consists of human activity recognition problems derived from accelerometers, gyroscopes and motion capture data. \textbf{BasicMotions} captures smartwatch accelerometer and gyroscope readings from subjects performing everyday activities with the classes resting, walking, running and playing badminton. \textbf{Epilepsy} simulates daily tasks and mimicked seizures with data tracked using an accelerometer. \textbf{Cricket} and \textbf{RacketSports} also involve classifying sports-related arm movements from multi-axis IMUs placed on the wrist, detecting umpire calls and stroke types respectively. \textbf{UWaveGestureLibrary} and \textbf{NATOPS} focus on recognising gestures from handheld devices or motion capture systems, with the first recording simple hand movements and the latter looking for the Naval Air Training and Operating Procedures Standardization motions used to control plane movement. \textbf{Handwriting} looks to track the motion of writing handwritten characters using smartwatch data.

Other motion data in the archive comes from different sources. \textbf{CharacterTrajectories} and \textbf{PenDigits} provide pen or stylus trajectories representing handwritten characters or digits respectively rather than IMU data, while \textbf{Libras} captures 15 different Brazilian sign language motions using videos of hand movements. \textbf{ArticularyWordRecognition} captures motions during the pronunciation of 25 English words using head sensors and tongue movements from an electromagnetic articulograph. Finally, \textbf{ERing} uses electric field sensors in a finger ring to detect hand postures and \textbf{EigenWorms} captures the motion of roundworms over long time windows using outlines to determine mutations.

The archive also contains a large collection of biosignal datasets built from ECG, EEG and MEG recordings. \textbf{AtrialFibrillation} consists of two channel ECG segments, with the task of predicting whether an episode of atrial fibrillation will terminate spontaneously or persist. \textbf{StandWalkJump} uses ECG segments to determine which physical activity a subject is performing. \textbf{FaceDetection} presents 1.5 second MEG recordings from 16 patients where the subject has been shown a picture of a face or a scrambled image, with the task being to determine which picture was shown. The group of \textbf{FingerMovements}, \textbf{HandMovementDirection}, \textbf{MotorImagery}, and \textbf{SelfRegulationSCP1/2} are derived from brain computer interface (BCI) experiments, where participants either execute or imagine specific movements while EEG or MEG is recorded over a varying amount of channels. The aim is to infer motor intent or cursor direction directly from the neural signal.

The last major grouping in the original archive are a collection of audio-based datasets. \textbf{JapaneseVowels} and \textbf{SpokenArabicDigits} are speech recognition benchmarks where channels represent cepstral or spectral features of spoken phonemes or digits over time. Still based on human speech, \textbf{PhonemeSpectra} encodes phonemes as multi-band spectrogram slices. For all of these the goal is to classify the phoneme or digit spoken. \textbf{DuckDuckGeese} and \textbf{InsectWingbeat} involve distinguishing between species from high-dimensional audio or wingbeat spectra. The last audio dataset is \textbf{Heartbeat} where the aim is to distinguish between normal and abnormal recordings from frequency spectrograms.

The remaining datasets fall into smaller categories. \textbf{EthanolConcentration} represents raw spectra of ethanol and water solutions contained in whisky bottles measured at different ethanol concentrations. The aim of the dataset is to detect lower concentrations of ethanol which can be fraudulent. \textbf{LSST} contains simulated multi-band astronomical light curves for astronomical objects created in preparation for the Large Synoptic Survey Telescope mission. \textbf{PEMS-SF} consists of traffic flow measurements from hundreds of highway sensors in San Francisco, where the goal is to classify the day of the week based on traffic patterns.

\subsection{EEG/MEG Classification Archive}

These datasets are drawn primarily from studies using electroencephalography (EEG), with one additional magnetoencephalography (MEG) dataset. EEG records electrical activity from the scalp using multiple electrodes, each producing a separate signal, together forming a multivariate time series. EEG data is widely used across medicine, psychology, and brain-computer interfacing (BCI). The full archive is described in~\cite{rushbrooke2026eeg}. The original archive contains 30 datasets, of which six are from the UEA collection, two are single-channel, and two are available only directly from the authors. We include the remaining 20 datasets in the Multiverse. Together, they span all three application areas and cover a broad range of characteristics, including channel counts, sampling frequencies, and series lengths.

\noindent\textbf{Alzheimers}~\cite{miltiadous2023alzheimers} contains recordings from subjects with Alzheimer's disease (n=36), frontotemporal dementia (n=23), and healthy controls (n=29). We extract a 60 second window from the middle of each eyes-closed resting-state recording to form a 3-class classification problem.

\noindent\textbf{PhotoStimulation}~\cite{ntetska2025photostimulation} also forms a 3-class problem using the same subject groups. Subjects viewed a flashing screen whose frequency increased from 5 Hz to 30 Hz, and we select an 18 second window at the 10 Hz stimulation stage.

\noindent \textbf{Blink}~\cite{ortiz2021bci} distinguishes between long and short blinks. Six subjects each performed 50 long or short blinks, and we extract a 2 second window around each blink, beginning 0.5 seconds before the trigger.

\noindent \textbf{EyesOpenShut}~\cite{roesler13eye} is an eye-state dataset in which the task is to determine whether a single participant's eyes are open or closed. We segment the recordings into one second intervals and remove extreme outliers caused by recording irregularities.

\noindent \textbf{OpenCloseFist}, \textbf{FeetHands}, \textbf{ImaginedOpenCloseFist}, and \textbf{ImaginedFeetHands}~\cite{schalk2022motorimagery,schalk2004bci2000} come from a motor movement collection. The first two involve real movement, whereas the latter two involve motor imagery. \textbf{OpenCloseFist} and \textbf{ImaginedOpenCloseFist} classify left versus right hand movement, whereas \textbf{FeetHands} and \textbf{ImaginedFeetHands} classify hand versus foot movement. The data was collected from 109 participants, each completing 45 trials per task.

\noindent \textbf{InnerSpeech}, \textbf{PronouncedSpeech}, and \textbf{VisualSpeech}~\cite{nieto2021innerspeech} classify one of four directions, up, down, left, or right, from imagined speech, spoken speech, or visual imagery, respectively. Ten participants took part, each completing at least 50 trials, and each instance is a 3.5 second window beginning 0.5 seconds before the direction prompt.

\noindent \textbf{LongIntervalTask} and \textbf{ShortIntervalTask}~\cite{arun2023timingtask} investigate the effect of Parkinson's disease on a simple timing task. Participants with Parkinson's disease and healthy controls were asked to hold down a button for either 3 or 7 seconds after a visual cue. We form two classification datasets, one for short intervals and one for long intervals, where the task is to distinguish participants with Parkinson's disease from healthy controls.

\noindent \textbf{MindReading}~\cite{klami2011icann} is the only MEG dataset. It consists of recordings from a single subject watching silent films across two sessions on consecutive days. The task is to predict which of five movie categories the subject was watching from one-second windows.

\noindent \textbf{SitStand}~\cite{guzman2022sitstandpaper,guzman2024sitstanddataset} is a motor-imagery dataset involving real and imagined movement between standing and sitting, producing a 4-class problem. Thirty-two participants completed 50 instances of each task, and we extract a 5 second window around each movement, starting 200ms before the trigger.

\noindent \textbf{SongFamiliarity}~\cite{girard2025songfamiliarity} investigates whether EEG can be used to detect song recognition. Twenty-nine participants listened to 50 music excerpts, indicated when a song sounded familiar, and then selected the correct song from four options. We define the class label according to whether the correct song was selected.

\noindent \textbf{ButtonPress} is a simple calibration dataset containing recordings from a single participant either pressing a button or remaining idle. We combine 120 button presses with 120 resting-state segments drawn randomly from a 30 minute recording.

\noindent \textbf{FeedbackButton} is related to \textbf{ButtonPress}, with the task again being to detect whether a button is pressed. In this experiment, participants were shown a simple prompt and responded by either pressing or not pressing the button, after which they received positive or negative feedback based on their action. We use a 900ms window starting 200ms before the initial stimulus and ending before the button press. The dataset contains recordings from 57 participants.

\noindent \textbf{MatchingPennies}~\cite{appelhoff2018matchingpennies} is based on a psychology task in which participants try to trick a computer by raising one arm while imagining raising the other. Seven participants took part, raising either their left or right hand 100 times each and receiving live feedback on the computer's prediction.

\noindent \textbf{LowCost}~\cite{peterson2021lowcost} is a BCI dataset designed to evaluate a lower-cost EEG system. The task is to classify idle versus imagined dominant-hand movement. Ten participants completed 60 recordings for each class.

\begin{table}
    \caption{The 20 new MTSC archive datasets from the EEG/MEG classification archive multivariate datasets, all of type Biosignal.}
    \label{tab:new_eeg_datasets}
    \footnotesize
    \centering
    \begin{tabular}{ | p{3.5cm} | p{1.2cm} | p{1.2cm} | p{1.3cm} | p{1.25cm} | p{1cm} | p{1.3cm} | }
        \hline
        Dataset name & Train & Test & Channels & Length & Classes & Type \\
        \hline
        Alzheimers & 45 & 43 & 19 & 15000 & 3 & Biosignal \\ \hline
        Blink & 500 & 450 & 4 & 510 & 2 & Biosignal \\ \hline
        ButtonPress & 120 & 120 & 32 & 1000 & 2 & Biosignal \\ \hline
        EyesOpenShut & 56 & 42 & 14 & 128 & 2 & Biosignal \\ \hline
        FeedbackButton & 1700 & 1150 & 61 & 450 & 2 & Biosignal \\ \hline
        FeetHands & 2835 & 1890 & 64 & 640 & 2 & Biosignal \\ \hline
        ImaginedFeetHands & 2835 & 1935 & 64 & 640 & 2 & Biosignal \\ \hline
        ImaginedOpenCloseFist & 2835 & 1890 & 64 & 640 & 2 & Biosignal \\ \hline
        InnerSpeech & 250 & 200 & 128 & 3584 & 4 & Biosignal \\ \hline
        LongIntervalTask & 1660 & 1120 & 63 & 4750 & 2 & Biosignal \\ \hline
        LowCost & 600 & 600 & 15 & 375 & 2 & Biosignal \\ \hline
        MatchingPennies & 700 & 700 & 10 & 10000 & 2 & Biosignal \\ \hline
        MindReading & 727 & 653 & 204 & 200 & 5 & Biosignal \\ \hline
        OpenCloseFist & 2835 & 1890 & 64 & 640 & 2 & Biosignal \\ \hline
        PhotoStimulation & 37 & 36 & 19 & 9000 & 3 & Biosignal \\ \hline
        PronouncedSpeech & 250 & 200 & 128 & 3584 & 4 & Biosignal \\ \hline
        ShortIntervalTask & 1360 & 1420 & 63 & 2750 & 2 & Biosignal \\ \hline
        SitStand & 3200 & 3200 & 17 & 1250 & 4 & Biosignal \\ \hline
        SongFamiliarity & 700 & 750 & 32 & 3000 & 2 & Biosignal \\ \hline
        VisualSpeech & 250 & 200 & 128 & 3584 & 4 & Biosignal \\ \hline
    \end{tabular}
\end{table}

\subsection{{\monster}}

\begin{table}
    \caption{The 18 new MTSC archive datasets from the Monash {\monster} archive multivariate datasets.}
    \label{tab:new_monster_datasets}
    \footnotesize
    \centering
    \begin{tabular}{ | p{3cm} | p{1.45cm} | p{1.45cm} | p{1.3cm} | p{1.25cm} | p{1cm} | p{1.3cm} | }
        \hline
        Dataset name & Train & Test & Channels & Length & Classes & Type \\
        \hline
        CrowdSourced   & 9,456      & 2,833      & 14  & 256 & 2    & Biosignal \\ \hline
        DREAMERA       & 140,638    & 29,608     & 14  & 256 & 2    & Biosignal \\ \hline
        DREAMERV       & 140,638    & 29,608     & 14  & 256 & 2    & Biosignal \\ \hline
        FordChallenge  & 29,003     & 7,254      & 30  & 40  & 2    & Sensor \\ \hline
        LenDB          & 983,483    & 261,459    & 3   & 540 & 2    & Sensor \\ \hline
        Opportunity    & 12,184     & 5,202      & 113 & 100 & 5    & Motion \\ \hline
        PAMAP2         & 28,961     & 9,895      & 52  & 100 & 12   & Motion \\ \hline
        S2Agri-10pc-17 & 4,661,858  & 1,189,023  & 10  & 24  & 17   & Image \\ \hline
        S2Agri-10pc-34 & 4,661,858  & 1,189,023  & 10  & 24  & 29   & Image \\ \hline
        S2Agri-17      & 47,328,752 & 11,940,071 & 10  & 24  & 17   & Image \\ \hline
        S2Agri-34      & 47,328,752 & 11,940,071 & 10  & 24  & 34** & Image \\ \hline
        Skoda          & 11,290     & 2,827      & 60  & 100 & 11   & Motion \\ \hline
        STEW           & 21,384     & 7,128      & 14  & 256 & 2    & Biosignal \\ \hline
        Tiselac        & 79,907     & 19,780     & 10  & 23  & 9    & Image \\ \hline
        UCIActivity    & 7,898      & 2,401      & 9   & 128 & 6    & Motion \\ \hline
        USCActivity    & 43,060     & 13,168     & 6   & 100 & 12   & Motion \\ \hline
        WISDM          & 11,989     & 5,177      & 3   & 100 & 6    & Motion \\ \hline
        WISDM2         & 98,094     & 50,940     & 3   & 100 & 6    & Motion \\ \hline

        \multicolumn{7}{l}{** The test split is missing some classes that are present in the training split.} \\
    \end{tabular}
\end{table}

We take 18 datasets directly from the {\monster} archive~\cite{dempster2025monster} and include EEG, HAR, satellite-image, seismological, and other sensor-derived time series. These are summarised in Table~\ref{tab:new_monster_datasets}. The TimeSen2Crop introduced in the benchmark has since been excluded due to issues with its labelling. We exclude it from the Multiverse archive as well.

\noindent \textbf{CrowdSourced}, \textbf{DREAMERA}, \textbf{DREAMERV}, and \textbf{STEW} are all EEG datasets, where the task is to predict cognitive state or activity using data captured from multiple participants with various EEG headsets. In each case, the data was recorded over 14 channels at, or downsampled to, a sampling frequency of 128 Hz. The time series have been processed so that each series represents 2 seconds of data. \textit{CrowdSourced} involves classifying the states \textit{eyes open} or \textit{eyes closed}. The \textit{Dreamer} datasets involve classifying \textit{arousal} (A) or \textit{valence} (V). \textit{STEW} involves classifying \textit{cognitive workload} (high or low).

\noindent \textbf{FordChallenge} contains data recorded from a variety of physiological, environmental, and vehicular sensors while a subject is driving a car. The data was recorded over 30 channels at a sampling rate of 10 Hz. Each time series represents 4 seconds of data. The classification task involves predicting whether a driver is \textit{alert} or \textit{distracted}.

\noindent \textbf{LenDB} consists of seismological data recorded from various monitoring stations across the globe. The data was recorded over 3 channels at a sampling rate of 20 Hz. Each time series represents just under 30 seconds of data. The task is to distinguish between \textit{earthquake} and \textit{noise}.

\noindent \textbf{Opportunity}, \textbf{PAMAP2}, \textbf{Skoda}, \textbf{UCIActivity}, \textbf{USCActivity}, \textbf{WISDM}, and    \textbf{WISDM2} are all human activity recognition (HAR) datasets, where the task is to detect the activity of the experimental subject based on data recorded from various wearable sensors. \textit{Opportunity} contains data collected using a variety of sensors over a total of 113 channels at a sampling rate of 30 Hz. The task involves identifying five different activities: \textit{stand}, \textit{walk}, \textit{sit}, \textit{lie}, and \textit{null} (i.e., inactive). \textit{PAMAP2} contains data collected using multiple sensors over a total of 52 channels at a sampling rate of 100 Hz. The task is to identify one of 12 activity states. \textit{Skoda} contains data collected using multiple sensors during car maintenance tasks at a sampling rate of approximately 100 Hz. The task is to identify one of 10 distinct gestures. \textit{UCIActivity} contains data recorded using smartphones over 9 channels at a sampling rate of 50 Hz. The task is to identify one of six activities: \textit{walking}, \textit{walking upstairs}, \textit{walking downstairs}, \textit{sitting}, \textit{standing}, and \textit{lying down}. \textit{USCActivity} contains data recorded from various sensors over 6 channels at a sampling rate of 100 Hz. The task is to identify one of 12 activities. \textit{WISDM} and \textit{WISDM2} contain data recorded using accelerometers over three channels at a sampling rate of 20 Hz. The task is to identify one of six activities: \textit{walking}, \textit{jogging}, \textit{stairs}, \textit{sitting}, \textit{standing}, and \textit{lying down}.

\noindent The \textbf{S2Agri} datasets (\textbf{S2Agri-10pc-17}, \textbf{S2Agri-10pc-34}, \textbf{S2Agri-17}, and \textbf{S2Agri-34}) and \textbf{Tiselac} consist of pixel-level satellite-image time series collected over France and Reunion Island, respectively, where the task is to classify ground cover type or land use. In each case, each time series represents the changing values for a given pixel over time, and different channels represent different spectral bands, or values derived from them. \textit{Tiselac} consists of time series derived from images of Reunion Island captured by the Landsat 8 satellite, sampled approximately every 16 days, with a total of 10 channels. The \textit{S2Agri} datasets consist of time series derived from images of France captured by Sentinel-2 satellites, sampled approximately every 10 days, with a total of 10 channels. \textit{S2Agri-17} and \textit{S2Agri-34} represent two versions of the dataset with different label sets. \textit{S2Agri-10pc-17} and \textit{S2Agri-10pc-34} are subsets of the same datasets containing approximately 10\% of the original land parcels, where each land parcel is a distinct geographic region, i.e., these are not simply random subsamples of the larger datasets. As a quirk of the way S2Agri-34 has been set up to avoid using cases from the same `land parcel', some classes are missing in the test set for all the cross-validation splits used in the released version of the data.

\subsection{Rehab-Pile Datasets}

These datasets come from Rehab-Pile, a benchmark archive for physical rehabilitation assessment using skeleton-based motion sequences~\cite{fawaz2025rehab}. It aggregates seven public repositories and treats each exercise as a separate dataset. In total, Rehab-Pile contains 39 classification datasets and 21 regression datasets. We include only the classification datasets in the Multiverse. We summarise the repositories in Table~\ref{tab:rehab_pile_summary}, and the individual datasets added to the Multiverse in Table~\ref{tab:new_rehab_datasets}.

\begin{figure}[!htb]
    \centering
    \includegraphics[width=0.8\linewidth]{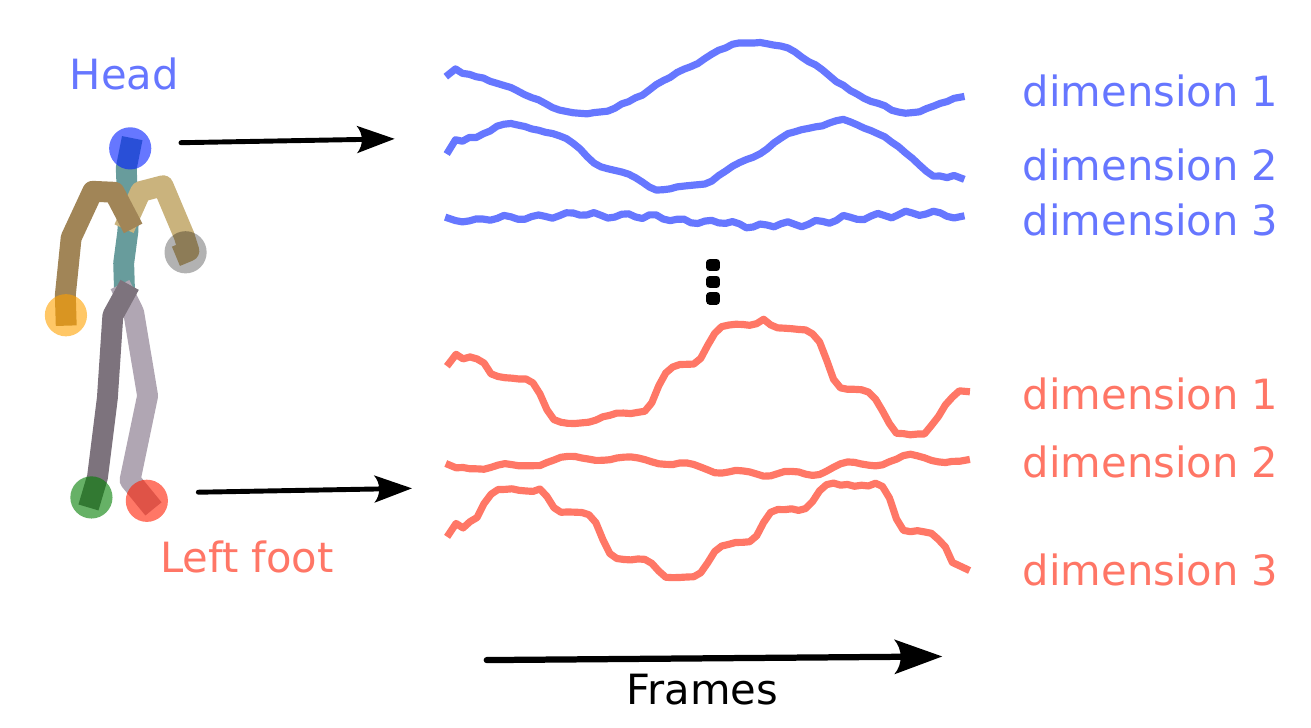}
    \caption{Given a skeleton sequence of $F$ frames, where each frame contains $J$ joints described by $D$-dimensional features, the sequence is represented as a multivariate time series of length $F$ with $J \cdot D$ channels.}
    \label{fig:skeleton-to-mts}
\end{figure}

Each skeleton sequence is represented as a multivariate time series. For a sequence of $F$ frames, where each frame contains $J$ joints described in $D$ dimensions, the resulting series has length $F$ and $J \cdot D$ channels, as illustrated in Figure~\ref{fig:skeleton-to-mts}. The IRDS, KERAAL, KIMORE, and SPHERE sequences were collected using Kinect technology~\cite{kinectSensor-rehab}. In most cases, the task is to predict whether an exercise was performed correctly, unless stated otherwise. Because sequences within a dataset are recorded under consistent acquisition settings, the authors of Rehab-Pile~\cite{fawaz2025rehab} resample all sequences to a common length equal to the mean sequence length using the Fourier-based resampling method provided by \textit{SciPy}~\cite{virtanen2020scipy}.

\begin{table*}[!htb]
    \centering
    \caption{Summary of the Rehab-Pile dataset repositories included in the Multiverse archive.}
    \label{tab:rehab_pile_summary}
    \footnotesize
    \begin{tabular}{lccccc p{6.3cm}}
        \hline
        Repository & Subjects & Ex. & Joints & Dim. & Ch. & Notes \\
        \hline
        IRDS    & 29  & 9  & 25 & 3 & 75 & Binary, one dataset per exercise \\
        KERAAL  & 11  & 3  & 11 & 7 & 77 & Binary and multiclass, 6 total \\
        KIMORE  & 71  & 5  & 18 & 3 & 54 & Regression reformulated as binary classification \\
        KINECAL & --  & 4  & 25 & 3 & 75 & Binary clinical risk prediction \\
        SPHERE  & 6   & 1  & 15 & 3 & 45 & Anomaly detection reformulated as binary classification \\
        UCDHE   & 106 & 2  & 8  & 2 & 16 & Binary and multiclass, 4 total \\
        UIPRMD  & 10  & 10 & 22 & 3 & 66 & Binary, one dataset per exercise \\
        \hline
    \end{tabular}
\end{table*}

    \begin{table}
        \caption{The 39 new MTSC archive datasets from the classification portion of the Rehab-Pile archive, all of type Motion.}
        \label{tab:new_rehab_datasets}
        \footnotesize
        \centering
        \begin{tabular}{ | p{3.5cm} | p{1.2cm} | p{1.2cm} | p{1.3cm} | p{1.25cm} | p{1cm} | p{1.3cm} | }
            \hline
            Dataset name & Train & Test & Channels & Length & Classes & Type \\
            \hline
            IRDS-EFL & 223 & 26 & 75 & 60 & 2 & Motion \\ \hline
            IRDS-EFR & 256 & 15 & 75 & 75 & 2 & Motion \\ \hline
            IRDS-SAL & 213 & 56 & 75 & 82 & 2 & Motion \\ \hline
            IRDS-SAR & 218 & 33 & 75 & 76 & 2 & Motion \\ \hline
            IRDS-SFE & 237 & 15 & 75 & 95 & 2 & Motion \\ \hline
            IRDS-SFL & 253 & 116 & 75 & 111 & 2 & Motion \\ \hline
            IRDS-SFR & 295 & 15 & 75 & 106 & 2 & Motion \\ \hline
            IRDS-STL & 226 & 38 & 75 & 66 & 2 & Motion \\ \hline
            IRDS-STR & 221 & 60 & 75 & 66 & 2 & Motion \\ \hline
            KERAAL-CTK & 271 & 14 & 77 & 276 & 2 & Motion\\ \hline
            KERAAL-CTK-MC & 271 & 14 & 77 & 276 & 4** & Motion\\ \hline
            KERAAL-ELK & 247 & 10 & 77 & 295 & 2 & Motion\\ \hline
            KERAAL-ELK-MC & 243 & 14 & 77 & 295 & 4** & Motion\\ \hline
            KERAAL-RTK & 256 & 14 & 77 & 264 & 2 & Motion\\ \hline
            KERAAL-RTK-MC & 257 & 13 & 77 & 264 & 4** & Motion\\ \hline
            KIMORE-LA-C & 64 & 7 & 54 & 725 & 2 & Motion \\ \hline
            KIMORE-LT-C & 64 & 7 & 54 & 798 & 2 & Motion \\ \hline
            KIMORE-PR-C & 64 & 7 & 54 & 847 & 2 & Motion \\ \hline
            KIMORE-Sq-C & 64 & 7 & 54 & 557 & 2 & Motion \\ \hline
            KIMORE-TR-C & 64 & 7 & 54 & 813 & 2 & Motion \\ \hline
            KINECAL-3WFV & 48 & 12 & 75 & 257 & 2 & Motion \\ \hline
            KINECAL-GGFV & 49 & 13 & 75 & 296 & 2 & Motion \\ \hline
            KINECAL-QSEC & 69 & 18 & 75 & 592 & 2 & Motion \\ \hline
            KINECAL-QSEO & 68 & 17 & 75 & 593 & 2 & Motion \\ \hline
            SPHERE-WUS & 44 & 4 & 45 & 214 & 2 & Motion \\ \hline
            UCDHE-MP & 1495 & 362 & 16 & 161 & 2 & Motion \\ \hline
            UCDHE-MP-MC & 1495 & 362 & 16 & 161 & 4 & Motion \\ \hline
            UCDHE-Rowing & 1992 & 440 & 16 & 161 & 2 & Motion \\ \hline
            UCDHE-Rowing-MC & 1992 & 440 & 16 & 161 & 5 & Motion \\ \hline
            UIPRMD-DS-C & 144 & 36 & 66 & 81 & 2 & Motion \\ \hline
            UIPRMD-HS-C & 92 & 18 & 66 & 69 & 2 & Motion \\ \hline
            UIPRMD-IL-C & 84 & 18 & 66 & 77 & 2 & Motion \\ \hline
            UIPRMD-SASLR-C & 126 & 20 & 66 & 63 & 2 & Motion \\ \hline
            UIPRMD-SL-C & 122 & 18 & 66 & 85 & 2 & Motion \\ \hline
            UIPRMD-SSA-C & 108 & 18 & 66 & 74 & 2 & Motion \\ \hline
            UIPRMD-SSE-C & 108 & 18 & 66 & 67 & 2 & Motion \\ \hline
            UIPRMD-SSIER-C & 100 & 20 & 66 & 74 & 2 & Motion \\ \hline
            UIPRMD-SSS-C & 90 & 18 & 66 & 66 & 2 & Motion \\ \hline
            UIPRMD-STS-C & 132 & 36 & 66 & 88 & 2 & Motion \\ \hline

            \multicolumn{7}{l}{** These datasets are missing classes in the test set which are present in the train set.} \\
        \end{tabular}
    \end{table}

\noindent \textbf{IRDS.} The IntelliRehabDS (IRDS) repository~\cite{IRDS-rehab} contains data for nine rehabilitation exercises collected from 14 healthy subjects and 15 patients. The nine datasets correspond to left and right elbow flexion-extension (EFL, EFR), left and right shoulder front raises (SFL, SFR), left and right shoulder lateral raises (SAL, SAR), left and right side leg raises (STL, STR), and an overhead arm raise with hands together (SFE).

\noindent \textbf{KERAAL.} The KERAAL repository~\cite{keraal,keraal2} contains upper-body rehabilitation data from five healthy subjects and six patients collected in a clinical study. Participants performed three exercises: lateral bending of the trunk (ELK), trunk rotation (RTK), and upper limbs flexed at 90 degrees (CTK). Unlike most of the other repositories, KERAAL also includes detailed annotations of execution errors. This yields a multiclass version of each exercise, producing three additional classification datasets with the suffix -MC. See~\cite{fawaz2025rehab} for details of the class definitions. Due to the cross-validation setup used, some classes are missing in the test set for the released multiclass folds.

\noindent \textbf{KIMORE.} The Kinematic Assessment of Movement for Remote Monitoring of Physical Rehabilitation (KIMORE) repository~\cite{kimore} contains data from 71 subjects, including 40 healthy participants and 31 patients, performing five exercises. Each subject performed multiple repetitions of the same exercise, resulting in one sequence per subject containing all repetitions. The five exercises are lifting of the arms (LA), lateral tilt of the trunk with the arms extended (LT), trunk rotation (TR), pelvis rotations on the transverse plane (PR), and squatting (Sq). KIMORE was originally designed as a regression task, with the goal of predicting an expert-assigned score between 0 and 100 that reflects exercise quality. Rehab-Pile~\cite{fawaz2025rehab} reformulates this as a binary classification problem by assigning scores greater than 50 to class \textit{good} and scores of 50 or lower to class \textit{bad}. To distinguish these problems from the original archive, we add the suffix -C.

\noindent \textbf{KINECAL.} Unlike the other rehabilitation repositories, which focus on exercise quality, the KINECAL repository~\cite{kinecal} aims to determine whether a patient is clinically at risk based on observed performance. The retained datasets are therefore framed as binary clinical risk prediction tasks with classes \textit{clinically at risk} and \textit{clinically not at risk}. The repository contains 11 exercises, but only four were retained by~\cite{fawaz2025rehab}, since the remaining exercises contain only a single class and are therefore unsuitable for classification. The retained datasets are Get Up And Go Front View (GGFV), 3m Walk Front View (3WFV), Quiet Standing Eyes Open (QSEO), and Quiet Standing Eyes Closed (QSEC).

\noindent \textbf{SPHERE.} The Sensor Platform for Healthcare in a Residential Environment (SPHERE) repository~\cite{sphere} contains data from six subjects performing a single exercise, Walking Up Stairs (WUS). This dataset was originally intended for anomaly detection. In~\cite{fawaz2025rehab}, it is reformulated as a binary classification problem in which the goal is to determine whether a sequence contains an anomaly.

\noindent \textbf{UCDHE.} The University College Dublin Human Exercises (UCDHE) repository~\cite{ucdhe-1,ucdhe-2,ucdhe-3} contains two exercises, Military Press (MP) and Rowing, performed by 52 and 54 healthy subjects, respectively. The sequences are extracted from video using the 2D OpenPose model~\cite{open-pose}, making this repository distinct from the 3D motion-capture based sources above. As with KERAAL, each exercise is available in two versions. The first is a binary classification task that predicts whether the exercise was performed correctly. The second is a multiclass version in which the \textit{incorrect} class is subdivided according to the specific execution error.

\noindent \textbf{UIPRMD.} The University of Idaho Physical Rehabilitation Movement Data (UIPRMD) repository~\cite{uiprmd} contains 10 exercises performed by 10 healthy subjects. It was originally proposed for both regression and classification, but we include only the classification tasks in the Multiverse. The 10 exercises are Side Lunge (SL), Standing Shoulder Extension (SSE), Deep Squat (DS), Standing Active Straight Leg Raise (SASLR), Standing Shoulder Abduction (SSA), Standing Shoulder Scaption (SSS), Hurdle Step (HS), Standing Shoulder Internal-External Rotation (SSIER), Inline Lunge (IL), and Sit to Stand (STS).

\subsection{Discretised Regression Archive Datasets}

We convert the 15 multivariate problems from the TSR dataset archive for classification by discretising the target variables for each problem. Summary information for these problems is presented in Table~\ref{tab:new_tsr_datasets}.

    \begin{table}
        \caption{The 15 MTSC archive datasets from the discretised regression archive. For datasets with unequal length series, both the minimum and maximum series lengths are shown. HouseholdPowerConsumption is compressed to HPC for formatting.}
        \label{tab:new_tsr_datasets}
        \footnotesize
        \centering
        \begin{tabular}{ | p{4.5cm} | p{1.2cm} | p{1.2cm} | p{1.3cm} | p{1.25cm} | p{1cm} | p{1.3cm} | }
            \hline
            Dataset name & Train & Test & Channels & Length & Classes & Type \\
            \hline
            AppliancesEnergy\_disc & 95 & 42 & 24 & 144 & 2 & Power \\ \hline
            AustraliaRainfall\_disc & 112,186 & 48081 & 3 & 24 & 4 & Sensor \\ \hline
            BIDMC32HR\_disc & 5550 & 2399 & 2 & 4000 & 3 & Biosignal \\ \hline
            BIDMC32RR\_disc & 5471 & 2399 & 2 & 4000 & 3 & Biosignal \\ \hline
            BIDMC32SpO2\_disc & 5550 & 2399 & 2 & 4000 & 2 & Biosignal \\ \hline
            BeijingPM10Quality\_disc & 11,918 & 5048 & 9 & 24* & 2 & Sensor \\ \hline
            BeijingPM25Quality\_disc & 11,918 & 5048 & 9 & 24* & 2 & Sensor \\ \hline
            BenzeneConcentration\_disc & 3349 & 5163 & 8 & 240* & 2 & Sensor \\ \hline
            HPC1\_disc & 745 & 686 & 5 & 1440* & 3 & Power \\ \hline
            HPC2\_disc & 745 & 686 & 5 & 1440* & 3 & Power \\ \hline
            IEEEPPG\_disc & 1768 & 1328 & 5 & 1000 & 3 & Biosignal \\ \hline
            LiveFuelMoistureContent\_disc & 3493 & 1510 & 7 & 365 & 2 & Spectro \\ \hline
            NewsHeadlineSentiment\_disc & 58,213 & 24,951 & 3 & 144 & 3 & Digital \\ \hline
            NewsTitleSentiment\_disc & 58,213 & 24,951 & 3 & 144 & 3 & Digital \\ \hline
            PPGDalia\_disc & 43,215 & 21,482 & 4 & 256-512*** & 3 & Biosignal \\ \hline

            \multicolumn{7}{l}{* These datasets contain missing values.} \\
            \multicolumn{7}{l}{*** Series length differs by channel instead of case for PPGDalia.} \\
        \end{tabular}
    \end{table}

\noindent \textbf{AppliancesEnergy}, \textbf{HouseholdPowerConsumption1}, and \textbf{HouseholdPowerConsumption2} are energy-consumption problems built from home sensor data. AppliancesEnergy contains daily multivariate series from a smart home, including temperature and humidity measurements from multiple rooms together with outdoor and airport weather variables, with the target being total daily appliance energy consumption. The HouseholdPowerConsumption datasets contain daily series from a house in Sceaux, France, with channels corresponding to voltage, current, and three sub-metering readings. They differ only in the target, with HouseholdPowerConsumption1 predicting active power and HouseholdPowerConsumption2 reactive power.

\noindent \textbf{BenzeneConcentration}, \textbf{BeijingPM25Quality}, and \textbf{BeijingPM10Quality} are air-quality monitoring problems. BenzeneConcentration uses hourly recordings from chemical and environmental sensors in an Italian city to predict benzene concentration in the next hour. The two Beijing datasets are derived from a multi-site air-quality network, where each case is a daily nine-channel series formed from pollutant and meteorological variables. They share the same inputs and split, differing only in whether the target is daily PM2.5 or PM10 concentration.

\noindent \textbf{LiveFuelMoistureContent} and \textbf{AustraliaRainfall} are environmental prediction problems. LiveFuelMoistureContent uses one year of daily satellite-derived reflectance in seven spectral bands to estimate live fuel moisture content, an important factor in bushfire risk. AustraliaRainfall is derived from Australian weather-station data and uses hourly average, maximum, and minimum air temperature to predict total daily rainfall.

\noindent \textbf{PPGDalia}, \textbf{IEEEPPG}, \textbf{BIDMC32RR}, \textbf{BIDMC32HR}, and \textbf{BIDMC32SpO2} are medical datasets based on PPG signals and related physiological measurements. PPGDalia combines wrist PPG with three-axis accelerometer data from free-living activities, while IEEEPPG uses PPG and accelerometer signals recorded from participants running on a treadmill. Both are used to predict heart rate. The three BIDMC32 datasets contain paired PPG and ECG windows and differ only in the target, namely respiratory rate, heart rate, or blood oxygen saturation.

\noindent \textbf{NewsHeadlineSentiment} and \textbf{NewsTitleSentiment} are sentiment-analysis problems derived from social feedback time series for online news items. Each case is a three-channel series of reactions on Facebook, Google+, and LinkedIn over two days. The inputs and train-test splits are identical, but the targets differ: headline sentiment for NewsHeadlineSentiment and title sentiment for NewsTitleSentiment.

\begin{figure}[!h]
    	\centering
        \vspace{0.5cm}
        \begin{tabular}{ c c c }
            \includegraphics[width=0.3\linewidth, trim={0cm 0cm 1.3cm 0.5cm},clip]{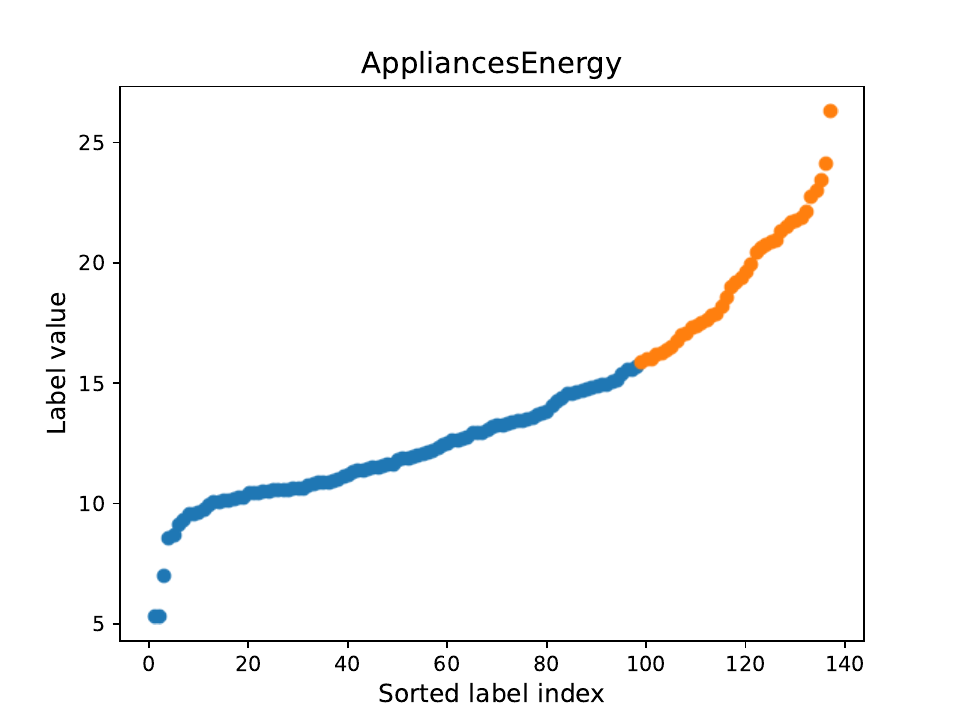} &
            \includegraphics[width=0.3\linewidth, trim={0cm 0cm 1.3cm 0.5cm},clip]{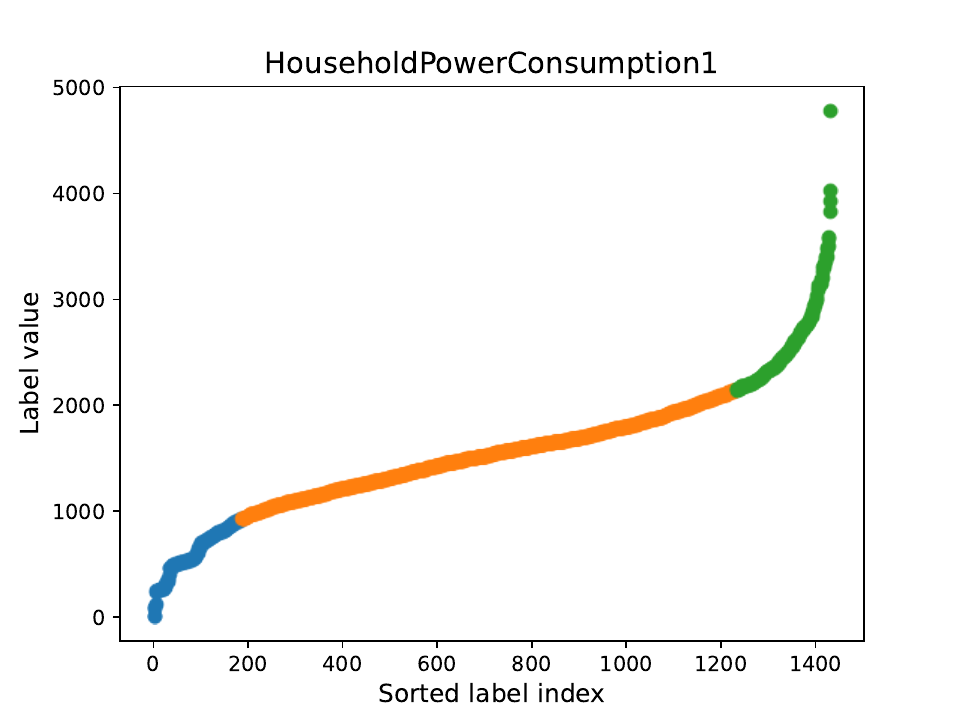} &
            \includegraphics[width=0.3\linewidth, trim={0cm 0cm 1.3cm 0.5cm},clip]{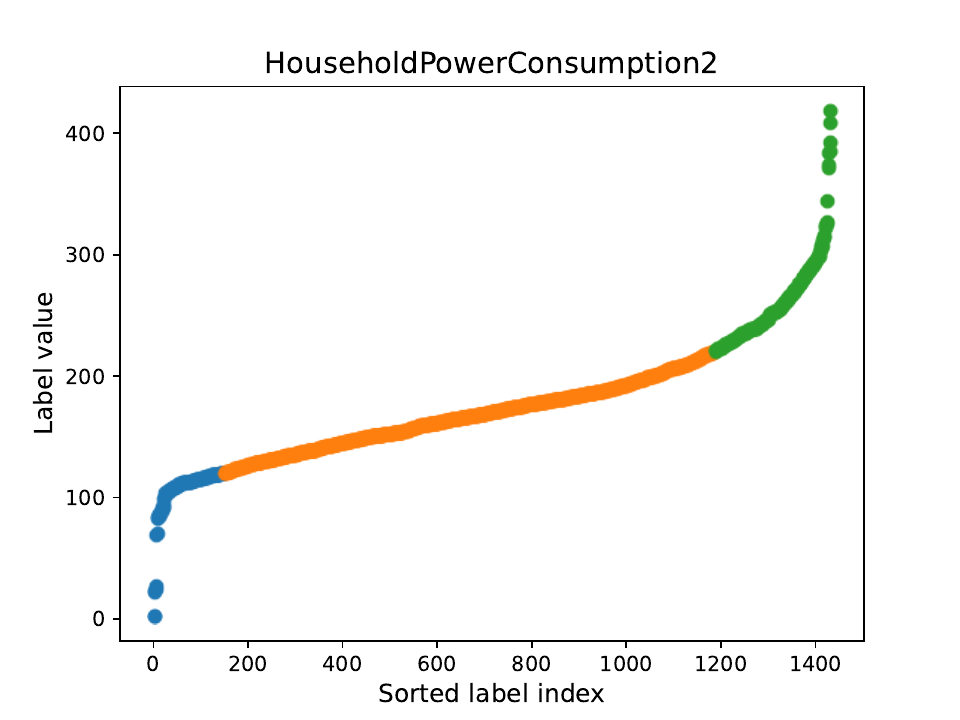} \\
            \includegraphics[width=0.3\linewidth, trim={0cm 0cm 1.3cm 0.5cm},clip]{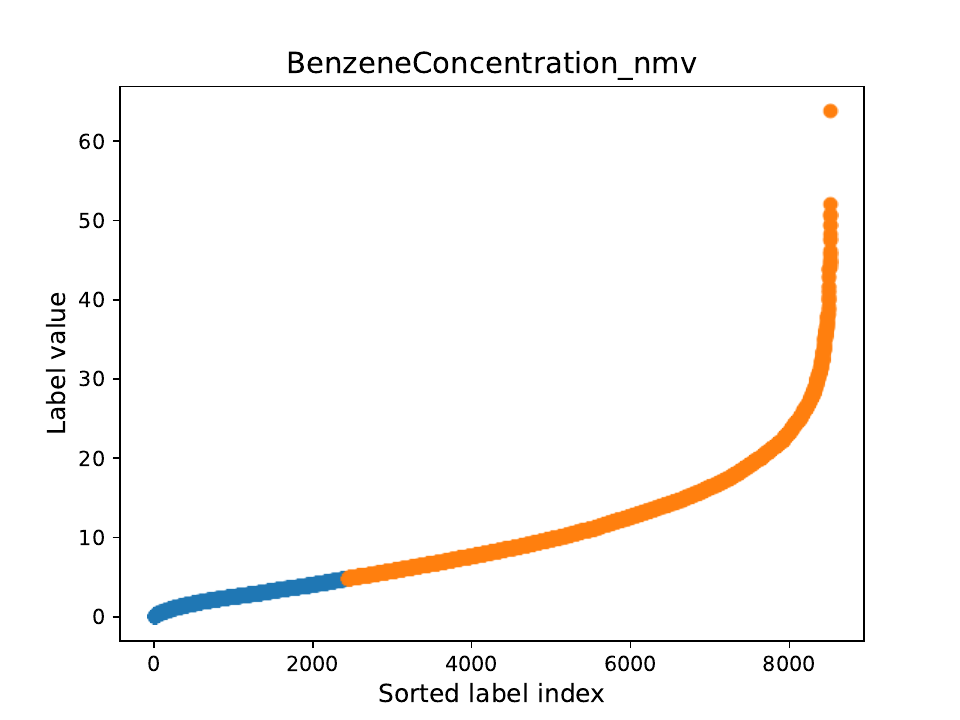} &
            \includegraphics[width=0.3\linewidth, trim={0cm 0cm 1.3cm 0.5cm},clip]{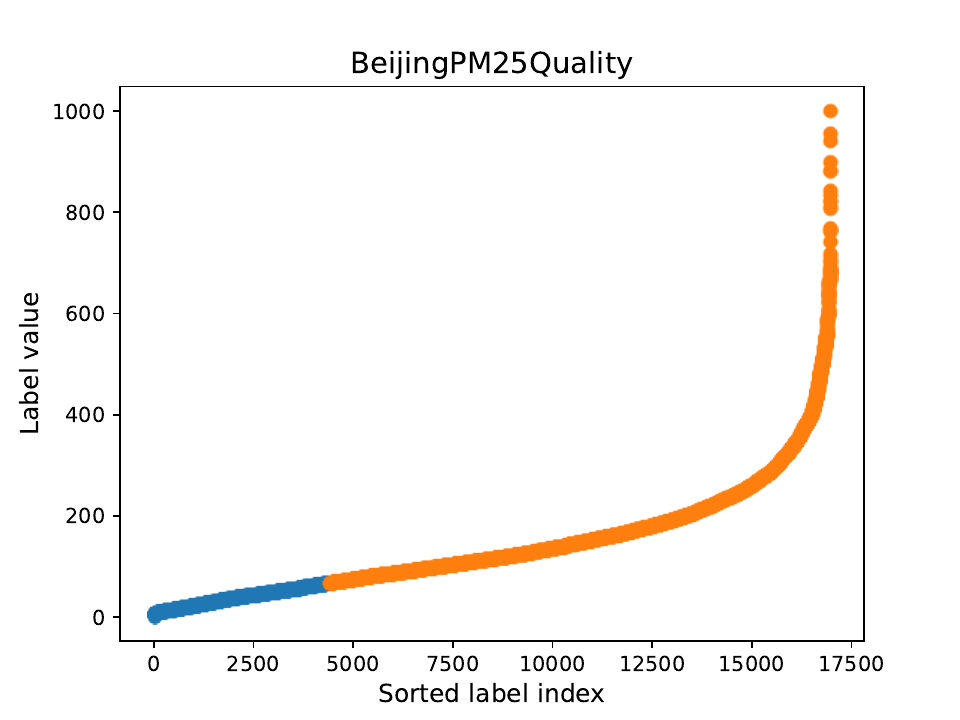} &
            \includegraphics[width=0.3\linewidth, trim={0cm 0cm 1.3cm 0.5cm},clip]{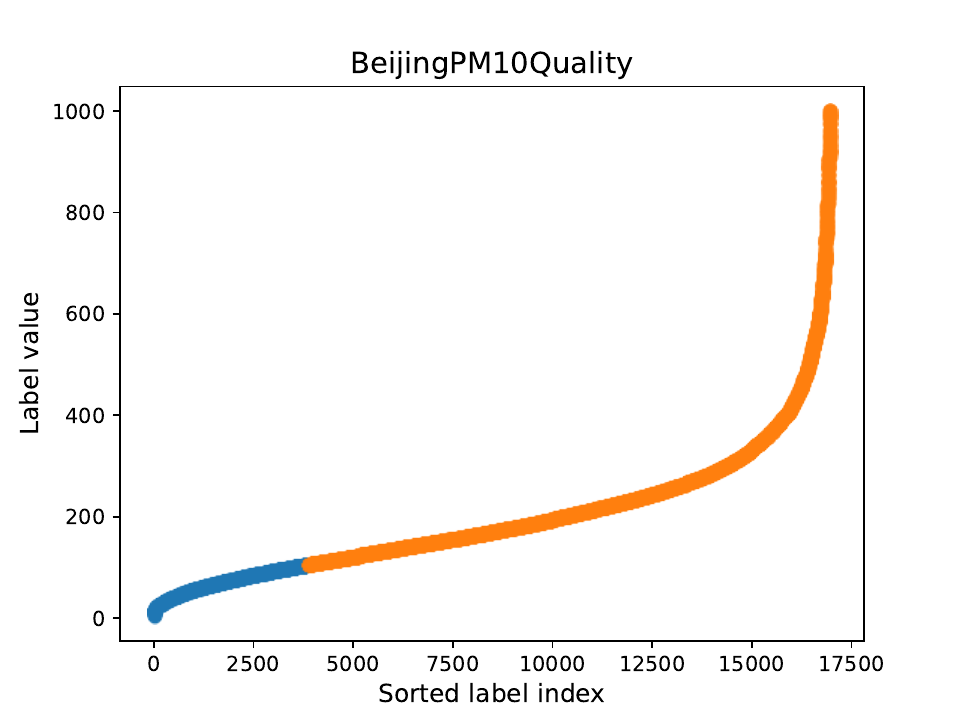} \\
            \includegraphics[width=0.3\linewidth, trim={0cm 0cm 1.3cm 0.5cm},clip]{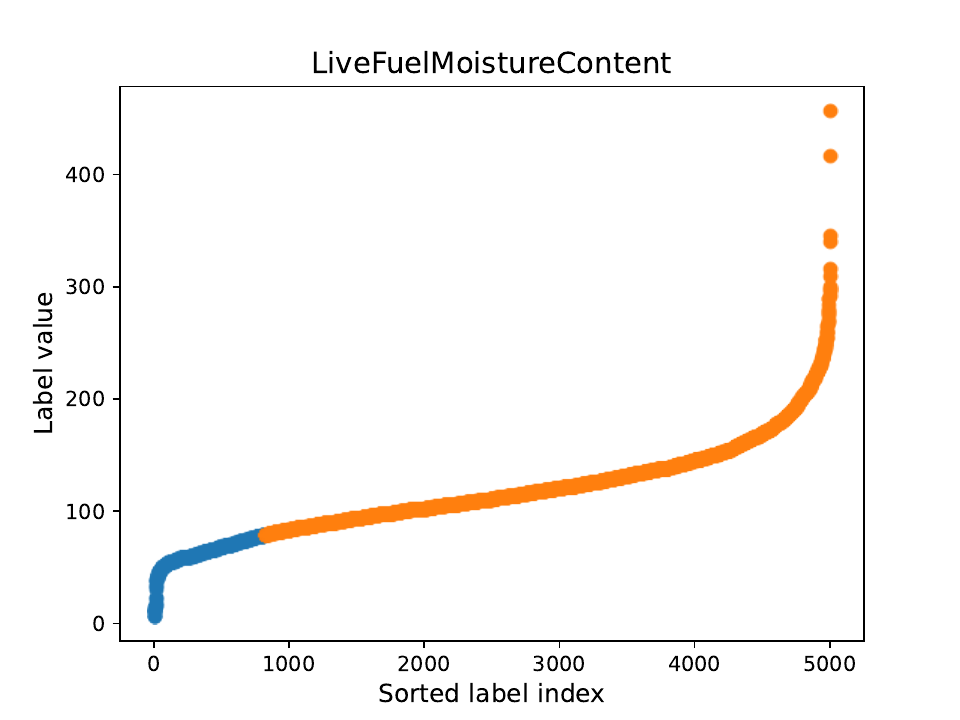} &
            \includegraphics[width=0.3\linewidth, trim={0cm 0cm 1.3cm 0.5cm},clip]{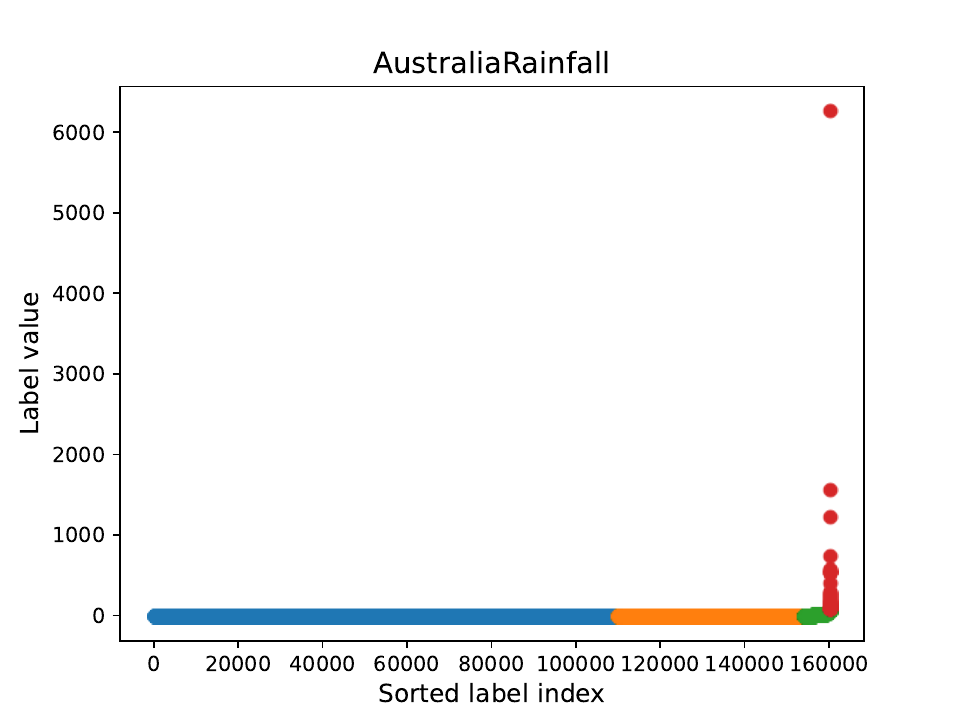} &
            \includegraphics[width=0.3\linewidth, trim={0cm 0cm 1.3cm 0.5cm},clip]{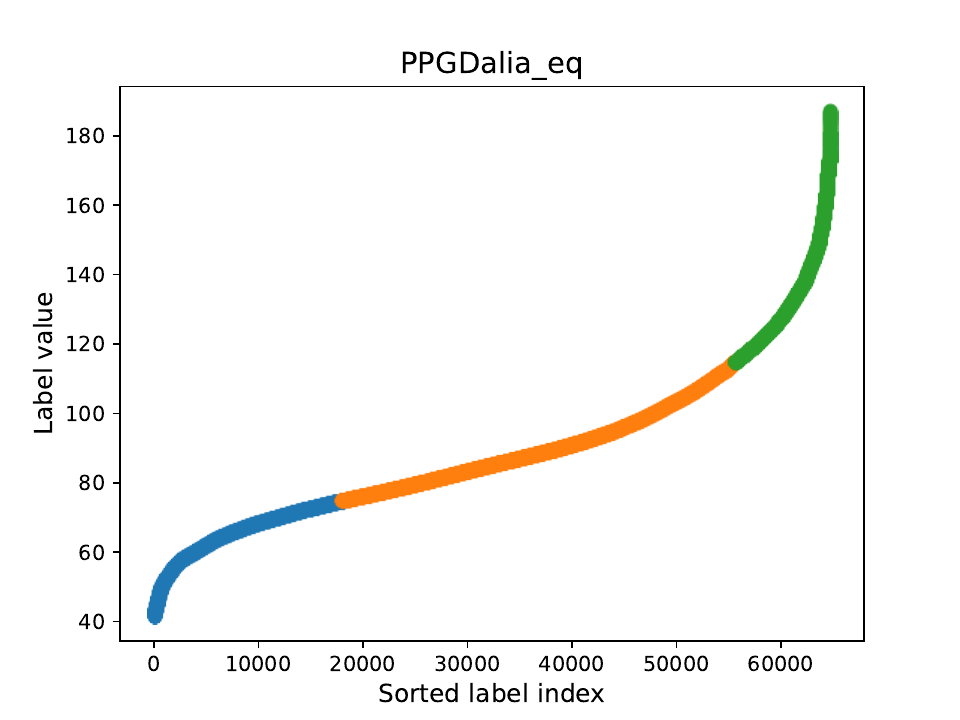} \\
            \includegraphics[width=0.3\linewidth, trim={0cm 0cm 1.3cm 0.5cm},clip]{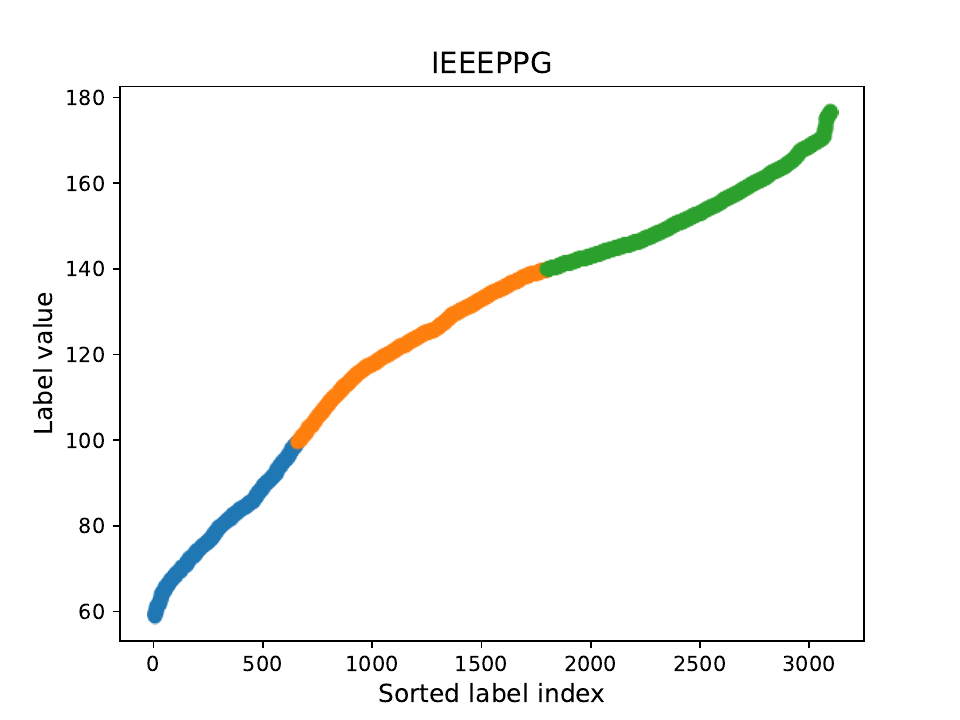} &
            \includegraphics[width=0.3\linewidth, trim={0cm 0cm 1.3cm 0.5cm},clip]{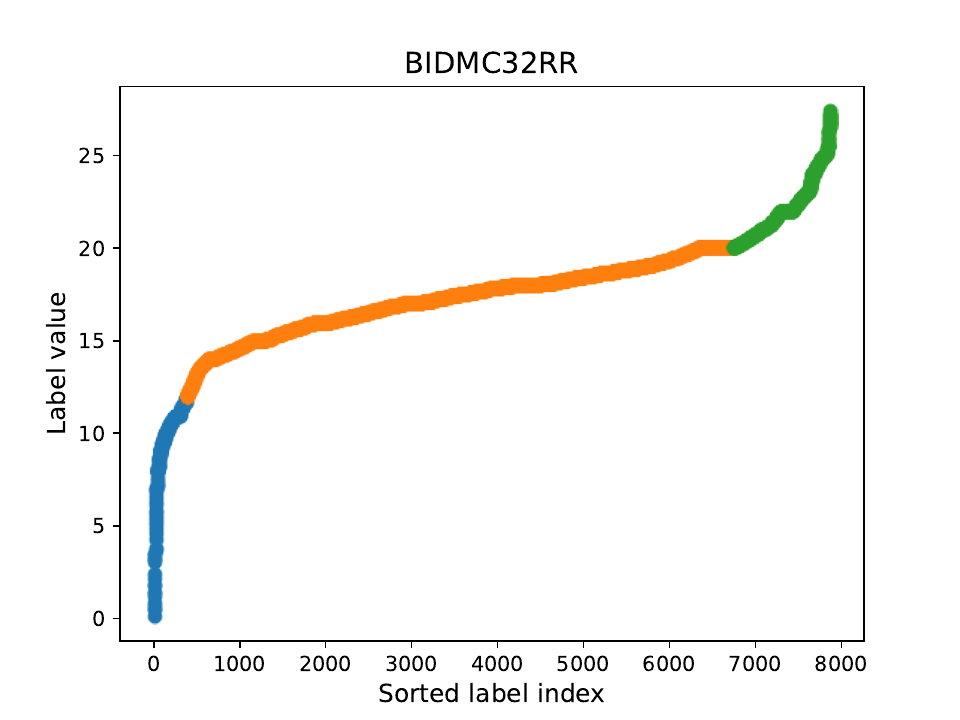} &
            \includegraphics[width=0.3\linewidth, trim={0cm 0cm 1.3cm 0.5cm},clip]{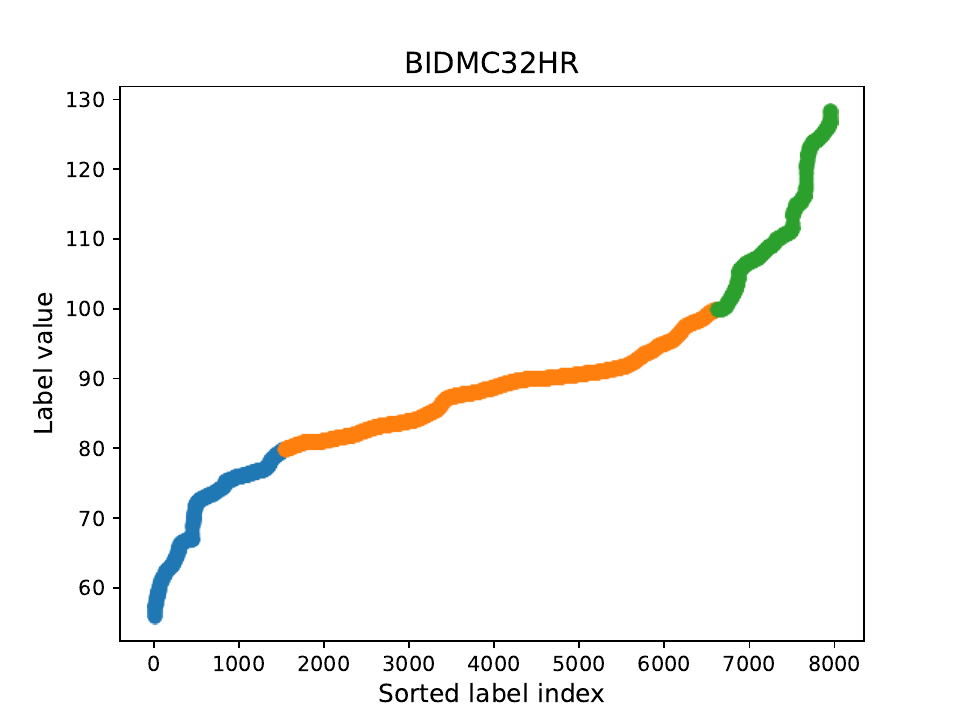} \\
            \includegraphics[width=0.3\linewidth, trim={0cm 0cm 1.3cm 0.5cm},clip]{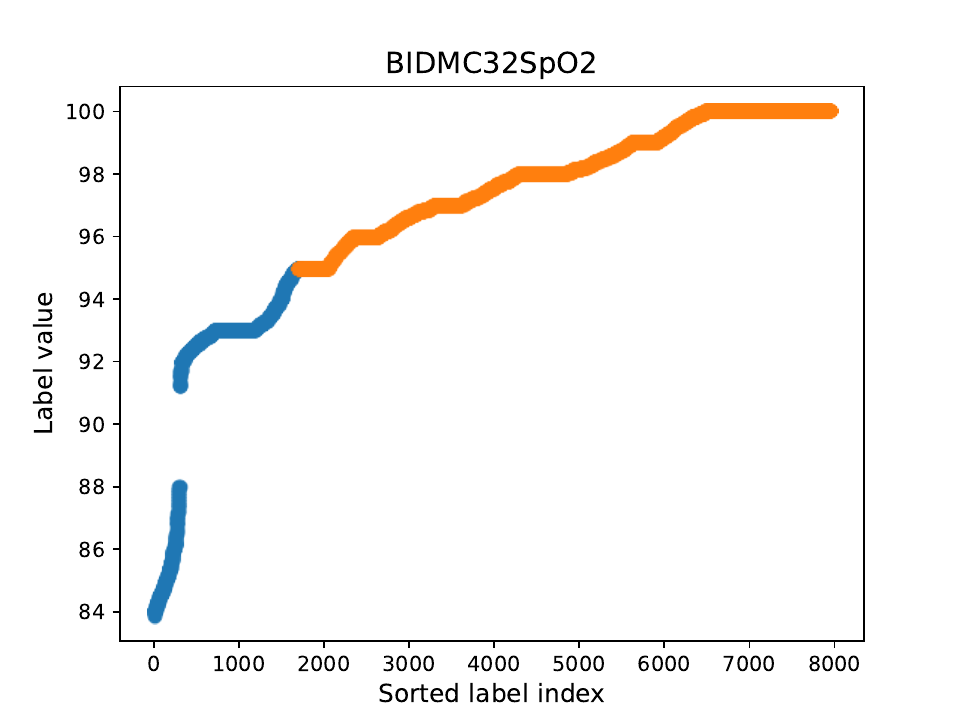} &
            \includegraphics[width=0.3\linewidth, trim={0cm 0cm 1.3cm 0.5cm},clip]{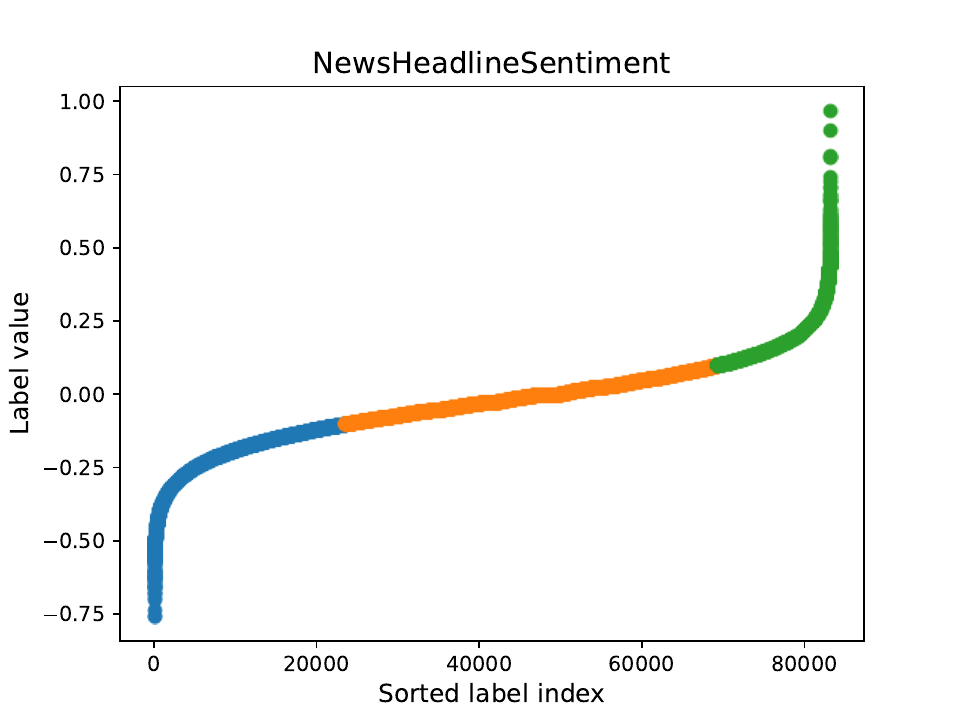} &
            \includegraphics[width=0.3\linewidth, trim={0cm 0cm 1.3cm 0.5cm},clip]{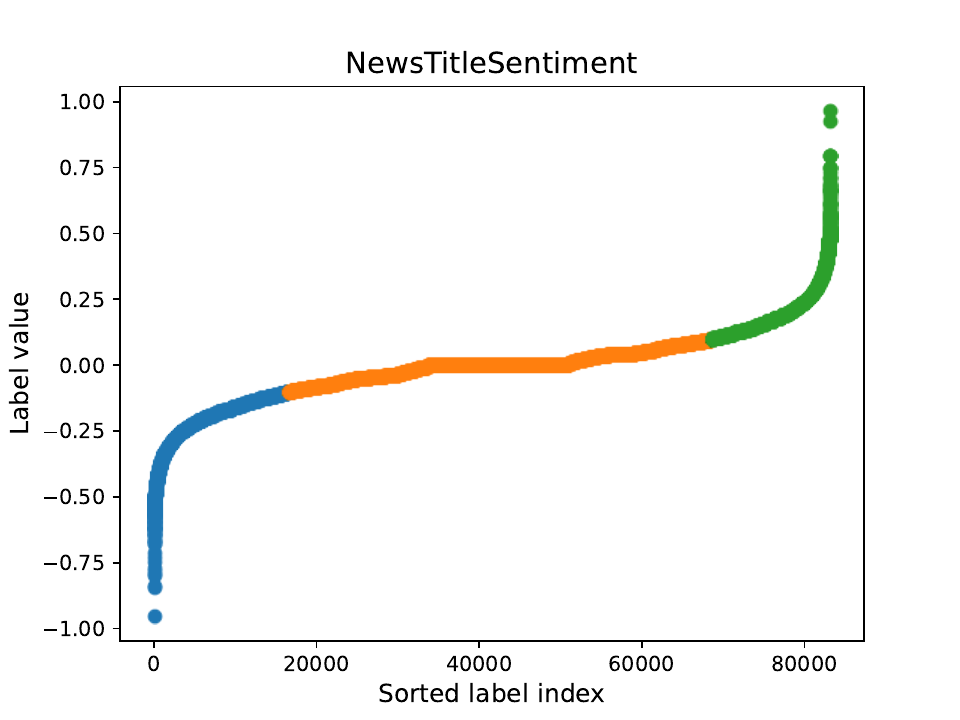} \\
        \end{tabular}
        \caption{The original target labels for the discretised regression problems. The targets for each case have been sorted and plotted in a scatter graph for each problem. The colour of each point represents its post-discretisation class, showing where class splits were made and the resulting class distribution.}
        \label{fig:disc_regression}
    \end{figure}

Table~\ref{tab:disc_regression} gives the class split points used to discretise the original regression targets (Appendix A), while Figure~\ref{fig:disc_regression} visualises the sorted target values together with the resulting class assignments. Where possible, we use practically meaningful thresholds, but for some datasets we choose split points based on the original target distribution while trying to maintain a reasonable class balance. We retain the original regression archive names and append the suffix \textit{\_disc} to denote the discretised classification versions.

The extended TSR archive introduced by~\cite{guijo2024unsupervised} adds many new regression datasets. It expands the archive to a total of 63 regression problems, including the original 19. In this paper, we limit our dataset conversion to the 15 multivariate datasets from the original regression archive. There is therefore further room for expansion using the extended TSR archive should the current batch of discretised datasets prove useful.

\subsection{Standalone Datasets}

In this section, we introduce new datasets to the archive that are not affiliated with any wider collection. There are 11 such datasets in total, which we introduce in subsections categorised by dataset type. Since much of the data collected since the original UEA archive release has been filtered into dedicated archives, the remaining entries are smaller datasets from categories such as non-rehabilitation motion and uncategorised sensor data. These datasets are summarised in Table~\ref{tab:new_datasets}.

    \begin{table}
        \caption{The 11 new standalone MTSC archive datasets. For datasets with unequal length series, both the minimum and maximum series lengths are shown.}
        \label{tab:new_datasets}
        \footnotesize
        \centering
        \begin{tabular}{ | p{5cm} | p{1.15cm} | p{1.15cm} | p{1.3cm} | p{1.25cm} | p{1cm} | p{1.3cm} | }
            \hline
            Dataset name & Train & Test & Channels & Length & Classes & Type \\
            \hline
            AsphaltObstaclesCoordinates & 390 & 391 & 3 & 111-736 & 4 & Motion \\ \hline
            AsphaltPavementTypeCoordinates & 1055 & 1056 & 3 & 66-2371 & 3 & Motion \\ \hline
            AsphaltRegularityCoordinates & 751 & 751 & 3 & 66-4201 & 2 & Motion \\ \hline
            AutomotiveRoadTrials & 153 & 77 & 8 & 300 & 2 & Sensor \\ \hline
            BoneIntensitiesAgeGroup & 600 & 445 & 8 & 270 & 3 & Image \\ \hline
            BoneProbAgeGroup & 600 & 445 & 8 & 256 & 3 & Image \\ \hline
            CounterMovementJump & 419 & 179 & 3 & 2001-7069 & 3 & Motion \\ \hline
            EmoPain & 968 & 355 & 30 & 200 & 3 & Motion \\ \hline
            Locust2022 & 6597 & 3299 & 13 & 96 & 2 & Sensor\\ \hline
            MotionSenseHAR & 966 & 265 & 12 & 1000 & 6 & Motion \\ \hline
            TactileTextureRecognition & 1586 & 681 & 6 & 2000 & 21 & Sensor \\ \hline
        \end{tabular}
    \end{table}

\subsubsection{Image Data}

The \textbf{BoneIntensitiesAgeGroup} and \textbf{BoneProbAgeGroup} datasets were created using data collected as part of a PhD project modelling bone ageing~\cite{davis2013predictive}. Univariate variations extracted from this data exist in the UCR archive, but a multivariate version was not included in the original release. Both datasets contain image outlines of hand bone X-rays. The time series are derived from the same images used in~\cite{cao2000digital}. An algorithm was used to automatically extract the hand outlines and then the outlines of the bones within. Each channel contains the series extracted from one of the phalanges of the little finger, middle finger and thumb. The little finger and middle finger have three bones (the proximal, middle and distal phalanges) while the thumb has two. The datasets differ in the values extracted from these outlines. For BoneIntensitiesAgeGroup each instance is a vector of pixel intensities from a region of interest around the bone, while for BoneProbAgeGroup each instance is the probability distribution of pixel intensities ranging from zero to 255 from the region of interest. Part of the research for which these data were used was to determine whether these outlines could be helpful in bone age prediction. The aim of both datasets is to determine whether the bones are in the following age ranges: up to six years old, seven to twelve, or thirteen to nineteen.

    \begin{figure}[ht]
    	\centering
        \begin{tabular}{ c c }
            \includegraphics[width=.4\linewidth, trim={0cm 0cm 0cm -0.5cm},clip]{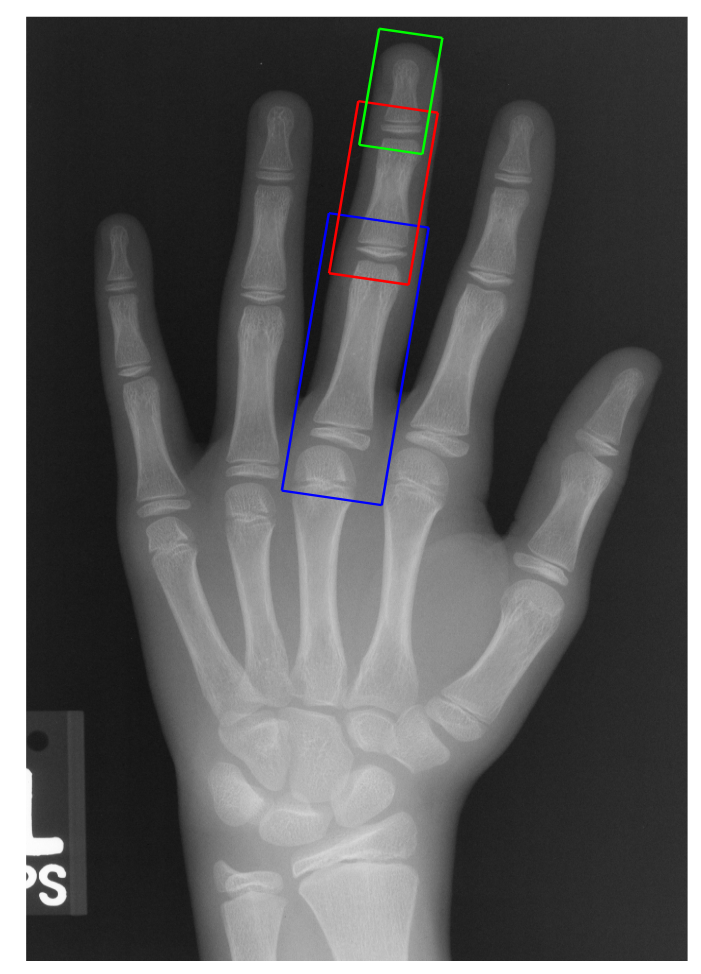} &
            \includegraphics[width=.45\linewidth, trim={0cm 0cm 0cm 0cm},clip]{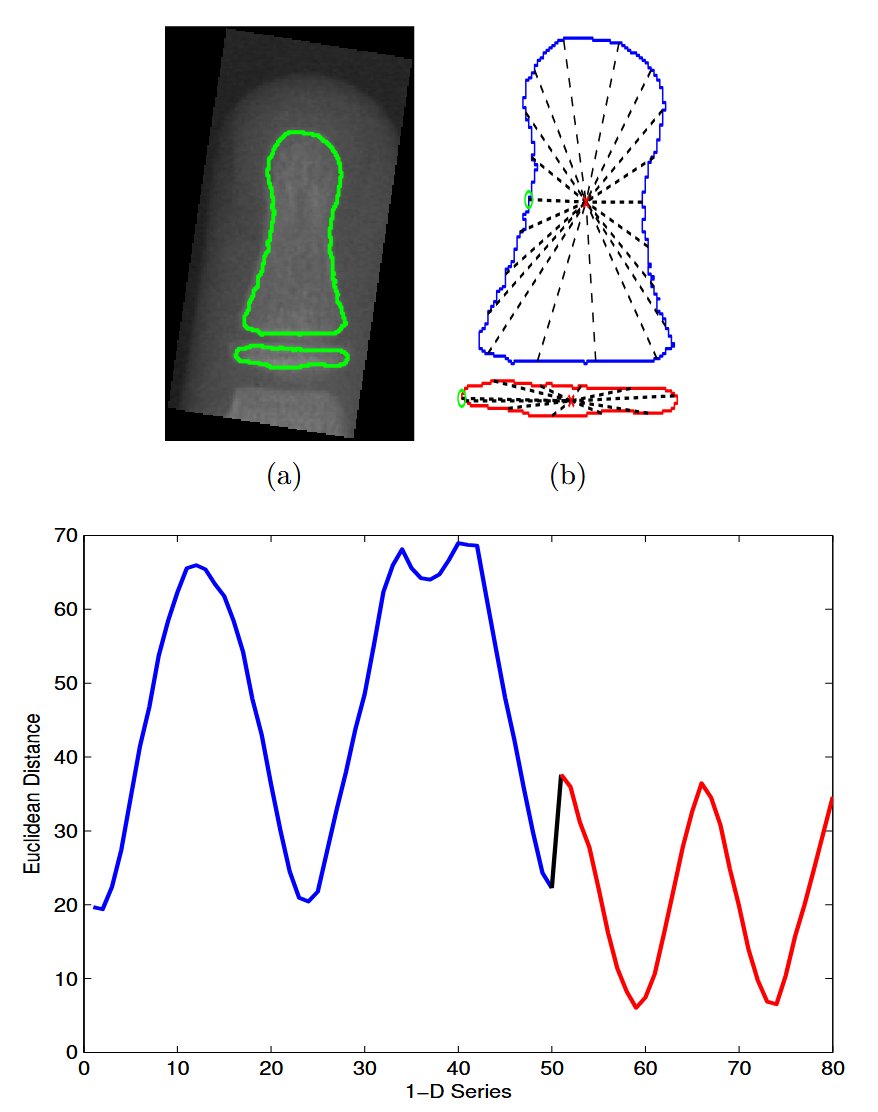} \\
            (A) & (B)
        \end{tabular}
        \caption{An example hand X-ray, with the regions of interest around each of the phalanges of the middle finger (A). An example of a bone segmentation being converted into a univariate series (B). Images from the original PhD thesis~\cite{davis2013predictive}, all credit the author.}
        \label{fig:bones}
    \end{figure}

\subsubsection{Motion Data}

\begin{figure}[!htb]
    	\centering
        \begin{tabular}{ c c }
            \includegraphics[width=.5\linewidth, trim={0cm 0cm 0cm -0.5cm},clip]{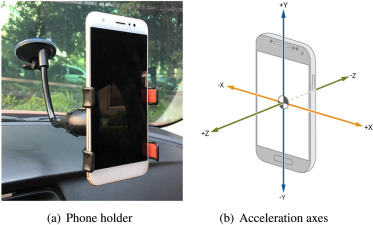} &
            \includegraphics[width=.5\linewidth, trim={0cm 0cm 0cm 0cm},clip]{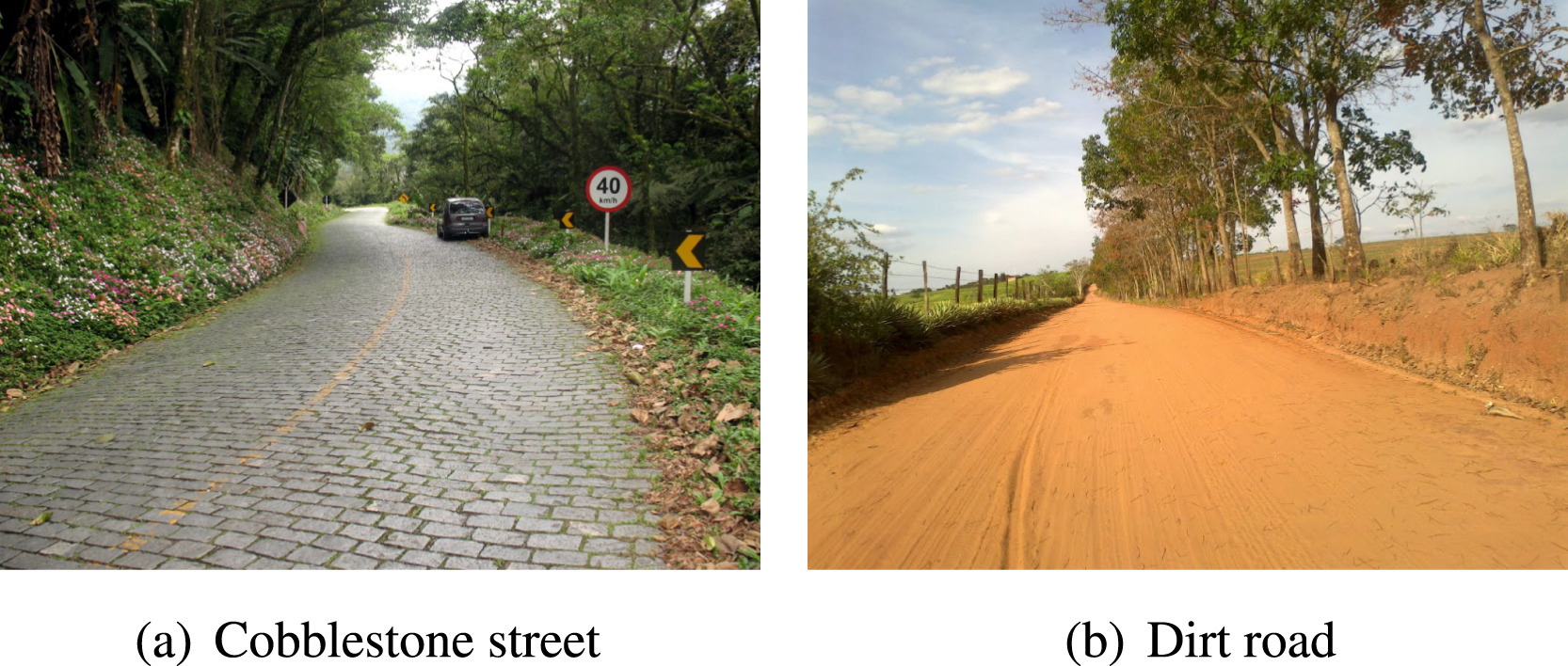} \\
            (A) & (B)
        \end{tabular}
        \caption{Photo examples of the device used to record data (A) and the type of road conditions present in the different street classes from the AsphaltPavementTypeCoordinates dataset (B). This figure is borrowed from Figure 1 and 3 of \cite{souza2018asphalt} which introduced the data. All credit to the original authors.}
        \label{fig:asphalt}
    \end{figure}

The first three motion datasets we introduce are the ``Asphalt'' datasets first presented in \cite{souza2018asphalt}. The aim of these datasets is to predict the condition of a road based on motion data. The data were recorded on a smartphone installed inside a vehicle using a flexible suction holder. This offers the potential for automated monitoring and assessment of road conditions, leading to earlier and less costly interventions with road faults. The Android application \textit{Asfault}~\cite{souza2017towards} was used to collect accelerometer data in the form of the three physical axes, latitude, longitude, and velocity from GPS. A sampling rate of 100 Hz was used for each time series. These datasets cover three separate problems related to road surface conditions. \textbf{AsphaltRegularityCoordinates} looks at road deterioration using driver comfort as a metric. A road is classified as regular or deteriorated. The \textbf{AsphaltPavementTypeCoordinates} problem is to classify the surface type of the road as either dirt, cobblestone or asphalt. The \textbf{AsphaltObstaclesCoordinates} problem is to classify whether the vehicle is crossing one of a set of common road obstacles: speed bumps; vertical patches; raised pavement markers; and raised crosswalks. All three datasets contain unequal length time series. The data recording setup and examples of the types of road being classified for the pavement type problem are shown in Figure~\ref{fig:asphalt}.

\begin{figure}[!htb]
    	\centering
        \begin{tabular}{ c c c }
            \includegraphics[width=.2\linewidth, trim={0cm 0cm 0cm -0.5cm},clip]{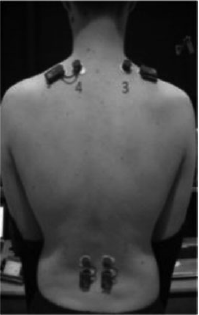} &
            \includegraphics[width=.2\linewidth, trim={0cm 0cm 0cm 0cm},clip]{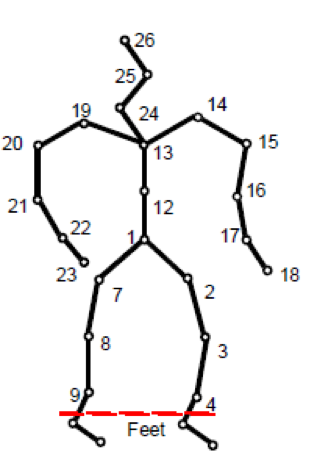} &
            \includegraphics[width=.5\linewidth, trim={0cm 0cm 0cm 0cm},clip]{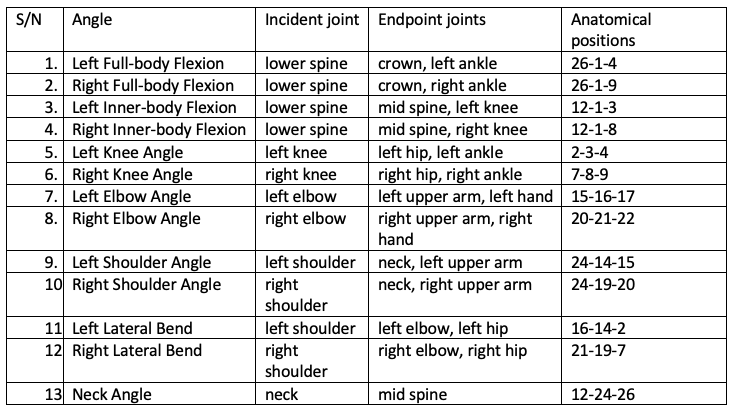} \\
            (A) & (B) & (C)
        \end{tabular}
        \caption{Illustration of the surface electromyography (sEMG) (A) and angle (B and C) sensors. This figure is borrowed from Figure 1 and 2 of the challenge GitHub (\url{https://github.com/Mvrjustid/EmoPainChallenge2020}). All credit to the competition organisers.}
        \label{fig:emopain}
    \end{figure}

The second set of three motion datasets we include in the archive are HAR problems. \textbf{EmoPain} is a dataset from the 2020 EmoPain Challenge~\cite{egede2020emopain} pain recognition from movement task. The goal of the dataset is to detect pain levels in subjects with chronic pain while they perform movements. This is split into three classes: no pain, low pain and high pain. The data are generated from 26 sensors measuring joint angles and four EMG sensors measuring muscle activity. All cases are created from 14 participants with chronic lower back pain and 9 healthy participants, who make up the train and validation set from the challenge. Figure~\ref{fig:emopain} visualises the 30 sensors used to generate the dataset.

\textbf{MotionSenseHAR} records accelerometer and gyroscope data (attitude, gravity, user acceleration, and rotation rate) from 24 participants~\cite{malekzadeh2019mobile}. The participants varied in gender, age, weight, and height, and performed six activities in 15 trials under the same environmental conditions. The data were collected using a mobile phone placed in the participants' front pocket at a sampling rate of 50 Hz. The label to predict is the activity taking place: downstairs, upstairs, walking, jogging, sitting, or standing. The original aim of the dataset was to discover personal attributes of the participants such as gender and weight in addition to the activity to create techniques which can classify while anonymising personal data such as this.

\begin{figure}[!htb]
    	\centering
        \includegraphics[width=.8\linewidth, trim={0cm 0cm 0cm -0.2cm},clip]{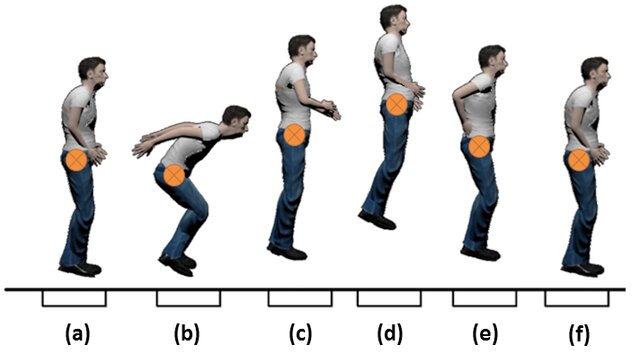}
        \caption{An example of a counter movement jump split into separate stages: standing (a), preparation (b), take-off (c), flight (d), landing (e), and recovery (f). This figure is borrowed from Figure 1 of \cite{hellmers2017understanding} (unrelated to the source of this dataset). All credit to the original authors.}
        \label{fig:cmj}
    \end{figure}

\textbf{CounterMovementJump} consists of X, Y and Z accelerometer data measured with a Shimmer 3 inertial measurement unit (IMU) on each participant's dominant foot. Ten participants were asked to complete the counter movement jump test with acceptable form, with their legs bending during flight, and with a stumble upon landing. This was repeated 20 times by each participant for each scenario, though some cases were lost due to Bluetooth issues. The classification task is to determine which type of jump was performed using the provided accelerometer data. This dataset was first used in \cite{le2019interpretable}. Figure~\ref{fig:cmj} shows the motion performed for a standard counter movement jump.

\subsubsection{Sensor Data}

\textbf{Locust2022} aims to detect locust presence based on environmental variables in the state of Yucatan. The original data was obtained from SENASICA\footnote{\url{https://www.gob.mx/senasica}}, containing the attributes Date, Latitude, Longitude and Result. Result was kept as the class label, while the temporal and spatial information were used to generate time series using the Google Earth Engine\footnote{\url{https://earthengine.google.com/}}. Each case represents an exploration on a specific date and location made by the locust prevention team, inspecting the place to indicate whether they found locusts or not. The results of these inspections are used as the class label for each case. From the date of the exploration, time series data are taken for the previous 96 days for the following environmental variables and used as channels: daytime land surface temperature; normalised difference vegetation index; pressure; specific humidity; rain precipitation rate; air temperature; wind speed; root zone soil moisture; daily total precipitation (sum of all forms converted to water-equivalent); incident shortwave radiation flux density (taken as an average over the daylight period of the day); daily maximum 2-meter air temperature; daily minimum 2-meter air temperature; daily average partial pressure of water vapour.

Donated by the author of \cite{theissler2017detecting}, we include the \textbf{AutomotiveRoadTrials} dataset as a classification task. Originally collected as part of a thesis project~\cite{theissler2013detecting}, the data was used to evaluate anomaly detection for faults in automotive systems. The data were collected from a 2002 Renault Twingo used for the test drives. Data was recorded using the on-board diagnostics interface. The eight recorded signals were: calculated load value, engine coolant temperature, short term fuel trim, intake manifold absolute pressure, engine RPM, vehicle speed, ignition timing advance, and absolute throttle position. As well as recording regular driving data, in some tests, the following faults were injected during the test drive: erroneous injection, erroneous ignition, unavailable engine temperature, and erroneous engine temperature. More information on these faults and how they were induced is available in the original publication. For the AutomotiveRoadTrials dataset in the multivariate archive, the problem is a binary classification task to predict whether a 300 second test drive segment contains any faults or not.

\begin{figure}[!htb]
    	\centering
        \includegraphics[width=1\linewidth, trim={0cm 0cm 0cm -0.4cm},clip]{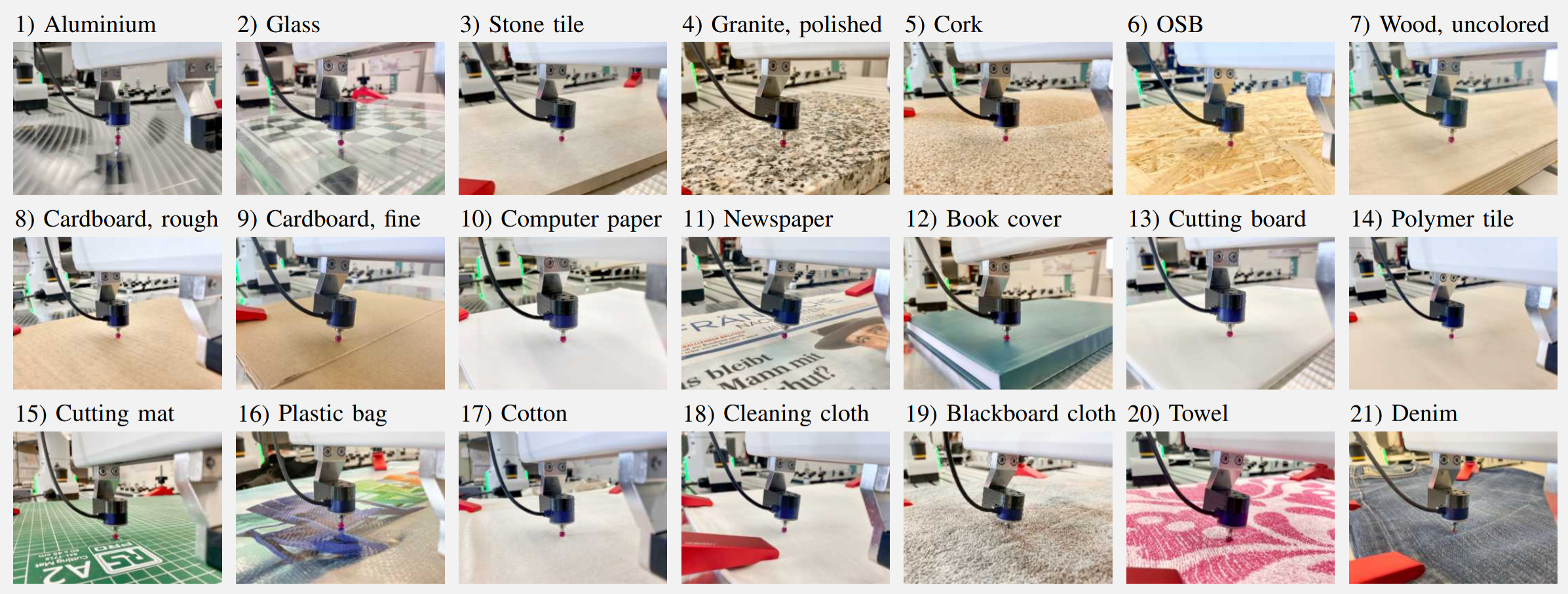}
        \caption{The classes used in TactileTextureRecognition, with visual examples of the robot arm and textures. This figure is borrowed from Figure 5 of \cite{markert2021fingertip}. All credit to the original authors.}
        \label{fig:texture}
    \end{figure}

The \textbf{TactileTextureRecognition} dataset consists of force/torque time series recorded with a Franka Emika Panda robot arm equipped with a miniature Wittenstein HEX21 6-axis sensor and rigid probe fingertip. The data was collected in \cite{markert2021fingertip} and used to create three datasets, each with the series split into 500 millisecond, 1000 millisecond and 2000 millisecond segments. We use the 2000 millisecond version of the data with no preprocessing applied. The goal of the dataset is to determine which material the robot is applying force to. During data acquisition, the robot performed lateral sliding motions over 21 planar surface textures mounted on a flat aluminium plate including metals, stone tiles, wood-based panels, polymers, textiles and paper products. The probe moved at a constant speed of 3 centimetres per second along a path to form an isosceles orthogonal triangle with a 15 cm edge length. The normal force applied by the robot during this movement was set to 4 Newtons. The sensor sampled the three force and three torque components at 1 kHz, producing for each material 30 recordings of 20 seconds. These series are split into multiple 2000 millisecond cases. Each case is associated with one of the 21 texture classes, examples of which are shown in Figure~\ref{fig:texture}.

\subsection{Processed Unequal Length and Missing Value Datasets}

Of the datasets in the archive, five contain missing values and nine contain unequal length time series. While these are common scenarios, and having such data in the archive is useful for evaluating pre-processing methods and algorithms capable of processing such characteristics, not all approaches are built to handle such data. Rather than leave it to individual researchers to select one of the multitude of methods available to pre-process these datasets, we provide versions of each problem with equal length series and no missing values. Table~\ref{tab:cleaned_datasets} summarises these cleaned variants. Datasets ending in ``\_eq'' have been equalised, while those ending in ``\_nmv'' have had missing values removed.

All datasets with missing values come from discretised versions of datasets from the time series regression archive~\cite{guijo2024unsupervised}. The regression archive comes with pre-processed versions of these datasets already, using linear interpolation to fill missing values. We use the same method for the discretised versions. PPGDalia also comes with an equalised variant which is truncated to the shortest channel length. For CounterMovementJump we truncate parts with no activity from both the beginning and end of the series down to a length of 3000. If the series is already less than 3000 in length, we pad the series start with noise matching values found when there is no activity. For the remaining unequal length datasets, we equalise via FFT-based resampling applied independently to each channel, adjusting the series to a common target length by spectral zero-padding when upsampling, and truncation when downsampling.

    \begin{table}
        \caption{The 14 cleaned versions of datasets from the MTSC archive with unequal length series or missing values. This includes \textit{\_eq} datasets which are variants with equal length series and \textit{\_nmv} datasets which are variants with no missing values. HouseholdPowerConsumption is compressed to HPC for formatting.}
        \label{tab:cleaned_datasets}
        \footnotesize
        \centering
        \begin{tabular}{ | p{5.4cm} | p{1.05cm} | p{1.05cm} | p{1.2cm} | p{1.2cm} | p{1cm} | p{1.25cm} | }
            \hline
            Dataset name & Train & Test & Channels & Length & Classes & Type \\
            \hline
            AsphaltObstaclesCoordinates\_eq & 390 & 391 & 3 & 298 & 4 & Motion \\ \hline
            AsphaltPavementTypeCoordinates\_eq & 1055 & 1056 & 3 & 400 & 3 & Motion \\ \hline
            AsphaltRegularityCoordinates\_eq & 751 & 751 & 3 & 384 & 2 & Motion \\ \hline
            BeijingPM10Quality\_disc\_nmv & 11918 & 5048 & 9 & 24 & 2 & Sensor \\ \hline
            BeijingPM25Quality\_disc\_nmv & 11918 & 5048 & 9 & 24 & 2 & Sensor \\ \hline
            BenzeneConcentration\_disc\_nmv & 3349 & 5163 & 8 & 240 & 2 & Sensor \\ \hline
            CharacterTrajectories\_eq & 1422 & 1436 & 3 & 119 & 20 & Motion \\ \hline
            CounterMovementJump\_eq & 419 & 179 & 3 & 3000 & 3 & Motion \\ \hline
            HPC1\_disc\_nmv & 745 & 686 & 5 & 1440 & 3 & Power \\ \hline
            HPC2\_disc\_nmv & 745 & 686 & 5 & 1440 & 3 & Power \\ \hline
            InsectWingbeat\_eq & 25000 & 25000 & 200 & 20 & 10 & Audio \\ \hline
            JapaneseVowels\_eq & 270 & 370 & 12 & 25 & 9 & Audio \\ \hline
            PPGDalia\_disc\_eq & 43215 & 21482 & 4 & 256 & 3 & Biosignal \\ \hline
            SpokenArabicDigits\_eq & 6599 & 2199 & 13 & 65 & 10 & Audio \\ \hline
        \end{tabular}
    \end{table}

\section{Multiverse-core: A Practical Archive Subset}
\label{sec:subset}

In this section, we propose a subset of the full Multiverse archive for initial algorithmic development. Our aim with this set is to produce a leaner collection of datasets which researchers can reasonably use to benchmark approaches they are developing without the massive computing resource requirements the full archive necessitates. We aim to balance the datasets selected by excluding datasets from overrepresented problem types such as rehabilitation and EEG, issues with the full Multiverse archive we have discussed previously. We name this subset Multiverse-core (MV-core). We only consider dataset versions with equal length and no missing values. Our subset contains 66 datasets total, which is half the size of the full Multiverse archive and contains 40 more datasets than the current equal length UEA archive. In the remainder of this section we provide justification for the datasets we have selected for and excluded from this recommended subset.

The first set of problems we exclude from our subset are due to size. We define a metric based on the amount of resources required to store a dataset in memory, and use this as a basis to gauge how large the dataset is comparatively to others.
\begin{equation*}
\text{Size (GB)} \approx \frac{(n_{\mathrm{train}}+n_{\mathrm{test}}) \cdot d \cdot m \cdot 32}{8 \cdot 10^{9}}
\end{equation*}
We do not include any dataset with an approximated memory cost greater than 0.5 GB in the subset. This excludes the following 22 datasets: \textbf{InsectWingbeat} from the original archive. \textbf{SitStand}, \textbf{SongFamiliarity}, \textbf{MatchingPennies}, \textbf{FeetHands}, \textbf{ImaginedOpenCloseFist}, \textbf{OpenCloseFist}, \textbf{ImaginedFeetHands}, \textbf{InnerSpeech}, \textbf{PronouncedSpeech}, \textbf{VisualSpeech}, \textbf{ShortIntervalTask}, and \textbf{LongIntervalTask} from the EEG archive. \textbf{Opportunity}, \textbf{PAMAP2}, \textbf{DREAMERA}, \textbf{DREAMERV}, \textbf{S2Agri-10pc-17}, and \textbf{S2Agri-10pc-34}, \textbf{LenDB}, \textbf{S2Agri-17}, \textbf{S2Agri-34} from the {\monster} archive.

The addition of the 39 Rehab-Pile archive datasets creates a large bias towards skeleton-based motion data and rehabilitation exercises. It could be argued the original 30 dataset archive already has a disproportionate amount of HAR datasets included, and we wish to avoid further extending that. From the rehabilitation archive we exclude 33 datasets from the subset, retaining six of the new rehabilitation datasets. The selected datasets are: \textbf{IRDS-SFL}, \textbf{KERAAL-RTK}, \textbf{KIMORE-PR-C}, \textbf{KINECAL-QSEO}, \textbf{UCDHE-Rowing-MC}, and \textbf{UIPRMD-DS-C}. We aimed to include datasets from multiple studies and a variety of exercise types. Datasets which appear to be trivial or contained oddities in their train-test split such as the multiclass KERAAL problems were excluded.

We exclude five datasets where it is dubious whether any information can be extracted from the data. For \textbf{BIDMC32RR\_disc}, \textbf{LiveFuelMoistureContent\_disc}, \textbf{NewsHeadlineSentiment\_disc}, and \textbf{NewsTitleSentiment\_disc} selecting the majority class is the best approach for the wide range of algorithms tested. In the original archive, \textbf{SelfRegulationSCP2} was introduced with the warning {\em ``Note that it is not clear if there is any information contained in this dataset that is useful for the classification task. A view on the result suggests that it is not. The best has error 45.5\%.''}~\cite{bagnall18mtsc}. With the archive containing another similar problem as well as this warning, we choose to exclude it from the subset.

Finally, we handpick seven additional biosignal and motion datasets for exclusion, as these types of problem remain overrepresented. \textbf{PPGDalia\_disc} contains different length channels, and has a large amount of data truncated to equalise it. \textbf{FingerMovements}, \textbf{FeedbackButton} and \textbf{ButtonPress} are EEG datasets with overlap in problem type to other archive datasets. For motion data, \textbf{AsphaltRegularityCoordinates} is similar to the other asphalt datasets which have been introduced. Likewise, \textbf{WISDM2} overlaps with the other WISDM dataset in the archive. \textbf{BasicMotions} alongside being a HAR problem is trivial for the vast majority of algorithms.

Across the different archive subsections introduced above, the Multiverse-core subset includes 26 of the 30 original UEA datasets; 6 of the 20 EEG archive datasets (10 of 26 including shared data from the original archive); 8 of the 18 {\monster} datasets; 6 of the 39 Rehab-Pile datasets; 10 of the 15 discretised regression datasets; and 10 of 11 new standalone multivariate problems. Table~\ref{tab:core_stats} provides summary statistics for the core subset. As would be expected given our subset goals, there are proportionally fewer datasets with extreme sizes across all categories, though these characteristics are still present. Compared to the full archive the subset has fewer binary classification problems, largely due to the removal of many rehabilitation problems which aimed to classify correct and incorrect exercise motions. While there is still a skew towards the motion and biosignal problem types we defined, this has been reduced from the full archive with a more diverse set of problem types and methods of data generation within the categories.

    \begin{table}[t]
        \caption{Summary statistics for the 66 Multiverse-core datasets. Provides statistics for the train set sizes (a), test set sizes (b), number of classes (c), number of time points (d), number of dimensions (e) of the archive datasets, and dataset type (f).}
        \label{tab:core_stats}
        \footnotesize
        \centering

        \begin{tabular}{c c c}
    		\begin{tabular}{l|c}
    			\hline
                $\leq$200 & 21 (31.82\%) \\
                201-2000 & 25 (37.88\%) \\
                2001-20000 & 15 (22.73\%) \\
                $>$20000 & 5 (7.58\%) \\
                Minimum & 12  \\
                Maximum & 112,186 \\  \hline
    		\end{tabular}
            &
    		\begin{tabular}{l|c}
    			\hline
                $\leq$200 & 23 (34.85\%) \\
                201-2000 & 23 (34.85\%) \\
                2001-20000 & 19 (28.79\%) \\
                $>$20000 & 1 (1.52\%) \\
                Minimum & 7 \\
                Maximum & 48,081 \\  \hline
    		\end{tabular}
    		&
            \begin{tabular}{l|c}
    			\hline
                2 & 23 (34.85\%) \\
                3-5 & 22 (33.33\%) \\
                6-15 & 16 (24.24\%) \\
                $>$15 & 5 (7.58\%) \\
                Minimum & 2 \\
                Maximum & 39  \\  \hline
    		\end{tabular} \\[1cm]
    		(a) Train Size & (b) Test Size & (c) No. Classes \\[0.25cm]
	
        	\begin{tabular}{l|c}
        		\hline
                $\leq$50 & 10 (15.15\%) \\
                51-500 & 36 (54.55\%) \\
                501-5000 & 17 (25.76\%) \\
                $>$5000 & 3 (4.55\%) \\
                Minimum & 8 \\
                Maximum & 17,984 \\  \hline
        	\end{tabular}
            &
            \begin{tabular}{l|c}
        		\hline
                $\leq$5 & 21 (31.82\%) \\
                6-20 & 29 (43.94\%) \\
                21-100 & 12 (18.18\%) \\
                $>$100 & 4 (6.06\%) \\
                Minimum & 2  \\
                Maximum & 1345  \\  \hline
        	\end{tabular}
            &
            \begin{tabular}{l|c}
        		\hline
                Audio & 5 (7.58\%) \\
                Biosignal & 17 (25.75\%) \\
                Motion & 27 (40.91\%) \\
                Image & 3 (4.55\%) \\
                Power & 3 (4.55\%) \\
                Sensor & 10 (15.15\%) \\
                Spectro & 1 (1.52\%) \\ \hline
        	\end{tabular} \\[1.25cm] 	
            (d) No. Time points & (e) No. Channels & (f) Dataset Types
	    \end{tabular}
	\end{table}

\section{Experiments and Results}
\label{sec:results}

To showcase the Multiverse archive, we provide a practical and reproducible first benchmark. Our aim is not to conduct a new exhaustive bake off, but rather to provide an initial practical and reproducible benchmark for the Multiverse archive. We therefore include multivariate-capable classifiers that cover the main methodological families identified in earlier comparative studies \cite{ruiz21mtsc,middlehurst24bakeoff}, while restricting attention to methods implemented in \textit{aeon} \cite{aeon24jmlr}. This ensures broad coverage of established MTSC design patterns within a common software framework. Methods not currently available in \textit{aeon}, or known to have substantial runtime or memory demands, are left for future work. The results should therefore be interpreted as a representative baseline, rather than as a definitive comparison of all current MTSC algorithms.

We run a total of 17 classifiers on 133 MTSC archive datasets. Where relevant, we use dataset variants with equal length series and no missing values. The experiments are limited to a single run on the default train-test split, with no resampling or repeated runs with different random seeds. To support reproducibility, we set the random seed for all relevant algorithms to 0. Each classifier-dataset run is limited to 3 days of runtime on a single CPU. Although most experiments did not require this, runs that initially failed were allowed up to 200 GB of memory and 40 CPU threads, where supported by the implementation. No further attempts were made for algorithms that could not process a dataset within these resource limits or that failed for other reasons. This is not a comprehensive evaluation of algorithms on the archive, but rather an initial validation and a demonstration of how current multivariate-capable algorithms perform on the expanded archive.

We compare algorithms using a variation of the critical difference diagram~\cite{Demsar2006comparisons}. Pairwise Wilcoxon signed-rank tests replace the post-hoc Nemenyi test when comparing approaches. Cliques are formed using the Holm correction as recommended by~\cite{garcia2008extension,benavoli2016should}. We also use the multi-comparison matrix (MCM) described in \cite{ismail23multiple}. We compare algorithms using accuracy as the performance metric. Our accompanying website contains guidance on downloading and reproducing all experiments, in addition to all results we have generated to date\footnote{\url{https://github.com/aeon-toolkit/multiverse}}.

\begin{figure}[t]
    	\centering
        \includegraphics[width=1\linewidth, trim={0cm 0cm 0cm 0cm},clip]{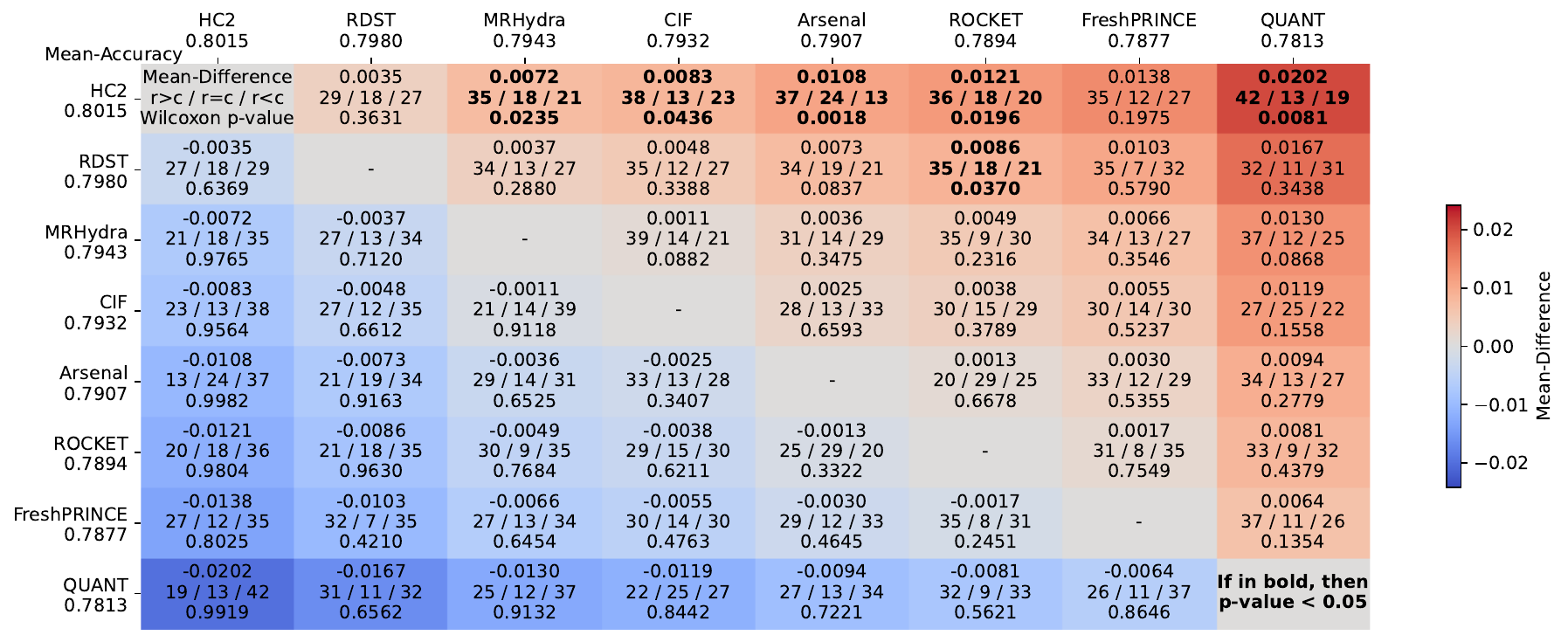}
        \caption{An MCM diagram for the top-performing multivariate-capable classifiers on 74 Multiverse datasets.}
        \label{fig:132mcm}
    \end{figure}

\begin{figure}[t]
    	\centering
        \begin{tabular}{ c c }
            \includegraphics[width=.5\linewidth, trim={0cm 0cm 0cm 0cm},clip]{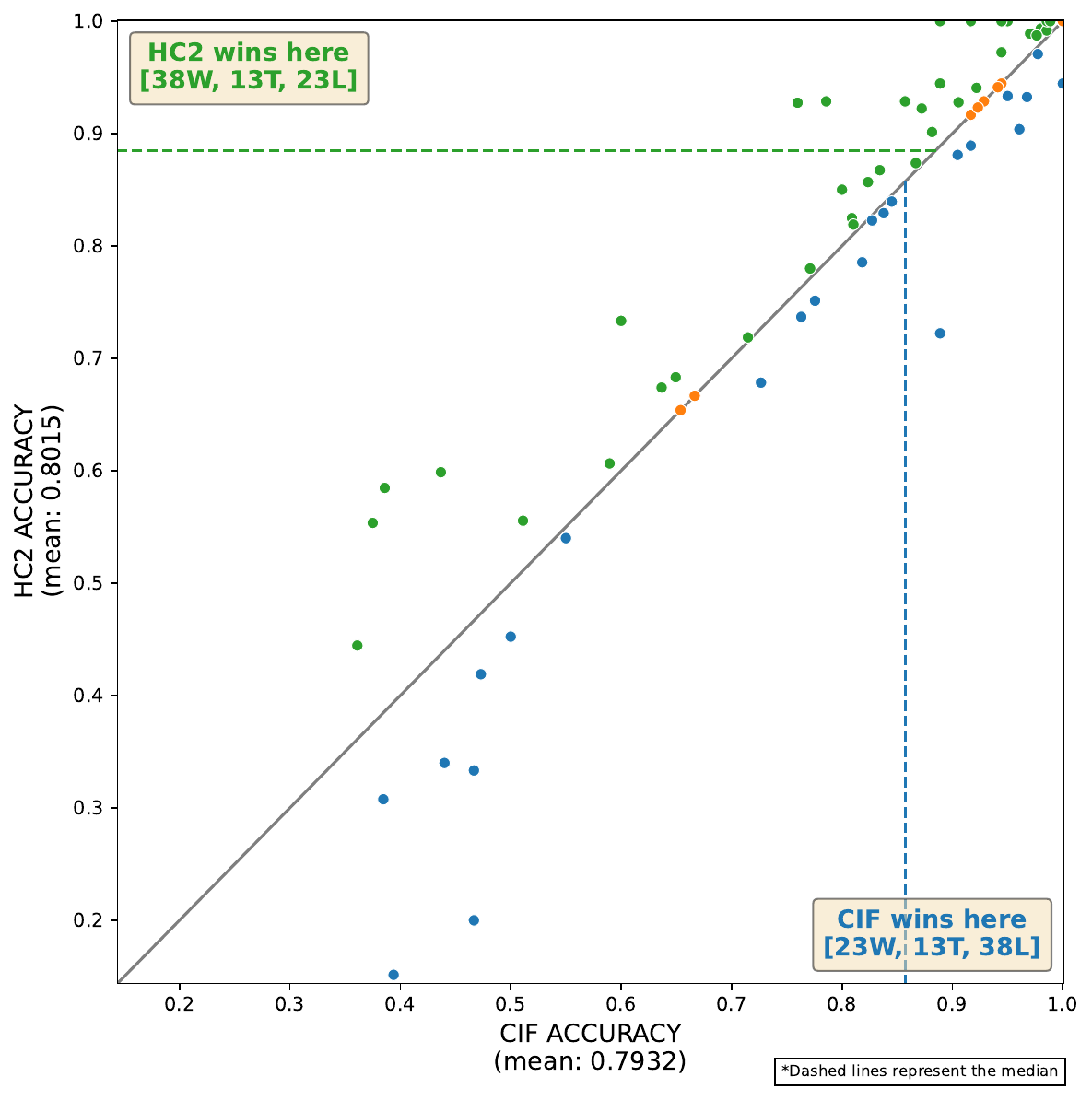} &
            \includegraphics[width=.5\linewidth, trim={0cm 0cm 0cm 0cm},clip]{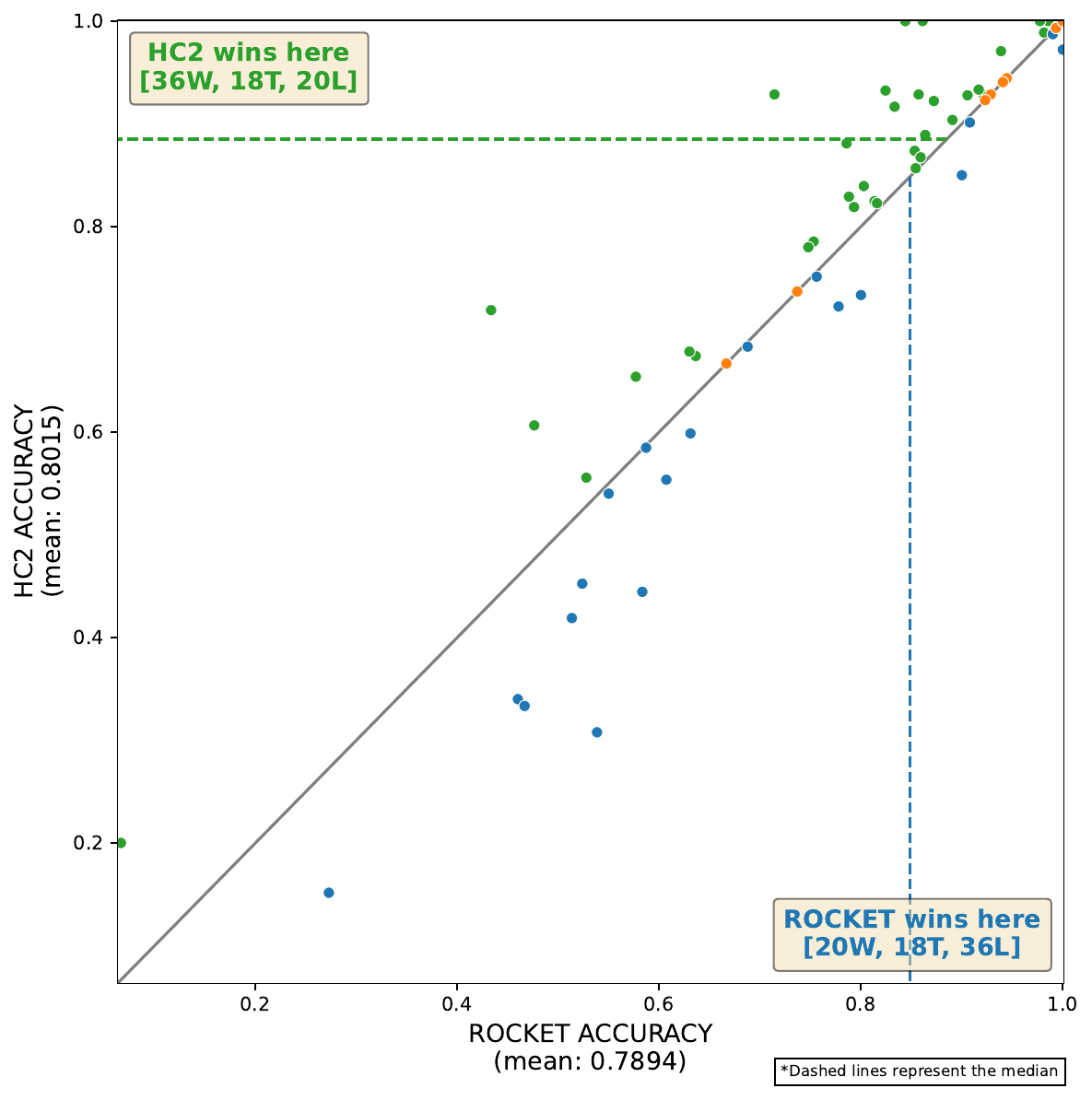} \\
            (A) & (B)
        \end{tabular}
        \caption{A scatter diagram comparing HC2 to CIF (A) and HC2 to ROCKET (B) on 74 Multiverse datasets.}
        \label{fig:scatter}
    \end{figure}

Of the 133 datasets, only 74 have complete results for every algorithm. The vast majority of these failures are due to runtime and memory constraints. For S2Agri-17 and S2Agri-34, the largest datasets available, only a dummy classifier predicting the majority class could produce results. Loading these datasets alone has a considerable memory cost. For completeness, we provide full results tables containing the accuracy on all datasets where results were produced in Appendix~\ref{app:results}. Figure~\ref{fig:132cd} presents a critical difference diagram for our algorithms on the 74 fully completed datasets. HC2 ranks highest, but is not significantly different from multiple algorithms. For the top eight algorithms, we produce an MCM diagram shown in Figure~\ref{fig:132mcm}. No significant differences are observed among these algorithms, except that HC2 is significantly better than MRHydra, Arsenal, ROCKET, and QUANT. Figure~\ref{fig:scatter} shows scatter diagrams comparing accuracy on individual datasets for HC2 against CIF and ROCKET, two top performers from the MTSC bake off.

    \begin{figure}[t]
    	\centering
        \includegraphics[width=.95\linewidth, trim={0cm 0cm 0cm 0cm},clip]{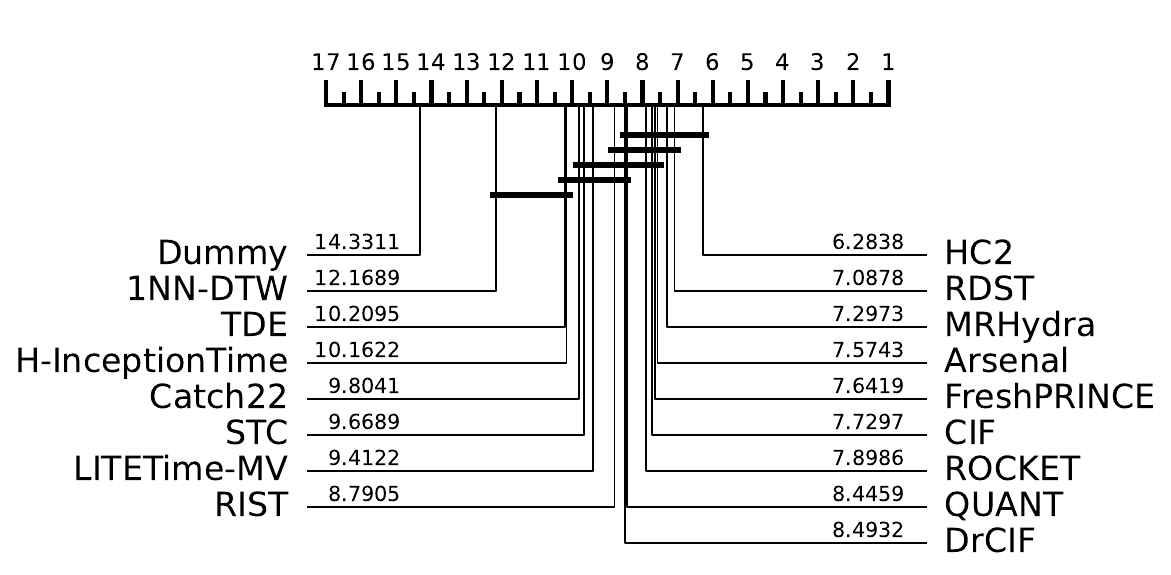}
        \caption{An accuracy rank critical difference diagram for multivariate-capable classifiers on 74 Multiverse datasets.}
        \label{fig:132cd}
    \end{figure}

\begin{figure}[!h]
    	\centering
        \includegraphics[width=.95\linewidth, trim={0cm 0cm 0cm 0cm},clip]{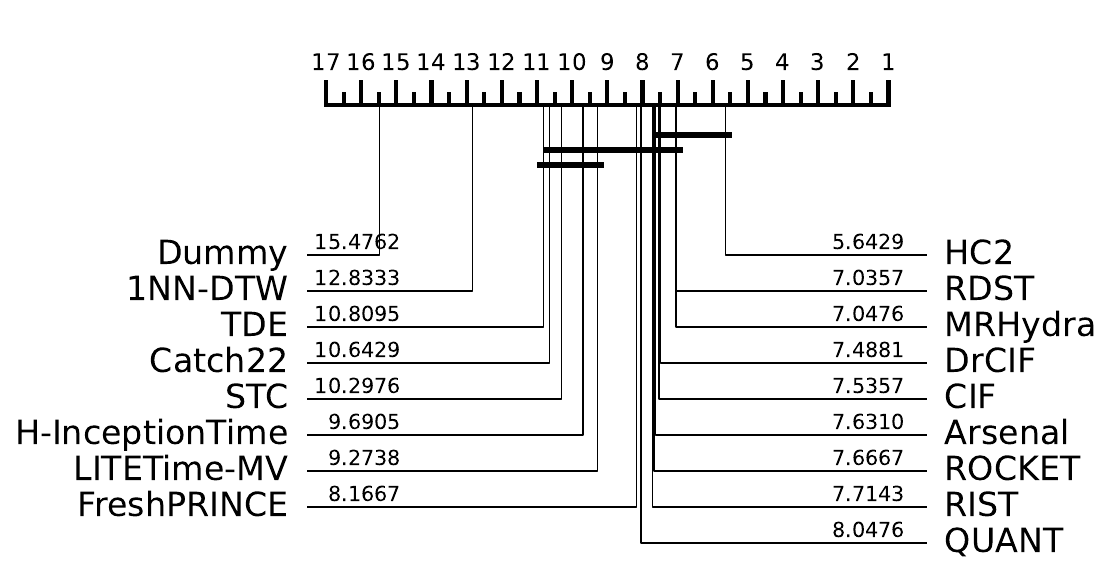}
        \caption{An accuracy rank critical difference diagram for multivariate-capable classifiers on 42 Multiverse-core datasets.}
        \label{fig:66cd}
    \end{figure}

\begin{figure}[ht]
    	\centering
        \includegraphics[width=.95\linewidth, trim={0cm 0.6cm 0cm 0cm},clip]{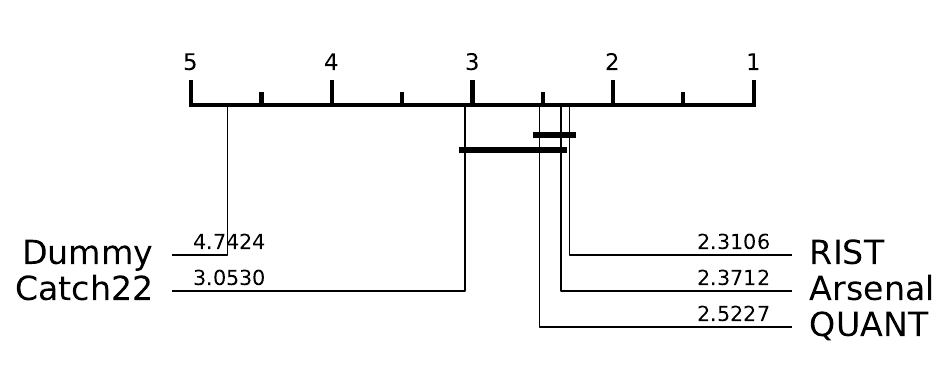}
        \caption{An accuracy rank critical difference diagram for algorithms that completed all 66 Multiverse-core datasets.}
        \label{fig:66allcd}
    \end{figure}

We next present results for the Multiverse-core subset. For this collection, 42 of the 66 datasets have complete results for every algorithm. The main difference compared with the previous results is the removal of many skeletal rehabilitation problems, which are relatively small and straightforward for most algorithms to complete. Figure~\ref{fig:66cd} shows the results for all algorithms in a critical difference diagram. Although the relative positions of many algorithms are similar, interval-based approaches and hybrids using interval features appear to achieve slightly higher average ranks. Five algorithms completed all 66 datasets under our resource constraints, and the corresponding critical difference diagram is shown in Figure~\ref{fig:66allcd}. All of these results are available on the associated webpage\footnote{\url{https://github.com/aeon-toolkit/multiverse}} which we will continue to maintain and update as new results are generated and new classifiers are proposed.

\section{Conclusion}
\label{sec:conclusion}

We have presented the Multiverse archive, a substantial expansion of the original UEA multivariate time series classification archive. The new archive increases the number of unique classification problems from 30 to 133, and to 147 released datasets when preprocessed variants for missing values and unequal length series are included. Beyond scale alone, the archive consolidates datasets from multiple sources into a single repository, rebrands the collection to reflect broader community ownership, and provides a common framework for access, reproducibility, and result comparison. To make the resource more practical for routine empirical work, we also proposed Multiverse-core, a practical 66 dataset subset designed to reduce computational burden while retaining a broad range of problem types.

To support use of the archive, we reported a baseline benchmark over 17 multivariate-capable classifiers available in the \texttt{aeon} toolkit, using the default train-test splits and equal length, no-missing-value versions where appropriate. These experiments are intended as an initial validation and reference point rather than a definitive bake off. On the 74 datasets for which all methods completed, HC2 achieved the best average rank, although the leading group of methods remained competitive and many pairwise differences were not statistically significant. On Multiverse-core, only 42 of the 66 datasets had complete results for every method, and only five algorithms completed all 66 datasets, reinforcing the practical motivation for a curated benchmark subset even after filtering for tractability.

The archive is therefore both a benchmark resource and an infrastructure contribution. It gives the community a broader and more realistic basis for developing and evaluating MTSC methods, while also making clear that scalability, robustness, and reproducibility remain important open challenges. The archive is distributed through the Multiverse archive and supported through integration with aeon, making datasets, metadata, and benchmark results easier to access within an open and reproducible software ecosystem. There is considerable scope to extend the accompanying results with additional algorithms, stronger resampling-based evaluation, and broader coverage of modern deep learning and foundation-style approaches. Longer term, we envisage the archive evolving as a community resource through the addition of new datasets, expanded metadata, canonical cross-validation folds with consideration of dataset circumstances, improved support for missing values and unequal length series, and continually updated benchmark results. We also expect Multiverse-core to evolve as practical experience accumulates, helping define a stable but representative subset for routine empirical studies. We hope the Multiverse archive, together with \texttt{aeon} and the Multiverse leaderboard, will provide a useful foundation for future comparative studies, more informative bake offs, and a clearer view of progress in multivariate time series classification.

\section*{Acknowledgements}
This work is supported by the UK Engineering and Physical Sciences Research Council (EPSRC) grant number EP/W030756/2. The experiments were carried out on the IRIDIS High Performance Computing Facility at the University of Southampton. We would like to thank all those responsible for helping maintain the time series classification archives, researchers who have donated data to these archives as well as ours, and those contributing to open-source implementations of the algorithms we use.

\bibliographystyle{abbrv}
\bibliography{main,TSRegression,applications,machinelearning,TSCMaster,rehab}

\appendix

\section{Further Archive Details}
\label{app:datasets}

In this section, we provide tables of information for the multiverse. Table~\ref{tab:full_archive} combines the previous Section~\ref{sec:datasets} tables into a single table displaying number of cases, channels, time points and classes for each dataset. Table~\ref{tab:archive_classes} shows the proportion of each class present in each archive dataset. Table~\ref{tab:disc_regression} describes the discretisation for the regression problems.

    \begin{footnotesize}
    \begin{longtable}{ | p{5.5cm} | p{1.1cm} | p{1.1cm} | p{1.1cm} | p{1.25cm} | p{0.9cm} | p{1.25cm} | }
        \caption{Details for all 147 MTSC archive datasets. For datasets with unequal length series, both the minimum and maximum series lengths are shown. HouseholdPowerConsumption is compressed to HPC for formatting.}
        \label{tab:full_archive} \\

        \hline
        Dataset name & Train & Test & Channels & Length & Classes & Type \\
        \hline
        Alzheimers & 45 & 43 & 19 & 15000 & 3 & Biosignal \\ \hline
        AppliancesEnergy\_disc & 95 & 42 & 24 & 144 & 2 & Power \\ \hline
        ArticularyWordRecognition & 275 & 300 & 9 & 144 & 25 & Motion \\ \hline
        AsphaltObstaclesCoordinates & 390 & 391 & 3 & 111-736 & 4 & Motion \\ \hline
        AsphaltObstaclesCoordinates\_eq & 390 & 391 & 3 & 298 & 4 & Motion \\ \hline
        AsphaltPavementTypeCoordinates & 1055 & 1056 & 3 & 66-2371 & 3 & Motion \\ \hline
        AsphaltPavementTypeCoordinates\_eq & 1055 & 1056 & 3 & 400 & 3 & Motion \\ \hline
        AsphaltRegularityCoordinates & 751 & 751 & 3 & 66-4201 & 2 & Motion \\ \hline
        AsphaltRegularityCoordinates\_eq & 751 & 751 & 3 & 384 & 2 & Motion \\ \hline
        AtrialFibrillation & 15 & 15 & 2 & 640 & 3 & Biosignal \\ \hline
        AustraliaRainfall\_disc & 112,186 & 48,081 & 3 & 24 & 4 & Sensor \\ \hline
        AutomotiveRoadTrials & 153 & 77 & 8 & 300 & 2 & Sensor \\ \hline
        BasicMotions & 40 & 40 & 6 & 100 & 4 & Motion \\ \hline
        BeijingPM10Quality\_disc & 11918 & 5048 & 9 & 24 & 2 & Sensor \\ \hline
        BeijingPM10Quality\_disc\_nmv & 11918 & 5048 & 9 & 24 & 2 & Sensor \\ \hline
        BeijingPM25Quality\_disc & 11918 & 5048 & 9 & 24 & 2 & Sensor \\ \hline
        BeijingPM25Quality\_disc\_nmv & 11918 & 5048 & 9 & 24 & 2 & Sensor \\ \hline
        BenzeneConcentration\_disc & 3349 & 5163 & 8 & 240 & 2 & Sensor \\ \hline
        BenzeneConcentration\_disc\_nmv & 3349 & 5163 & 8 & 240 & 2 & Sensor \\ \hline
        BIDMC32HR\_disc & 5550 & 2399 & 2 & 4000 & 3 & Biosignal \\ \hline
        BIDMC32RR\_disc & 5471 & 2399 & 2 & 4000 & 3 & Biosignal \\ \hline
        BIDMC32SpO2\_disc & 5550 & 2399 & 2 & 4000 & 2 & Biosignal \\ \hline
        Blink & 500 & 450 & 4 & 510 & 2 & Biosignal \\ \hline
        BoneIntensitiesAgeGroup & 600 & 445 & 8 & 270 & 3 & Image \\ \hline
        BoneProbAgeGroup & 600 & 445 & 8 & 256 & 3 & Image \\ \hline
        ButtonPress & 120 & 120 & 32 & 1000 & 2 & Biosignal \\ \hline
        CharacterTrajectories & 1422 & 1436 & 3 & 60-182 & 20 & Motion \\ \hline
        CharacterTrajectories\_eq & 1422 & 1436 & 3 & 119 & 20 & Motion \\ \hline
        CounterMovementJump & 419 & 179 & 3 & 2001-7069 & 3 & Motion \\ \hline
        CounterMovementJump\_eq & 419 & 179 & 3 & 3000 & 3 & Motion \\ \hline
        Cricket & 108 & 72 & 6 & 1197 & 12 & Motion \\ \hline
        CrowdSourced & 9456 & 2833 & 14 & 256 & 2 & Biosignal \\ \hline
        DREAMERA & 140638 & 29608 & 14 & 256 & 2 & Biosignal \\ \hline
        DREAMERV & 140638 & 29608 & 14 & 256 & 2 & Biosignal \\ \hline
        DuckDuckGeese & 50 & 50 & 1345 & 270 & 5 & Audio \\ \hline
        EigenWorms & 128 & 131 & 6 & 17984 & 5 & Motion \\ \hline
        EmoPain & 968 & 355 & 30 & 200 & 3 & Motion \\ \hline
        Epilepsy & 137 & 138 & 3 & 206 & 4 & Motion \\ \hline
        ERing & 30 & 270 & 4 & 65 & 6 & Motion \\ \hline
        EthanolConcentration & 261 & 263 & 3 & 1751 & 4 & Spectro \\ \hline
        EyesOpenShut & 56 & 42 & 14 & 128 & 2 & Biosignal \\ \hline
        FaceDetection & 5890 & 3524 & 144 & 62 & 2 & Biosignal \\ \hline
        FeedbackButton & 1700 & 1150 & 61 & 450 & 2 & Biosignal \\ \hline
        FeetHands & 2835 & 1890 & 64 & 640 & 2 & Biosignal \\ \hline
        FingerMovements & 316 & 100 & 28 & 50 & 2 & Biosignal \\ \hline
        FordChallenge & 29003 & 7254 & 30 & 40 & 2 & Sensor \\ \hline
        HandMovementDirection & 160 & 74 & 10 & 400 & 4 & Biosignal \\ \hline
        Handwriting & 150 & 850 & 3 & 152 & 26 & Motion \\ \hline
        Heartbeat & 204 & 205 & 61 & 405 & 2 & Audio \\ \hline
        HPC1\_disc & 745 & 686 & 5 & 1440 & 3 & Power \\ \hline
        HPC1\_disc\_nmv & 745 & 686 & 5 & 1440 & 3 & Power \\ \hline
        HPC2\_disc & 745 & 686 & 5 & 1440 & 3 & Power \\ \hline
        HPC2\_disc\_nmv & 745 & 686 & 5 & 1440 & 3 & Power \\ \hline
        IEEEPPG\_disc & 1768 & 1328 & 5 & 1000 & 3 & Biosignal \\ \hline
        ImaginedFeetHands & 2835 & 1935 & 64 & 640 & 2 & Biosignal \\ \hline
        ImaginedOpenCloseFist & 2835 & 1890 & 64 & 640 & 2 & Biosignal \\ \hline
        InnerSpeech & 250 & 200 & 128 & 3584 & 4 & Biosignal \\ \hline
        InsectWingbeat & 25000 & 25000 & 200 & 2-22 & 10 & Audio \\ \hline
        InsectWingbeat\_eq & 25000 & 25000 & 200 & 20 & 10 & Audio \\ \hline
        IRDS-EFL & 223 & 26 & 75 & 60 & 2 & Motion \\ \hline
        IRDS-EFR & 256 & 15 & 75 & 75 & 2 & Motion \\ \hline
        IRDS-SAL & 213 & 56 & 75 & 82 & 2 & Motion \\ \hline
        IRDS-SAR & 218 & 33 & 75 & 76 & 2 & Motion \\ \hline
        IRDS-SFE & 237 & 15 & 75 & 95 & 2 & Motion \\ \hline
        IRDS-SFL & 253 & 116 & 75 & 111 & 2 & Motion \\ \hline
        IRDS-SFR & 295 & 15 & 75 & 106 & 2 & Motion \\ \hline
        IRDS-STL & 226 & 38 & 75 & 66 & 2 & Motion \\ \hline
        IRDS-STR & 221 & 60 & 75 & 66 & 2 & Motion \\ \hline
        JapaneseVowels & 270 & 370 & 12 & 7-29 & 9 & Audio \\ \hline
        JapaneseVowels\_eq & 270 & 370 & 12 & 25 & 9 & Audio \\ \hline
        KERAAL-CTK & 271 & 14 & 77 & 276 & 2 & Motion \\ \hline
        KERAAL-CTK-MC & 271 & 14 & 77 & 276 & 4** & Motion \\ \hline
        KERAAL-ELK & 247 & 10 & 77 & 295 & 2 & Motion \\ \hline
        KERAAL-ELK-MC & 243 & 14 & 77 & 295 & 4** & Motion \\ \hline
        KERAAL-RTK & 256 & 14 & 77 & 264 & 2 & Motion \\ \hline
        KERAAL-RTK-MC & 257 & 13 & 77 & 264 & 4** & Motion \\ \hline
        KIMORE-LA-C & 64 & 7 & 54 & 725 & 2 & Motion \\ \hline
        KIMORE-LT-C & 64 & 7 & 54 & 798 & 2 & Motion \\ \hline
        KIMORE-PR-C & 64 & 7 & 54 & 847 & 2 & Motion \\ \hline
        KIMORE-Sq-C & 64 & 7 & 54 & 557 & 2 & Motion \\ \hline
        KIMORE-TR-C & 64 & 7 & 54 & 813 & 2 & Motion \\ \hline
        KINECAL-3WFV & 48 & 12 & 75 & 257 & 2 & Motion \\ \hline
        KINECAL-GGFV & 49 & 13 & 75 & 296 & 2 & Motion \\ \hline
        KINECAL-QSEC & 69 & 18 & 75 & 592 & 2 & Motion \\ \hline
        KINECAL-QSEO & 68 & 17 & 75 & 593 & 2 & Motion \\ \hline
        LenDB & 983483 & 261459 & 3 & 540 & 2 & Sensor \\ \hline
        Libras & 180 & 180 & 2 & 45 & 15 & Motion \\ \hline
        LiveFuelMoistureContent\_disc & 3493 & 1510 & 7 & 365 & 2 & Spectro \\ \hline
        Locust2022 & 6597 & 3299 & 13 & 96 & 2 & Sensor \\ \hline
        LongIntervalTask & 1660 & 1120 & 63 & 4750 & 2 & Biosignal \\ \hline
        LowCost & 600 & 600 & 15 & 375 & 2 & Biosignal \\ \hline
        LSST & 2459 & 2466 & 6 & 36 & 14 & Sensor \\ \hline
        MatchingPennies & 700 & 700 & 10 & 10000 & 2 & Biosignal \\ \hline
        MindReading & 727 & 653 & 204 & 200 & 5 & Biosignal \\ \hline
        MotionSenseHAR & 966 & 265 & 12 & 1000 & 6 & Motion \\ \hline
        MotorImagery & 278 & 100 & 64 & 3000 & 2 & Biosignal \\ \hline
        NATOPS & 180 & 180 & 24 & 51 & 6 & Motion \\ \hline
        NewsHeadlineSentiment\_disc & 58213 & 24951 & 3 & 144 & 3 & Digital \\ \hline
        NewsTitleSentiment\_disc & 58213 & 24951 & 3 & 144 & 3 & Digital \\ \hline
        OpenCloseFist & 2835 & 1890 & 64 & 640 & 2 & Biosignal \\ \hline
        Opportunity & 12184 & 5202 & 113 & 100 & 5 & Motion \\ \hline
        PAMAP2 & 28961 & 9895 & 52 & 100 & 12 & Motion \\ \hline
        PEMS-SF & 267 & 173 & 963 & 144 & 7 & Sensor \\ \hline
        PenDigits & 7494 & 3498 & 2 & 8 & 10 & Motion \\ \hline
        PhonemeSpectra & 3315 & 3353 & 11 & 217 & 39 & Audio \\ \hline
        PhotoStimulation & 37 & 36 & 19 & 9000 & 3 & Biosignal \\ \hline
        PPGDalia\_disc & 43215 & 21482 & 4 & 256-512*** & 3 & Biosignal \\ \hline
        PPGDalia\_disc\_eq & 43215 & 21482 & 4 & 256 & 3 & Biosignal \\ \hline
        PronouncedSpeech & 250 & 200 & 128 & 3584 & 4 & Biosignal \\ \hline
        RacketSports & 151 & 152 & 6 & 30 & 4 & Motion \\ \hline
        S2Agri-10pc-17 & 4661858 & 1189023 & 10 & 24 & 17 & Image \\ \hline
        S2Agri-10pc-34 & 4661858 & 1189023 & 10 & 24 & 29 & Image \\ \hline
        S2Agri-17 & 47328752 & 11940071 & 10 & 24 & 17 & Image \\ \hline
        S2Agri-34 & 47328752 & 11940071 & 10 & 24 & 34** & Image \\ \hline
        SelfRegulationSCP1 & 268 & 293 & 6 & 896 & 2 & Biosignal \\ \hline
        SelfRegulationSCP2 & 200 & 180 & 7 & 1152 & 2 & Biosignal \\ \hline
        ShortIntervalTask & 1360 & 1420 & 63 & 2750 & 2 & Biosignal \\ \hline
        SitStand & 3200 & 3200 & 17 & 1250 & 4 & Biosignal \\ \hline
        Skoda & 11290 & 2827 & 60 & 100 & 11 & Motion \\ \hline
        SongFamiliarity & 700 & 750 & 32 & 3000 & 2 & Biosignal \\ \hline
        SPHERE-WUS & 44 & 4 & 45 & 214 & 2 & Motion \\ \hline
        SpokenArabicDigits & 6599 & 2199 & 13 & 4-93 & 10 & Audio \\ \hline
        SpokenArabicDigits\_eq & 6599 & 2199 & 13 & 65 & 10 & Audio \\ \hline
        StandWalkJump & 12 & 15 & 4 & 2500 & 3 & Biosignal \\ \hline
        STEW & 21384 & 7128 & 14 & 256 & 2 & Biosignal \\ \hline
        TactileTextureRecognition & 1586 & 681 & 6 & 2000 & 21 & Sensor \\ \hline
        Tiselac & 79907 & 19780 & 10 & 23 & 9 & Image \\ \hline
        UCDHE-MP & 1495 & 362 & 16 & 161 & 2 & Motion \\ \hline
        UCDHE-MP-MC & 1495 & 362 & 16 & 161 & 4 & Motion \\ \hline
        UCDHE-Rowing & 1992 & 440 & 16 & 161 & 2 & Motion \\ \hline
        UCDHE-Rowing-MC & 1992 & 440 & 16 & 161 & 5 & Motion \\ \hline
        UCIActivity & 7898 & 2401 & 9 & 128 & 6 & Motion \\ \hline
        UIPRMD-DS-C & 144 & 36 & 66 & 81 & 2 & Motion \\ \hline
        UIPRMD-HS-C & 92 & 18 & 66 & 69 & 2 & Motion \\ \hline
        UIPRMD-IL-C & 84 & 18 & 66 & 77 & 2 & Motion \\ \hline
        UIPRMD-SASLR-C & 126 & 20 & 66 & 63 & 2 & Motion \\ \hline
        UIPRMD-SL-C & 122 & 18 & 66 & 85 & 2 & Motion \\ \hline
        UIPRMD-SSA-C & 108 & 18 & 66 & 74 & 2 & Motion \\ \hline
        UIPRMD-SSE-C & 108 & 18 & 66 & 67 & 2 & Motion \\ \hline
        UIPRMD-SSIER-C & 100 & 20 & 66 & 74 & 2 & Motion \\ \hline
        UIPRMD-SSS-C & 90 & 18 & 66 & 66 & 2 & Motion \\ \hline
        UIPRMD-STS-C & 132 & 36 & 66 & 88 & 2 & Motion \\ \hline
        USCActivity & 43060 & 13168 & 6 & 100 & 12 & Motion \\ \hline
        UWaveGestureLibrary & 120 & 320 & 3 & 315 & 8 & Motion \\ \hline
        VisualSpeech & 250 & 200 & 128 & 3584 & 4 & Biosignal \\ \hline
        WISDM & 11989 & 5177 & 3 & 100 & 6 & Motion \\ \hline
        WISDM2 & 98094 & 50940 & 3 & 100 & 6 & Motion \\ \hline

        \multicolumn{7}{l}{* These datasets contain missing values.} \\
        \multicolumn{7}{l}{** These datasets are missing classes in the test set which are present in the train set.} \\
        \multicolumn{7}{l}{*** Series length differs by channel instead of case for PPGDalia.} \\
    \end{longtable}

    \begin{longtable}{ | p{5.5cm} | p{8.5cm} | }
        \caption{The class values for all 133 archive datasets, excluding cleaned variants. Each class value is followed by the percentage of the dataset it represents (both train and test sets concatenated). The class names used in the \textit{.ts} files for each problem are displayed. HouseholdPowerConsumption is compressed to HPC for formatting.}
        \label{tab:archive_classes} \\

        \hline
        Dataset name & Classes \\
        \hline
        Alzheimers & `0.0: 40.91\%, `1.0: 32.95\%, `2.0: 26.14\% \\ \hline
        AppliancesEnergy\_disc & `0.0: 71.53\%, `1.0: 28.47\% \\ \hline
        ArticularyWordRecognition & `1.0: 4.0\%, `10.0: 4.0\%, `11.0: 4.0\%, `12.0: 4.0\%, `13.0: 4.0\%, `14.0: 4.0\%, `15.0: 4.0\%, `16.0: 4.0\%, `17.0: 4.0\%, `18.0: 4.0\%, `19.0: 4.0\%, `2.0: 4.0\%, `20.0: 4.0\%, `21.0: 4.0\%, `22.0: 4.0\%, `23.0: 4.0\%, `24.0: 4.0\%, `25.0: 4.0\%, `3.0: 4.0\%, `4.0: 4.0\%, `5.0: 4.0\%, `6.0: 4.0\%, `7.0: 4.0\%, `8.0: 4.0\%, `9.0: 4.0\% \\ \hline
        AsphaltObstaclesCoordinates & `raised\_crosswalk: 20.49\%, `raised\_markers: 23.94\%, `speed\_bump: 27.14\%, `vertical\_patch: 28.43\% \\ \hline
        AsphaltPavementTypeCoordinates & `cobblestone: 24.96\%, `dirt: 36.38\%, `flexible: 38.65\% \\ \hline
        AsphaltRegularityCoordinates & `deteriorated: 49.27\%, `regular: 50.73\% \\ \hline
        AtrialFibrillation & `n: 33.33\%, `s: 33.33\%, `t: 33.33\% \\ \hline
        AustraliaRainfall\_disc & `0.0: 68.52\%, `1.0: 27.64\%, `2.0: 3.76\%, `3.0: 0.08\% \\ \hline
        AutomotiveRoadTrials & `fault: 22.17\%, `normal: 77.83\% \\ \hline
        BasicMotions & `badminton: 25.0\%, `running: 25.0\%, `standing: 25.0\%, `walking: 25.0\% \\ \hline
        BeijingPM10Quality\_disc & `0.0: 23.01\%, `1.0: 76.99\% \\ \hline
        BeijingPM25Quality\_disc & `0.0: 25.99\%, `1.0: 74.01\% \\ \hline
        BenzeneConcentration\_disc & `0.0: 28.65\%, `1.0: 71.35\% \\ \hline
        BIDMC32HR\_disc & `0.0: 19.35\%, `1.0: 63.97\%, `2.0: 16.68\% \\ \hline
        BIDMC32RR\_disc & `0.0: 4.88\%, `1.0: 80.8\%, `2.0: 14.32\% \\ \hline
        BIDMC32SpO2\_disc & `0.0: 21.39\%, `1.0: 78.61\% \\ \hline
        Blink & `longblink: 47.37\%, `shortblink: 52.63\% \\ \hline
        BoneIntensitiesAgeGroup & `1: 20.67\%, `2: 42.01\%, `3: 37.32\% \\ \hline
        BoneProbAgeGroup & `1: 20.67\%, `2: 42.01\%, `3: 37.32\% \\ \hline
        ButtonPress & `0.0: 50.0\%, `1.0: 50.0\% \\ \hline
        CharacterTrajectories & `10: 4.55\%, `11: 4.93\%, `12: 4.58\%, `13: 4.34\%, `14: 4.16\%, `15: 4.65\%, `16: 4.58\%, `17: 5.42\%, `18: 4.37\%, `19: 4.79\%, `1: 5.98\%, `20: 5.98\%, `2: 4.93\%, `3: 4.97\%, `4: 5.49\%, `5: 6.51\%, `6: 4.83\%, `7: 4.44\%, `8: 6.09\%, `9: 4.37\% \\ \hline
        CounterMovementJump & `0.0: 33.44\%, `1.0: 33.28\%, `2.0: 33.28\% \\ \hline
        Cricket & `1.0: 8.33\%, `10.0: 8.33\%, `11.0: 8.33\%, `12.0: 8.33\%, `2.0: 8.33\%, `3.0: 8.33\%, `4.0: 8.33\%, `5.0: 8.33\%, `6.0: 8.33\%, `7.0: 8.33\%, `8.0: 8.33\%, `9.0: 8.33\% \\ \hline
        CrowdSourced & `0: 50.01\%, `1: 49.99\% \\ \hline
        DREAMERA & `0: 52.05\%, `1: 47.95\% \\ \hline
        DREAMERV & `0: 58.9\%, `1: 41.1\% \\ \hline
        DuckDuckGeese & `black-bellied\_whistling\_duck: 20.0\%, `canadian\_goose: 20.0\%, `greylag\_goose: 20.0\%, `pink-footed\_goose: 20.0\%, `white-faced\_whistling\_duck: 20.0\% \\ \hline
        EigenWorms & `1: 42.47\%, `2: 16.99\%, `3: 13.51\%, `4: 17.37\%, `5: 9.65\% \\ \hline
        EmoPain & `0: 77.25\%, `1: 14.29\%, `2: 8.47\% \\ \hline
        Epilepsy & `epilepsy: 24.73\%, `running: 26.55\%, `sawing: 21.82\%, `walking: 26.91\% \\ \hline
        ERing & `1: 16.67\%, `2: 16.67\%, `3: 16.67\%, `4: 16.67\%, `5: 16.67\%, `6: 16.67\% \\ \hline
        EthanolConcentration & `e35: 25.0\%, `e38: 25.0\%, `e40: 25.19\%, `e45: 24.81\% \\ \hline
        EyesOpenShut & `0: 55.1\%, `1: 44.9\% \\ \hline
        FaceDetection & `0: 50.0\%, `1: 50.0\% \\ \hline
        FeedbackButton & `0.0: 50.0\%, `1.0: 50.0\% \\ \hline
        FeetHands & `0.0: 49.65\%, `1.0: 50.35\% \\ \hline
        FingerMovements & `left: 50.0\%, `right: 50.0\% \\ \hline
        FordChallenge & `0: 36.5\%, `1: 63.5\% \\ \hline
        HandMovementDirection & `backward: 23.5\%, `forward: 29.91\%, `left: 23.5\%, `right: 23.08\% \\ \hline
        Handwriting & `1.0: 4.4\%, `10.0: 3.3\%, `11.0: 2.7\%, `12.0: 4.2\%, `13.0: 4.7\%, `14.0: 3.9\%, `15.0: 3.7\%, `16.0: 3.8\%, `17.0: 3.8\%, `18.0: 3.7\%, `19.0: 3.6\%, `2.0: 4.0\%, `20.0: 3.6\%, `21.0: 3.9\%, `22.0: 5.3\%, `23.0: 3.6\%, `24.0: 4.1\%, `25.0: 3.2\%, `26.0: 4.3\%, `3.0: 3.8\%, `4.0: 4.7\%, `5.0: 3.9\%, `6.0: 3.9\%, `7.0: 3.5\%, `8.0: 3.4\%, `9.0: 3.0\% \\ \hline
        Heartbeat & `abnormal: 72.13\%, `normal: 27.87\% \\ \hline
        HPC1\_disc & `0.0: 13.0\%, `1.0: 73.03\%, `2.0: 13.98\% \\ \hline
        HPC2\_disc & `0.0: 10.55\%, `1.0: 72.33\%, `2.0: 17.12\% \\ \hline
        IEEEPPG\_disc & `0.0: 21.25\%, `1.0: 36.82\%, `2.0: 41.93\% \\ \hline
        ImaginedFeetHands & `0.0: 50.06\%, `1.0: 49.94\% \\ \hline
        ImaginedOpenCloseFist & `0.0: 50.46\%, `1.0: 49.54\% \\ \hline
        InnerSpeech & `0.0: 23.56\%, `1.0: 25.78\%, `2.0: 26.67\%, `3.0: 24.0\% \\ \hline
        InsectWingbeat & `aedes\_female: 10.0\%, `aedes\_male: 10.0\%, `fruit\_flies: 10.0\%, `house\_flies: 10.0\%, `quinx\_female: 10.0\%, `quinx\_male: 10.0\%, `stigma\_female: 10.0\%, `stigma\_male: 10.0\%, `tarsalis\_female: 10.0\%, `tarsalis\_male: 10.0\% \\ \hline
        IRDS-EFL & `1: 82.73\%, `2: 17.27\% \\ \hline
        IRDS-EFR & `1: 85.24\%, `2: 14.76\% \\ \hline
        IRDS-SAL & `1: 83.27\%, `2: 16.73\% \\ \hline
        IRDS-SAR & `1: 84.86\%, `2: 15.14\% \\ \hline
        IRDS-SFE & `1: 83.73\%, `2: 16.27\% \\ \hline
        IRDS-SFL & `1: 60.43\%, `2: 39.57\% \\ \hline
        IRDS-SFR & `1: 66.45\%, `2: 33.55\% \\ \hline
        IRDS-STL & `1: 87.88\%, `2: 12.12\% \\ \hline
        IRDS-STR & `1: 88.61\%, `2: 11.39\% \\ \hline
        JapaneseVowels & `1: 9.53\%, `2: 10.16\%, `3: 18.44\%, `4: 11.56\%, `5: 9.22\%, `6: 8.44\%, `7: 10.94\%, `8: 12.5\%, `9: 9.22\% \\ \hline
        KERAAL-CTK & `c: 37.89\%, `e: 62.11\% \\ \hline
        KERAAL-CTK-MC & `c: 37.89\%, `e1: 27.02\%, `e2: 17.19\%, `e3: 17.89\% \\ \hline
        KERAAL-ELK & `c: 24.12\%, `e: 75.88\% \\ \hline
        KERAAL-ELK-MC & `c: 24.12\%, `e1: 17.51\%, `e2: 40.86\%, `e3: 17.51\% \\ \hline
        KERAAL-RTK & `c: 38.15\%, `e: 61.85\% \\ \hline
        KERAAL-RTK-MC & `c: 38.15\%, `e1: 18.52\%, `e2: 26.67\%, `e3: 16.67\% \\ \hline
        KIMORE-LA-C & `0: 12.68\%, `1: 87.32\% \\ \hline
        KIMORE-LT-C & `0: 30.99\%, `1: 69.01\% \\ \hline
        KIMORE-PR-C & `0: 36.62\%, `1: 63.38\% \\ \hline
        KIMORE-Sq-C & `0: 26.76\%, `1: 73.24\% \\ \hline
        KIMORE-TR-C & `0: 18.31\%, `1: 81.69\% \\ \hline
        KINECAL-3WFV & `0: 90.0\%, `1: 10.0\% \\ \hline
        KINECAL-GGFV & `0: 90.32\%, `1: 9.68\% \\ \hline
        KINECAL-QSEC & `0: 94.25\%, `1: 5.75\% \\ \hline
        KINECAL-QSEO & `0: 94.12\%, `1: 5.88\% \\ \hline
        LenDB & `0: 49.47\%, `1: 50.53\% \\ \hline
        Libras & `10: 6.67\%, `11: 6.67\%, `12: 6.67\%, `13: 6.67\%, `14: 6.67\%, `15: 6.67\%, `1: 6.67\%, `2: 6.67\%, `3: 6.67\%, `4: 6.67\%, `5: 6.67\%, `6: 6.67\%, `7: 6.67\%, `8: 6.67\%, `9: 6.67\% \\ \hline
        LiveFuelMoistureContent\_disc & `0.0: 16.43\%, `1.0: 83.57\% \\ \hline
        Locust2022 & `negativo: 91.11\%, `positivo: 8.89\% \\ \hline
        LongIntervalTask & `0.0: 32.37\%, `1.0: 67.63\% \\ \hline
        LowCost & `0.0: 50.0\%, `1.0: 50.0\% \\ \hline
        LSST & `15: 5.02\%, `16: 10.96\%, `42: 15.49\%, `52: 2.54\%, `53: 0.28\%, `62: 6.21\%, `64: 0.95\%, `65: 12.71\%, `67: 2.76\%, `6: 1.4\%, `88: 4.89\%, `90: 31.55\%, `92: 3.13\%, `95: 2.09\% \\ \hline
        MatchingPennies & `0.0: 50.0\%, `1.0: 50.0\% \\ \hline
        MindReading & `1: 21.74\%, `2: 24.06\%, `3: 15.07\%, `4: 19.57\%, `5: 19.57\% \\ \hline
        MotionSenseHAR & `dws: 7.55\%, `jog: 8.77\%, `sit: 25.59\%, `std: 23.15\%, `ups: 9.91\%, `wlk: 25.02\% \\ \hline
        MotorImagery & `finger: 50.0\%, `tongue: 50.0\% \\ \hline
        NATOPS & `1.0: 16.67\%, `2.0: 16.67\%, `3.0: 16.67\%, `4.0: 16.67\%, `5.0: 16.67\%, `6.0: 16.67\% \\ \hline
        NewsHeadlineSentiment\_disc & `0.0: 28.3\%, `1.0: 54.98\%, `2.0: 16.72\% \\ \hline
        NewsTitleSentiment\_disc & `0.0: 20.01\%, `1.0: 62.53\%, `2.0: 17.45\% \\ \hline
        OpenCloseFist & `0.0: 50.14\%, `1.0: 49.86\% \\ \hline
        Opportunity & `0: 18.5\%, `1: 39.65\%, `2: 23.1\%, `3: 15.8\%, `4: 2.94\% \\ \hline
        PAMAP2 & `0: 9.91\%, `10: 12.28\%, `11: 2.54\%, `1: 9.53\%, `2: 9.78\%, `3: 12.29\%, `4: 5.05\%, `5: 8.47\%, `6: 9.68\%, `7: 6.04\%, `8: 5.4\%, `9: 9.03\% \\ \hline
        PEMS-SF & `1.0: 14.09\%, `2.0: 12.95\%, `3.0: 14.55\%, `4.0: 14.77\%, `5.0: 14.77\%, `6.0: 14.32\%, `7.0: 14.55\% \\ \hline
        PenDigits & `0: 10.4\%, `1: 10.4\%, `2: 10.41\%, `3: 9.6\%, `4: 10.41\%, `5: 9.6\%, `6: 9.61\%, `7: 10.39\%, `8: 9.6\%, `9: 9.6\% \\ \hline
        PhonemeSpectra & `aa: 2.56\%, `ae: 2.56\%, `ah: 2.56\%, `ao: 2.56\%, `aw: 2.56\%, `ay: 2.56\%, `b: 2.56\%, `ch: 2.56\%, `d: 2.56\%, `dh: 2.56\%, `eh: 2.56\%, `er: 2.56\%, `ey: 2.56\%, `f: 2.56\%, `g: 2.56\%, `hh: 2.56\%, `ih: 2.56\%, `iy: 2.56\%, `jh: 2.56\%, `k: 2.56\%, `l: 2.56\%, `m: 2.56\%, `n: 2.56\%, `ng: 2.56\%, `ow: 2.56\%, `oy: 2.56\%, `p: 2.56\%, `r: 2.56\%, `s: 2.56\%, `sh: 2.56\%, `t: 2.56\%, `th: 2.56\%, `uh: 2.56\%, `uw: 2.56\%, `v: 2.56\%, `w: 2.56\%, `y: 2.56\%, `z: 2.56\%, `zh: 2.55\% \\ \hline
        PhotoStimulation & `0.0: 42.47\%, `1.0: 30.14\%, `2.0: 27.4\% \\ \hline
        PPGDalia\_disc & `0.0: 27.87\%, `1.0: 58.09\%, `2.0: 14.04\% \\ \hline
        PronouncedSpeech & `0.0: 25.78\%, `1.0: 24.44\%, `2.0: 24.89\%, `3.0: 24.89\% \\ \hline
        RacketSports & `badminton\_clear: 28.38\%, `badminton\_smash: 26.07\%, `squash\_backhandboast: 22.44\%, `squash\_forehandboast: 23.1\% \\ \hline
        S2Agri-10pc-17 & `0: 0.01\%, `10: 0.05\%, `11: 0.05\%, `12: 0.02\%, `13: 2.36\%, `14: 0.0\%, `15: 1.22\%, `16: 0.12\%, `1: 59.33\%, `2: 29.25\%, `3: 0.02\%, `4: 1.09\%, `5: 0.05\%, `6: 3.81\%, `7: 2.4\%, `8: 0.06\%, `9: 0.16\% \\ \hline
        S2Agri-10pc-34 & `0: 0.01\%, `10: 1.09\%, `11: 0.05\%, `12: 3.34\%, `13: 0.37\%, `14: 0.1\%, `15: 2.2\%, `16: 0.2\%, `17: 0.02\%, `18: 0.04\%, `19: 0.16\%, `1: 59.33\%, `20: 0.05\%, `21: 0.05\%, `22: 0.02\%, `23: 2.36\%, `24: 0.0\%, `25: 1.22\%, `26: 0.01\%, `27: 0.03\%, `28: 0.04\%, `2: 0.07\%, `3: 0.21\%, `4: 13.73\%, `5: 14.89\%, `6: 0.08\%, `7: 0.27\%, `8: 0.03\%, `9: 0.02\% \\ \hline
        S2Agri-17 & `0: 0.01\%, `10: 0.02\%, `11: 0.12\%, `12: 0.02\%, `13: 2.18\%, `14: 0.0\%, `15: 0.87\%, `16: 0.08\%, `1: 59.42\%, `2: 29.42\%, `3: 0.02\%, `4: 1.21\%, `5: 0.04\%, `6: 3.85\%, `7: 2.49\%, `8: 0.07\%, `9: 0.18\% \\ \hline
        S2Agri-34 & `0: 0.01\%, `10: 0.02\%, `11: 1.21\%, `12: 0.04\%, `13: 3.35\%, `14: 0.0\%, `15: 0.44\%, `16: 0.06\%, `17: 2.32\%, `18: 0.17\%, `19: 0.0\%, `1: 59.42\%, `20: 0.03\%, `21: 0.04\%, `22: 0.0\%, `23: 0.17\%, `24: 0.02\%, `25: 0.12\%, `26: 0.02\%, `27: 2.18\%, `28: 0.0\%, `29: 0.87\%, `2: 0.0\%, `30: 0.0\%, `31: 0.01\%, `32: 0.02\%, `33: 0.03\%, `3: 0.09\%, `4: 0.23\%, `5: 13.6\%, `6: 15.1\%, `7: 0.09\%, `8: 0.29\%, `9: 0.02\% \\ \hline
        SelfRegulationSCP1 & `negativity: 50.27\%, `positivity: 49.73\% \\ \hline
        SelfRegulationSCP2 & `negativity: 50.0\%, `positivity: 50.0\% \\ \hline
        ShortIntervalTask & `0.0: 32.37\%, `1.0: 67.63\% \\ \hline
        SitStand & `0.0: 25.0\%, `1.0: 25.0\%, `2.0: 25.0\%, `3.0: 25.0\% \\ \hline
        Skoda & `0: 23.58\%, `10: 5.49\%, `1: 8.87\%, `2: 10.43\%, `3: 9.98\%, `4: 7.23\%, `5: 4.4\%, `6: 4.14\%, `7: 7.64\%, `8: 8.42\%, `9: 9.82\% \\ \hline
        SongFamiliarity & `0.0: 46.07\%, `1.0: 53.93\% \\ \hline
        SPHERE-WUS & `0: 35.42\%, `1: 64.58\% \\ \hline
        SpokenArabicDigits & `10: 9.98\%, `1: 10.0\%, `2: 10.0\%, `3: 10.0\%, `4: 10.0\%, `5: 10.0\%, `6: 10.0\%, `7: 10.0\%, `8: 10.0\%, `9: 10.0\% \\ \hline
        StandWalkJump & `jumping: 33.33\%, `standing: 33.33\%, `walking: 33.33\% \\ \hline
        STEW & `0: 50.0\%, `1: 50.0\% \\ \hline
        TactileTextureRecognition & `aluminium: 3.97\%, `blackboard-cloth: 4.85\%, `book-cover: 4.9\%, `cardboard-fine: 5.03\%, `cardboard: 5.03\%, `cleaning-cloth: 4.94\%, `computer-paper: 4.76\%, `cork: 4.41\%, `cotton-bag: 5.07\%, `cutting-board: 4.68\%, `cutting-mat: 4.1\%, `glass: 4.72\%, `jeans: 4.5\%, `marble-tile: 5.07\%, `newspaper: 4.9\%, `osb: 4.98\%, `plastic-bag: 5.21\%, `polymer-tile: 4.72\%, `stone-tile: 4.94\%, `towel: 4.41\%, `wood: 4.81\% \\ \hline
        Tiselac & `0: 20.06\%, `1: 3.9\%, `2: 20.06\%, `3: 19.46\%, `4: 15.58\%, `5: 6.84\%, `6: 9.22\%, `7: 1.76\%, `8: 3.13\% \\ \hline
        UCDHE-MP & `e: 74.15\%, `n: 25.85\% \\ \hline
        UCDHE-MP-MC & `a: 24.23\%, `arch: 24.23\%, `n: 25.85\%, `r: 25.69\% \\ \hline
        UCDHE-Rowing & `e: 79.44\%, `n: 20.56\% \\ \hline
        UCDHE-Rowing-MC & `a: 20.23\%, `ext: 18.5\%, `n: 20.56\%, `r: 20.97\%, `rb: 19.74\% \\ \hline
        UCIActivity & `0: 16.72\%, `1: 14.99\%, `2: 13.65\%, `3: 17.25\%, `4: 18.51\%, `5: 18.88\% \\ \hline
        UIPRMD-DS-C & `0: 50.0\%, `1: 50.0\% \\ \hline
        UIPRMD-HS-C & `0: 50.0\%, `1: 50.0\% \\ \hline
        UIPRMD-IL-C & `0: 50.0\%, `1: 50.0\% \\ \hline
        UIPRMD-SASLR-C & `0: 50.0\%, `1: 50.0\% \\ \hline
        UIPRMD-SL-C & `0: 50.0\%, `1: 50.0\% \\ \hline
        UIPRMD-SSA-C & `0: 50.0\%, `1: 50.0\% \\ \hline
        UIPRMD-SSE-C & `0: 50.0\%, `1: 50.0\% \\ \hline
        UIPRMD-SSIER-C & `0: 50.0\%, `1: 50.0\% \\ \hline
        UIPRMD-SSS-C & `0: 50.0\%, `1: 50.0\% \\ \hline
        UIPRMD-STS-C & `0: 50.0\%, `1: 50.0\% \\ \hline
        USCActivity & `0: 13.57\%, `10: 5.88\%, `11: 5.87\%, `1: 9.42\%, `2: 9.59\%, `3: 7.54\%, `4: 7.02\%, `5: 6.28\%, `6: 3.81\%, `7: 9.29\%, `8: 8.39\%, `9: 13.33\% \\ \hline
        UWaveGestureLibrary & `1.0: 12.5\%, `2.0: 12.5\%, `3.0: 12.5\%, `4.0: 12.5\%, `5.0: 12.5\%, `6.0: 12.5\%, `7.0: 12.5\%, `8.0: 12.5\% \\ \hline
        VisualSpeech & `0.0: 24.0\%, `1.0: 25.33\%, `2.0: 26.0\%, `3.0: 24.67\% \\ \hline
        WISDM & `0: 9.73\%, `1: 37.68\%, `2: 1.34\%, `3: 1.03\%, `4: 10.22\%, `5: 39.99\% \\ \hline
        WISDM2 & `0: 14.73\%, `1: 9.26\%, `2: 22.27\%, `3: 1.93\%, `4: 9.69\%, `5: 42.13\% \\ \hline
    \end{longtable}
    \end{footnotesize}

    \begin{table}[]
        \caption{The original target label split points used to create class labels for the discretised regression problems. HouseholdPowerConsumption is compressed to HPC for formatting.}
        \label{tab:disc_regression}
        \footnotesize
        \centering
        \begin{tabular}{ | p{4cm} | p{5.2cm} | p{4.5cm} | }
            \hline
            Dataset & Label split point(s) & Split description  \\
            \hline
            AppliancesEnergy & \makecell[l]{Below average usage $<15.81$ \\ Above average usage: $\geq15.81$} & Midrange of target values \\ \hline
            HPC1 & \makecell[l]{Low consumption: $<936$ \\ Medium consumption: $\geq936, <2150$\\ High consumption: $\geq2150$} & Median of target values + and - standard deviation \\ \hline
            HPC2 & \makecell[l]{Low consumption: $<121$ \\ Medium consumption: $\geq121, <221$ \\ High consumption: $\geq221$} & Median of target values + and - standard deviation \\ \hline
            BenzeneConcentration & \makecell[l]{Below EU limit: $<5$ \\ Above EU limit: $\geq5$} & European Union air quality standards limit of 5 $\mu g/m^3$\tablefootnote{\url{https://web.archive.org/web/20241118204135/https://environment.ec.europa.eu/topics/air/air-quality/eu-air-quality-standards_en}}  \\ \hline
            BeijingPM25Quality & \makecell[l]{Below 2013 median: $<70$ \\ Above 2013 median: $\geq70$} & 2013 median of 74 Chinese cities from Table 1 of \cite{huang2018health} \\ \hline
            BeijingPM10Quality & \makecell[l]{Below 2013 median: $<108$ \\ Above 2013 median: $\geq108$} & 2013 median of 74 Chinese cities from Table 1 of \cite{huang2018health} \\ \hline
            LiveFuelMoistureContent & \makecell[l]{Fire risk FMC: $<80$ \\ Suitable FMC: $\geq80$} &  The risk of fire increases rapidly as fuel moisture content goes below 80\%~\cite{tan2020monash,yebra2018fuel} \\ \hline
            AustraliaRainfall & \makecell[l]{No rain: $=0$ \\ Slight rain: $>0, \leq12$ \\ Moderate rain: $>12, \leq96$ \\ Heavy rain: $>96$} & UK Met Office synoptic rain categorisation (hourly rate multiplied by 24 as the target is daily rainfall)\tablefootnote{\url{https://web.archive.org/web/20120114162401/http://www.metoffice.gov.uk/media/pdf/4/1/No._03_-_Water_in_the_Atmosphere.pdf}}  \\ \hline
            PPGDalia & \makecell[l]{Low heart rate: $<75$ \\ Medium heart rate: $\geq75, <115$ \\ High heart rate: $\geq115$} & Hand selected by archive authors \\ \hline
            IEEEPPG & \makecell[l]{Low heart rate: $<100$ \\ Medium heart rate: $\geq100, <140$ \\ High heart rate: $\geq140$} & Hand selected by archive authors \\ \hline
            BIDMC32RR & \makecell[l]{Low RR: $<12$ \\ Medium RR: $\geq12, \leq20$ \\ High RR: $>20$} & The regular respiratory rate for a resting adult\tablefootnote{\url{https://web.archive.org/web/20240416235146/https://www.ncbi.nlm.nih.gov/books/NBK537306/}} \\ \hline
            BIDMC32HR & \makecell[l]{Low heart rate: $<80$ \\ Medium heart rate: $\geq80, <100$ \\ High heart rate: $\geq100$} & Hand selected by archive authors  \\ \hline
            BIDMC32SpO2 & \makecell[l]{Low SpO2: $<95$ \\ Healthy SpO2: $\geq95$} &  Normal/abnormal blood oxygen saturation threshold\tablefootnote{\url{https://web.archive.org/web/20240920173035/https://www.ncbi.nlm.nih.gov/books/NBK525974/}} \\ \hline
            NewsHeadlineSentiment & \makecell[l]{Negative sentiment: $<-0.1$ \\ Neutral sentiment: $\geq-0.1, \leq0.1$ \\ Positive sentiment: $>0.1$} & Basic binning of scores, similar to UK civil service blog post\tablefootnote{\url{https://web.archive.org/web/20240616033751/https://dataingovernment.blog.gov.uk/2016/02/18/analysing-survey-comments-using-sentiment-scores/}} \\ \hline
            NewsTitleSentiment & \makecell[l]{Negative sentiment: $<-0.1$ \\ Neutral sentiment: $\geq-0.1, \leq0.1$ \\ Positive sentiment: $>0.1$} & Basic binning of scores, similar to UK civil service blog post \\ \hline
        \end{tabular}
    \end{table}

\section{Further Results}
\label{app:results}

In this section, we provide results not present in Section~\ref{sec:results}. Tables~\ref{tab:big_results1} and ~\ref{tab:big_results2} show the accuracy scores for all datasets of equal length with no missing values.

    \begin{landscape}
    \begin{footnotesize}
    \begin{longtable}{lrrrrrrrr}
        \caption{Accuracy result for the top 8 classifiers from Section~\ref{sec:results} for all 133 clean datasets. Experiments which produced no results are left blank.}
        \label{tab:big_results1} \\

        \toprule
         & Arsenal & CIF & FP & HC2 & MRHydra & QUANT & RDST & ROCKET \\
        \midrule
        Alzheimers & 0.442 & 0.419 & & 0.349 & \textbf{0.488} & 0.395 & 0.419 & 0.419 \\
        AppliancesEnergy\_disc & 0.857 & 0.905 & \textbf{0.952} & 0.881 & 0.833 & 0.929 & 0.786 & 0.786 \\
        ArticularyWordRecognition & \textbf{0.993} & 0.98 & 0.977 & \textbf{0.993} & \textbf{0.993} & 0.983 & \textbf{0.993} & \textbf{0.993} \\
        AsphaltObstaclesCoordinates\_eq & 0.854 & 0.824 & 0.839 & 0.857 & 0.859 & 0.818 & \textbf{0.867} & 0.854 \\
        AsphaltPavementTypeCoordinates\_eq & 0.86 & 0.917 & 0.942 & 0.889 & 0.912 & 0.933 & 0.896 & 0.864 \\
        AsphaltRegularityCoordinates\_eq & 0.936 & 0.977 & 0.991 & 0.971 & 0.988 & 0.988 & 0.983 & 0.939 \\
        AtrialFibrillation & 0.067 & \textbf{0.467} & 0.267 & 0.2 & 0.067 & 0.133 & 0.2 & 0.067 \\
        AustraliaRainfall\_disc & 0.764 & 0.769 & 0.775 & & & 0.776 & & \\
        AutomotiveRoadTrials & 0.766 & 0.792 & 0.481 & \textbf{0.844} & 0.805 & 0.805 & 0.792 & 0.74 \\
        BIDMC32HR\_disc & 0.934 & & & & 0.772 & 0.615 & 0.941 & 0.921 \\
        BIDMC32RR\_disc & 0.753 & & & & 0.766 & 0.678 & 0.786 & 0.784 \\
        BIDMC32SpO2\_disc & 0.713 & & & & 0.594 & 0.569 & 0.677 & \textbf{0.72} \\
        BasicMotions & \textbf{1.0} & \textbf{1.0} & \textbf{1.0} & \textbf{1.0} & \textbf{1.0} & \textbf{1.0} & \textbf{1.0} & \textbf{1.0} \\
        BeijingPM10Quality\_disc\_nmv & 0.778 & 0.824 & 0.842 & & 0.802 & \textbf{0.844} & 0.82 & 0.747 \\
        BeijingPM25Quality\_disc\_nmv & 0.815 & 0.867 & 0.883 & & 0.843 & \textbf{0.887} & 0.869 & 0.777 \\
        BenzeneConcentration\_disc\_nmv & 0.78 & 0.818 & 0.859 & 0.785 & 0.729 & 0.86 & 0.744 & 0.753 \\
        Blink & \textbf{1.0} & \textbf{1.0} & 0.998 & \textbf{1.0} & 0.978 & 0.993 & \textbf{1.0} & \textbf{1.0} \\
        BoneIntensitiesAgeGroup & 0.807 & 0.809 & 0.791 & \textbf{0.825} & 0.798 & 0.818 & 0.809 & 0.813 \\
        BoneProbAgeGroup & 0.674 & 0.649 & 0.674 & 0.683 & 0.676 & 0.631 & 0.688 & 0.688 \\
        ButtonPress & 0.983 & \textbf{1.0} & \textbf{1.0} & \textbf{1.0} & \textbf{1.0} & 0.992 & 0.883 & 0.983 \\
        CharacterTrajectories\_eq & 0.992 & 0.985 & 0.991 & 0.992 & 0.994 & 0.985 & 0.994 & 0.993 \\
        CounterMovementJump\_eq & \textbf{0.927} & 0.76 & 0.821 & \textbf{0.927} & \textbf{0.927} & 0.894 & 0.883 & 0.922 \\
        Cricket & \textbf{1.0} & 0.986 & 0.972 & \textbf{1.0} & 0.986 & 0.986 & \textbf{1.0} & \textbf{1.0} \\
        CrowdSourced & 0.717 & 0.763 & 0.665 & & 0.756 & 0.688 & 0.747 & 0.709 \\
        DREAMERA & 0.418 & & & & & \textbf{0.445} & & \\
        DREAMERV & 0.629 & & & & & 0.63 & & \\
        DuckDuckGeese & 0.46 & 0.44 & 0.54 & 0.34 & 0.54 & 0.48 & 0.42 & 0.46 \\
        ERing & \textbf{0.989} & 0.97 & 0.926 & \textbf{0.989} & 0.974 & 0.978 & 0.974 & 0.981 \\
        EigenWorms & 0.901 & & & & \textbf{0.985} & 0.893 & 0.908 & 0.916 \\
        EmoPain & 0.811 & 0.845 & 0.851 & 0.839 & 0.87 & 0.845 & 0.89 & 0.803 \\
        Epilepsy & 0.986 & 0.986 & 0.942 & \textbf{1.0} & \textbf{1.0} & \textbf{1.0} & 0.993 & 0.986 \\
        EthanolConcentration & 0.411 & 0.715 & 0.38 & 0.719 & 0.601 & 0.696 & 0.612 & 0.433 \\
        EyesOpenShut & 0.524 & 0.5 & 0.5 & 0.452 & 0.429 & \textbf{0.571} & 0.476 & 0.524 \\
        FaceDetection & 0.661 & 0.652 & & & 0.591 & 0.633 & 0.629 & 0.606 \\
        FeedbackButton &  &  & & &  & &  &  \\
        FeetHands & \textbf{0.698} & 0.683 & & & 0.647 & 0.677 & 0.626 & 0.668 \\
        FingerMovements & 0.54 & 0.55 & 0.53 & 0.54 & 0.56 & 0.5 & 0.54 & 0.55 \\
        FordChallenge & 0.84 & 0.911 & & & 0.856 & \textbf{0.934} & 0.797 & 0.798 \\
        HandMovementDirection & 0.5 & 0.473 & 0.459 & 0.419 & 0.324 & 0.527 & 0.338 & 0.514 \\
        Handwriting & 0.561 & 0.386 & 0.348 & 0.585 & 0.522 & 0.336 & 0.605 & 0.587 \\
        Heartbeat & 0.741 & 0.776 & 0.776 & 0.751 & 0.751 & 0.776 & 0.751 & 0.756 \\
        HouseholdPowerConsumption1\_disc\_nmv & 0.889 & 0.961 & \textbf{0.984} & 0.904 & 0.895 & 0.974 & 0.892 & 0.891 \\
        HouseholdPowerConsumption2\_disc\_nmv & 0.771 & 0.771 & 0.77 & 0.78 & \textbf{0.787} & 0.783 & 0.774 & 0.748 \\
        IEEEPPG\_disc & 0.606 & 0.437 & 0.293 & 0.599 & 0.606 & 0.268 & \textbf{0.715} & 0.631 \\
        IRDS-EFL & 0.615 & 0.654 & \textbf{0.885} & 0.654 & 0.769 & 0.731 & 0.769 & 0.577 \\
        IRDS-EFR & 0.667 & 0.667 & 0.667 & 0.667 & 0.667 & 0.667 & 0.667 & 0.667 \\
        IRDS-SAL & 0.518 & 0.375 & 0.393 & 0.554 & 0.321 & 0.339 & \textbf{0.768} & 0.607 \\
        IRDS-SAR & 0.273 & 0.394 & 0.303 & 0.152 & 0.697 & 0.515 & 0.364 & 0.273 \\
        IRDS-SFE & 0.667 & 0.667 & 0.8 & 0.667 & 0.733 & 0.667 & 0.667 & 0.667 \\
        IRDS-SFL & 0.81 & 0.81 & 0.81 & 0.819 & 0.767 & 0.81 & \textbf{0.871} & 0.793 \\
        IRDS-SFR & 0.8 & 0.6 & 0.4 & 0.733 & 0.6 & 0.533 & \textbf{0.867} & 0.8 \\
        IRDS-STL & 0.737 & 0.763 & \textbf{0.868} & 0.737 & 0.684 & 0.5 & 0.763 & 0.737 \\
        IRDS-STR & 0.917 & 0.95 & \textbf{0.967} & 0.933 & 0.933 & 0.933 & 0.933 & 0.917 \\
        ImaginedFeetHands & 0.685 & 0.687 & & & 0.636 & 0.678 & 0.628 & 0.663 \\
        ImaginedOpenCloseFist & \textbf{0.763} & 0.758 & & & 0.702 & 0.743 & 0.68 & 0.739 \\
        InnerSpeech & 0.255 & & & 0.27 & 0.255 & 0.25 & \textbf{0.31} & 0.225 \\
        InsectWingbeat\_eq & 0.536 & 0.713 & & & 0.656 & 0.727 & 0.658 & 0.382 \\
        JapaneseVowels\_eq & 0.838 & 0.968 & 0.959 & 0.932 & 0.978 & 0.957 & 0.973 & 0.824 \\
        KERAAL-CTK & 0.857 & 0.857 & \textbf{1.0} & 0.929 & 0.929 & 0.857 & \textbf{1.0} & 0.857 \\
        KERAAL-CTK-MC & 0.857 & \textbf{0.929} & 0.857 & \textbf{0.929} & 0.857 & \textbf{0.929} & \textbf{0.929} & 0.714 \\
        KERAAL-ELK & \textbf{0.8} & \textbf{0.8} & \textbf{0.8} & \textbf{0.8} & \textbf{0.8} & \textbf{0.8} & 0.7 & \textbf{0.8} \\
        KERAAL-ELK-MC & 0.929 & 0.857 & 0.929 & 0.929 & \textbf{1.0} & 0.714 & \textbf{1.0} & 0.857 \\
        KERAAL-RTK & 0.857 & 0.786 & 0.786 & \textbf{0.929} & 0.857 & 0.714 & 0.571 & \textbf{0.929} \\
        KERAAL-RTK-MC & 0.462 & 0.385 & 0.308 & 0.308 & 0.308 & 0.385 & \textbf{0.538} & \textbf{0.538} \\
        KIMORE-LA-C & 0.429 & 0.429 & 0.429 & 0.429 & \textbf{0.571} & 0.429 & 0.429 & 0.429 \\
        KIMORE-LT-C & 0.429 & 0.143 & 0.571 & 0.286 & \textbf{0.714} & 0.571 & \textbf{0.714} & 0.429 \\
        KIMORE-PR-C & 0.571 & 0.429 & 0.286 & 0.143 & \textbf{0.714} & 0.571 & 0.571 & 0.571 \\
        KIMORE-Sq-C & 0.571 & 0.286 & 0.429 & 0.286 & 0.286 & 0.286 & 0.429 & \textbf{0.714} \\
        KIMORE-TR-C & 0.143 & 0.143 & 0.143 & 0.143 & 0.143 & 0.143 & 0.143 & \textbf{0.286} \\
        KINECAL-3WFV & 0.917 & 0.917 & 0.917 & 0.917 & 0.917 & 0.917 & 0.833 & 0.833 \\
        KINECAL-GGFV & 0.923 & 0.923 & 0.923 & 0.923 & 0.923 & 0.923 & 0.923 & 0.923 \\
        KINECAL-QSEC & \textbf{0.944} & \textbf{0.944} & \textbf{0.944} & \textbf{0.944} & \textbf{0.944} & \textbf{0.944} & \textbf{0.944} & \textbf{0.944} \\
        KINECAL-QSEO & \textbf{0.941} & \textbf{0.941} & 0.882 & \textbf{0.941} & \textbf{0.941} & \textbf{0.941} & \textbf{0.941} & \textbf{0.941} \\
        LSST & 0.658 & 0.637 & \textbf{0.703} & 0.674 & 0.643 & 0.645 & 0.649 & 0.636 \\
        LenDB & & & & & & \textbf{0.956} & & \\
        Libras & 0.9 & 0.906 & 0.889 & 0.928 & 0.933 & 0.878 & 0.944 & 0.906 \\
        LiveFuelMoistureContent\_disc & 0.83 & 0.838 & 0.834 & 0.829 & 0.761 & 0.82 & 0.815 & 0.788 \\
        Locust2022 & 0.917 & 0.908 & 0.909 & & 0.916 & 0.91 & 0.914 & \textbf{0.918} \\
        LongIntervalTask & 0.691 & & & & \textbf{0.693} & 0.692 & 0.595 & 0.69 \\
        LowCost & 0.607 & \textbf{0.727} & 0.647 & 0.678 & 0.585 & 0.653 & 0.607 & 0.63 \\
        MatchingPennies & 0.601 & & & & 0.594 & 0.551 & 0.533 & 0.614 \\
        MindReading & 0.462 & 0.59 & 0.594 & 0.606 & 0.59 & 0.603 & 0.492 & 0.476 \\
        MotionSenseHAR & 0.992 & 0.989 & 0.985 & \textbf{1.0} & 0.996 & 0.989 & 0.985 & 0.977 \\
        MotorImagery & 0.52 & 0.5 & & \textbf{0.55} & 0.51 & 0.51 & 0.48 & 0.54 \\
        NATOPS & 0.883 & 0.872 & 0.928 & 0.922 & 0.906 & 0.867 & 0.883 & 0.872 \\
        NewsHeadlineSentiment\_disc & 0.549 & 0.539 & 0.535 & & & 0.524 & & 0.513 \\
        NewsTitleSentiment\_disc & \textbf{0.625} & 0.615 & 0.612 & & & 0.607 & & 0.602 \\
        OpenCloseFist & \textbf{0.776} & 0.753 & & & 0.708 & 0.74 & 0.66 & 0.745 \\
        Opportunity & 0.828 & 0.847 & & & 0.819 & \textbf{0.856} & 0.843 & 0.794 \\
        PAMAP2 & 0.854 & & & & 0.862 & 0.863 & \textbf{0.9} & 0.839 \\
        PEMS-SF & 0.832 & \textbf{1.0} & 0.988 & \textbf{1.0} & 0.827 & 0.988 & 0.913 & 0.844 \\
        PPGDalia\_disc\_eq & \textbf{0.816} & & 0.753 & & 0.775 & 0.729 & & 0.78 \\
        PenDigits & 0.982 & 0.967 & 0.969 & 0.98 & & 0.975 & 0.985 & 0.981 \\
        PhonemeSpectra & 0.298 & 0.266 & 0.329 & & 0.33 & 0.281 & 0.313 & 0.278 \\
        PhotoStimulation & 0.528 & 0.361 & 0.417 & 0.444 & 0.389 & 0.361 & 0.333 & \textbf{0.583} \\
        PronouncedSpeech & 0.28 & & & 0.32 & 0.27 & 0.3 & 0.255 & 0.27 \\
        RacketSports & 0.901 & 0.882 & 0.868 & 0.901 & 0.888 & 0.901 & 0.888 & \textbf{0.908} \\
        S2Agri-10pc-17 & & & & & & \textbf{0.759} & & \\
        S2Agri-10pc-34 & & & & & & \textbf{0.748} & & \\
        S2Agri-17 & & & & & & & & \\
        S2Agri-34 & & & & & & & & \\
        SPHERE-WUS & \textbf{1.0} & 0.5 & 0.5 & \textbf{1.0} & \textbf{1.0} & 0.75 & 0.75 & \textbf{1.0} \\
        STEW & \textbf{0.836} & 0.796 & & & 0.728 & 0.696 & 0.768 & 0.784 \\
        SelfRegulationSCP1 & 0.853 & 0.867 & 0.915 & 0.874 & \textbf{0.949} & 0.812 & 0.891 & 0.853 \\
        SelfRegulationSCP2 & 0.544 & 0.511 & \textbf{0.561} & 0.556 & 0.533 & 0.511 & 0.522 & 0.528 \\
        ShortIntervalTask & 0.694 & & & & 0.642 & 0.68 & 0.632 & 0.686 \\
        SitStand & 0.396 & & & & 0.42 & \textbf{0.422} & 0.35 & 0.398 \\
        Skoda & 0.93 & 0.953 & & & 0.952 & 0.954 & 0.952 & 0.945 \\
        SongFamiliarity & 0.521 & & & 0.515 & 0.513 & 0.525 & 0.516 & 0.507 \\
        SpokenArabicDigits\_eq & 0.992 & 0.976 & 0.979 & 0.987 & 0.989 & 0.974 & \textbf{0.994} & 0.99 \\
        StandWalkJump & 0.533 & 0.467 & 0.467 & 0.333 & \textbf{0.667} & 0.4 & 0.533 & 0.467 \\
        TactileTextureRecognition & 0.999 & 0.999 & 0.997 & \textbf{1.0} & 0.999 & 0.997 & \textbf{1.0} & 0.999 \\
        Tiselac & 0.691 & 0.834 & & & & 0.831 & & 0.646 \\
        UCDHE-MP & 0.848 & 0.834 & 0.837 & 0.867 & 0.845 & 0.84 & \textbf{0.887} & 0.859 \\
        UCDHE-MP-MC & 0.757 & 0.718 & 0.746 & 0.751 & 0.798 & 0.671 & \textbf{0.809} & 0.785 \\
        UCDHE-Rowing & 0.82 & 0.827 & 0.825 & 0.823 & 0.832 & 0.834 & 0.848 & 0.816 \\
        UCDHE-Rowing-MC & 0.741 & \textbf{0.811} & 0.198 & 0.793 & 0.798 & 0.789 & 0.768 & 0.73 \\
        UCIActivity & 0.953 & 0.954 & 0.975 & & 0.992 & 0.969 & 0.99 & 0.933 \\
        UIPRMD-DS-C & 0.833 & 0.917 & 0.917 & \textbf{1.0} & 0.944 & 0.833 & 0.694 & 0.861 \\
        UIPRMD-HS-C & \textbf{1.0} & \textbf{1.0} & \textbf{1.0} & \textbf{1.0} & \textbf{1.0} & \textbf{1.0} & \textbf{1.0} & \textbf{1.0} \\
        UIPRMD-IL-C & 0.722 & 0.889 & 0.889 & 0.722 & 0.889 & 0.889 & 0.667 & 0.778 \\
        UIPRMD-SASLR-C & \textbf{1.0} & 0.95 & \textbf{1.0} & \textbf{1.0} & 0.6 & 0.95 & 0.95 & \textbf{1.0} \\
        UIPRMD-SL-C & \textbf{1.0} & 0.889 & 0.722 & \textbf{1.0} & 0.944 & 0.889 & \textbf{1.0} & \textbf{1.0} \\
        UIPRMD-SSA-C & \textbf{1.0} & 0.944 & \textbf{1.0} & \textbf{1.0} & \textbf{1.0} & 0.944 & 0.889 & \textbf{1.0} \\
        UIPRMD-SSE-C & 0.944 & 0.889 & \textbf{1.0} & 0.944 & 0.889 & \textbf{1.0} & 0.944 & 0.944 \\
        UIPRMD-SSIER-C & \textbf{0.9} & 0.8 & 0.7 & 0.85 & 0.85 & 0.8 & \textbf{0.9} & \textbf{0.9} \\
        UIPRMD-SSS-C & 0.944 & \textbf{1.0} & 0.833 & 0.944 & \textbf{1.0} & 0.944 & 0.944 & 0.944 \\
        UIPRMD-STS-C & \textbf{1.0} & 0.944 & 0.972 & 0.972 & 0.972 & 0.944 & 0.917 & \textbf{1.0} \\
        USCActivity & 0.718 & 0.689 & & & 0.68 & 0.735 & & 0.663 \\
        UWaveGestureLibrary & 0.938 & 0.922 & 0.853 & \textbf{0.941} & 0.931 & 0.894 & 0.938 & \textbf{0.941} \\
        VisualSpeech & 0.245 & & & 0.29 & 0.235 & \textbf{0.3} & 0.25 & 0.235 \\
        WISDM & \textbf{0.892} & 0.83 & 0.883 & & \textbf{0.892} & 0.878 & 0.874 & 0.882 \\
        WISDM2 & 0.618 & \textbf{0.688} & & & & 0.655 & & 0.582 \\
        \bottomrule
    \end{longtable}

    \begin{longtable}{lrrrrrrrrr}
        \caption{Accuracy result for the remaining classifiers from Section~\ref{sec:results} for all 133 clean datasets. Experiments which produced no results are left blank.}
        \label{tab:big_results2} \\

        \toprule
         & 1NN-DTW & Catch22 & DrCIF & Dummy & H-IT & LITETime & RIST & STC & TDE \\
        \midrule
        Alzheimers & 0.279 & \textbf{0.488} & 0.442 & 0.419 & 0.326 & 0.372 & 0.442 & 0.372 & 0.372 \\
        AppliancesEnergy\_disc & 0.643 & 0.881 & 0.905 & 0.81 & 0.524 & 0.738 & 0.905 & 0.857 & 0.81 \\
        ArticularyWordRecognition & 0.987 & 0.983 & 0.987 & 0.04 & 0.99 & 0.983 & 0.987 & 0.973 & \textbf{0.993} \\
        AsphaltObstaclesCoordinates\_eq & 0.578 & 0.808 & 0.821 & 0.284 & 0.854 & 0.847 & 0.821 & 0.821 & 0.818 \\
        AsphaltPavementTypeCoordinates\_eq & 0.583 & 0.923 & 0.934 & 0.386 & 0.963 & \textbf{0.964} & 0.937 & 0.855 & 0.839 \\
        AsphaltRegularityCoordinates\_eq & 0.703 & 0.976 & 0.987 & 0.507 & \textbf{0.995} & \textbf{0.995} & 0.987 & 0.951 & 0.947 \\
        AtrialFibrillation & 0.2 & 0.133 & 0.067 & 0.333 & 0.333 & 0.2 & 0.2 & 0.2 & 0.333 \\
        AustraliaRainfall\_disc & 0.693 & 0.763 & 0.774 & 0.686 & 0.779 & \textbf{0.782} & 0.772 & & \\
        AutomotiveRoadTrials & 0.662 & 0.818 & 0.831 & 0.753 & & & 0.74 & \textbf{0.844} & 0.805 \\
        BIDMC32HR\_disc & & 0.789 & & 0.651 & 0.871 & 0.753 & \textbf{0.962} & 0.915 & \\
        BIDMC32RR\_disc & & 0.787 & & \textbf{0.794} & 0.627 & 0.702 & \textbf{0.794} & & \\
        BIDMC32SpO2\_disc & & 0.648 & & 0.715 & 0.712 & 0.682 & 0.715 & & \\
        BasicMotions & 0.975 & \textbf{1.0} & \textbf{1.0} & 0.25 & \textbf{1.0} & \textbf{1.0} & \textbf{1.0} & 0.95 & \textbf{1.0} \\
        BeijingPM10Quality\_disc\_nmv & 0.741 & 0.83 & 0.833 & 0.711 & 0.822 & 0.818 & 0.837 & 0.76 & \\
        BeijingPM25Quality\_disc\_nmv & 0.814 & 0.875 & 0.876 & 0.698 & 0.877 & 0.885 & 0.881 & 0.799 & \\
        BenzeneConcentration\_disc\_nmv & 0.821 & 0.517 & 0.826 & 0.69 & \textbf{0.964} & 0.892 & 0.817 & 0.808 & 0.69 \\
        Blink & 0.584 & 0.998 & 0.982 & 0.444 & 0.667 & 0.687 & 0.993 & 0.998 & 0.998 \\
        BoneIntensitiesAgeGroup & 0.524 & 0.733 & 0.809 & 0.476 & 0.775 & 0.784 & 0.818 & 0.813 & 0.804 \\
        BoneProbAgeGroup & 0.519 & 0.64 & 0.643 & 0.476 & 0.708 & \textbf{0.71} & 0.634 & 0.629 & 0.688 \\
        ButtonPress & 0.567 & \textbf{1.0} & \textbf{1.0} & 0.442 & 0.917 & 0.992 & \textbf{1.0} & 0.9 & \textbf{1.0} \\
        CharacterTrajectories\_eq & 0.988 & 0.98 & 0.985 & 0.065 & \textbf{0.996} & \textbf{0.996} & 0.981 & 0.975 & 0.98 \\
        CounterMovementJump\_eq & 0.575 & 0.743 & 0.838 & 0.335 & 0.654 & 0.709 & 0.922 & 0.726 & 0.816 \\
        Cricket & \textbf{1.0} & 0.986 & 0.958 & 0.083 & 0.986 & 0.986 & 0.986 & 0.972 & 0.972 \\
        CrowdSourced & 0.61 & 0.745 & 0.73 & 0.5 & 0.745 & \textbf{0.771} & 0.711 & 0.684 & \\
        DREAMERA & & 0.442 & & 0.393 & & & & & \\
        DREAMERV & & \textbf{0.632} & & 0.581 & & & & & \\
        DuckDuckGeese & 0.58 & 0.5 & 0.52 & 0.2 & \textbf{0.64} & 0.46 & 0.48 & 0.42 & 0.24 \\
        ERing & 0.915 & 0.922 & 0.985 & 0.167 & 0.933 & 0.933 & 0.978 & 0.893 & 0.956 \\
        EigenWorms & 0.618 & 0.931 & & 0.42 & 0.74 & 0.649 & 0.954 & 0.855 & 0.847 \\
        EmoPain & \textbf{0.927} & 0.885 & 0.848 & 0.783 & 0.676 & 0.842 & 0.879 & 0.831 & 0.811 \\
        Epilepsy & 0.964 & 0.986 & 0.993 & 0.268 & 0.971 & \textbf{1.0} & 0.986 & 0.986 & \textbf{1.0} \\
        EthanolConcentration & 0.323 & 0.357 & 0.673 & 0.251 & 0.24 & 0.27 & 0.62 & \textbf{0.749} & 0.521 \\
        EyesOpenShut & 0.476 & 0.5 & 0.5 & 0.5 & 0.429 & 0.429 & 0.524 & 0.476 & 0.429 \\
        FaceDetection & 0.529 & 0.556 & 0.619 & 0.5 & \textbf{0.689} & 0.647 & 0.6 & 0.629 & \\
        FeedbackButton &  &  & & &  & &  &  \\
        FeetHands & & 0.577 & 0.681 & 0.504 & 0.605 & 0.614 & 0.678 & 0.668 & \\
        FingerMovements & 0.53 & 0.5 & 0.48 & 0.49 & 0.57 & \textbf{0.64} & 0.52 & 0.58 & 0.5 \\
        FordChallenge & 0.655 & 0.911 & 0.916 & 0.623 & 0.81 & 0.877 & 0.907 & 0.623 & \\
        HandMovementDirection & 0.189 & 0.257 & 0.527 & 0.203 & 0.432 & 0.338 & 0.446 & \textbf{0.541} & 0.365 \\
        Handwriting & 0.607 & 0.256 & 0.353 & 0.038 & \textbf{0.647} & 0.631 & 0.386 & 0.412 & 0.561 \\
        Heartbeat & 0.717 & 0.761 & \textbf{0.785} & 0.722 & 0.756 & 0.712 & 0.776 & 0.722 & 0.654 \\
        HouseholdPowerConsumption1\_disc\_nmv & 0.764 & 0.975 & 0.942 & 0.778 & 0.921 & 0.911 & 0.934 & 0.876 & 0.856 \\
        HouseholdPowerConsumption2\_disc\_nmv & 0.646 & 0.732 & 0.777 & 0.722 & 0.783 & 0.736 & 0.767 & 0.773 & 0.749 \\
        IEEEPPG\_disc & 0.388 & 0.489 & 0.309 & 0.261 & 0.666 & 0.379 & 0.324 & 0.658 & 0.35 \\
        IRDS-EFL & 0.654 & 0.731 & 0.538 & 0.538 & 0.654 & 0.654 & 0.538 & 0.846 & 0.577 \\
        IRDS-EFR & 0.8 & 0.667 & 0.667 & 0.667 & 0.667 & \textbf{0.867} & 0.667 & 0.667 & 0.667 \\
        IRDS-SAL & 0.321 & 0.393 & 0.375 & 0.75 & 0.268 & 0.357 & 0.321 & 0.661 & 0.714 \\
        IRDS-SAR & 0.121 & 0.212 & 0.182 & 0.121 & 0.727 & \textbf{0.758} & 0.121 & 0.455 & 0.121 \\
        IRDS-SFE & \textbf{1.0} & 0.733 & 0.667 & 0.667 & 0.667 & 0.867 & 0.667 & 0.667 & 0.667 \\
        IRDS-SFL & 0.491 & 0.784 & 0.819 & 0.207 & 0.569 & 0.647 & 0.845 & 0.853 & 0.5 \\
        IRDS-SFR & \textbf{0.867} & 0.4 & 0.533 & \textbf{0.867} & 0.733 & 0.533 & 0.6 & 0.6 & \textbf{0.867} \\
        IRDS-STL & 0.526 & 0.737 & 0.526 & 0.711 & 0.158 & 0.553 & 0.711 & 0.763 & 0.711 \\
        IRDS-STR & 0.183 & 0.45 & 0.867 & 0.917 & 0.883 & 0.683 & 0.9 & \textbf{0.967} & 0.883 \\
        ImaginedFeetHands & & 0.564 & 0.688 & 0.5 & 0.582 & 0.619 & 0.681 & \textbf{0.689} & \\
        ImaginedOpenCloseFist & & 0.543 & 0.754 & 0.499 & 0.728 & 0.698 & 0.716 & \textbf{0.763} & \\
        InnerSpeech & & 0.22 & 0.24 & 0.235 & 0.235 & 0.28 & 0.24 & 0.23 & 0.27 \\
        InsectWingbeat\_eq & & 0.708 & & 0.1 & 0.715 & \textbf{0.733} & 0.7 & 0.363 & \\
        JapaneseVowels\_eq & 0.959 & 0.949 & 0.973 & 0.084 & 0.989 & \textbf{0.995} & 0.954 & 0.789 & 0.786 \\
        KERAAL-CTK & 0.929 & 0.857 & 0.857 & 0.5 & \textbf{1.0} & 0.857 & 0.857 & 0.857 & \textbf{1.0} \\
        KERAAL-CTK-MC & 0.857 & \textbf{0.929} & \textbf{0.929} & 0.5 & 0.857 & 0.857 & 0.786 & 0.5 & \textbf{0.929} \\
        KERAAL-ELK & 0.5 & \textbf{0.8} & \textbf{0.8} & \textbf{0.8} & & & \textbf{0.8} & \textbf{0.8} & \textbf{0.8} \\
        KERAAL-ELK-MC & 0.929 & 0.929 & 0.714 & 0.929 & 0.857 & 0.643 & 0.786 & 0.929 & 0.857 \\
        KERAAL-RTK & 0.643 & 0.786 & 0.786 & 0.571 & 0.714 & 0.857 & 0.714 & 0.714 & 0.571 \\
        KERAAL-RTK-MC & 0.308 & \textbf{0.538} & 0.385 & 0.385 & 0.308 & 0.385 & 0.308 & 0.385 & 0.385 \\
        KIMORE-LA-C & 0.429 & 0.429 & 0.429 & 0.429 & & & 0.429 & 0.429 & 0.429 \\
        KIMORE-LT-C & 0.429 & 0.429 & 0.286 & 0.429 & & & & 0.286 & 0.429 \\
        KIMORE-PR-C & 0.286 & 0.429 & 0.429 & 0.143 & & & 0.429 & 0.143 & 0.286 \\
        KIMORE-Sq-C & 0.571 & 0.286 & 0.286 & 0.286 & & & 0.286 & 0.286 & 0.286 \\
        KIMORE-TR-C & 0.143 & 0.143 & 0.143 & 0.143 & & & 0.143 & 0.143 & 0.143 \\
        KINECAL-3WFV & 0.917 & 0.917 & 0.917 & 0.917 & 0.833 & \textbf{1.0} & 0.917 & 0.833 & 0.917 \\
        KINECAL-GGFV & \textbf{1.0} & 0.923 & 0.923 & 0.923 & \textbf{1.0} & 0.923 & 0.923 & 0.923 & 0.923 \\
        KINECAL-QSEC & 0.889 & \textbf{0.944} & \textbf{0.944} & \textbf{0.944} & 0.889 & 0.889 & \textbf{0.944} & \textbf{0.944} & \textbf{0.944} \\
        KINECAL-QSEO & 0.882 & \textbf{0.941} & \textbf{0.941} & \textbf{0.941} & 0.824 & 0.882 & \textbf{0.941} & \textbf{0.941} & \textbf{0.941} \\
        LSST & 0.551 & 0.67 & 0.635 & 0.315 & 0.207 & 0.359 & 0.627 & 0.613 & 0.558 \\
        LenDB & & 0.892 & & 0.509 & & & & & \\
        Libras & 0.872 & 0.817 & 0.911 & 0.067 & 0.889 & 0.917 & \textbf{0.978} & 0.906 & 0.872 \\
        LiveFuelMoistureContent\_disc & 0.728 & 0.807 & 0.837 & \textbf{0.841} & 0.782 & 0.743 & 0.825 & 0.797 & 0.828 \\
        Locust2022 & 0.875 & 0.914 & 0.905 & 0.911 & 0.909 & 0.904 & 0.912 & 0.905 & \\
        LongIntervalTask & & 0.656 & & 0.679 & 0.678 & 0.675 & & 0.667 & \\
        LowCost & 0.538 & 0.552 & \textbf{0.727} & 0.5 & 0.575 & 0.573 & 0.65 & 0.662 & 0.648 \\
        MatchingPennies & & 0.529 & & 0.497 & 0.529 & 0.534 & & \textbf{0.626} & 0.554 \\
        MindReading & 0.41 & 0.548 & 0.567 & 0.231 & 0.466 & \textbf{0.72} & 0.557 & 0.459 & 0.527 \\
        MotionSenseHAR & 0.947 & 0.989 & \textbf{1.0} & 0.204 & 0.966 & 0.981 & \textbf{1.0} & 0.977 & \textbf{1.0} \\
        MotorImagery & 0.5 & 0.53 & 0.5 & 0.5 & 0.54 & 0.5 & 0.54 & 0.51 & 0.52 \\
        NATOPS & 0.883 & 0.872 & 0.861 & 0.167 & 0.961 & \textbf{0.967} & 0.861 & 0.872 & 0.844 \\
        NewsHeadlineSentiment\_disc & & 0.525 & 0.536 & \textbf{0.551} & 0.546 & 0.542 & 0.535 & & \\
        NewsTitleSentiment\_disc & & 0.608 & 0.613 & \textbf{0.625} & 0.572 & 0.61 & 0.609 & & \\
        OpenCloseFist & & 0.556 & 0.742 & 0.501 & 0.741 & 0.698 & 0.708 & 0.755 & \\
        Opportunity & & 0.836 & 0.851 & 0.483 & 0.835 & 0.844 & 0.843 & 0.777 & \\
        PAMAP2 & & 0.848 & & 0.109 & 0.819 & 0.834 & 0.879 & 0.7 & \\
        PEMS-SF & 0.711 & 0.983 & \textbf{1.0} & 0.116 & 0.798 & 0.728 & \textbf{1.0} & 0.988 & \textbf{1.0} \\
        PPGDalia\_disc\_eq & & 0.771 & & 0.544 & 0.811 & 0.791 & 0.785 & \textbf{0.816} & \\
        PenDigits & 0.977 & 0.949 & 0.973 & 0.104 & \textbf{0.989} & 0.988 & 0.979 & 0.97 & 0.884 \\
        PhonemeSpectra & 0.151 & 0.246 & 0.299 & 0.026 & \textbf{0.37} & 0.311 & 0.3 & 0.29 & \\
        PhotoStimulation & 0.25 & 0.333 & 0.417 & 0.417 & 0.25 & 0.194 & 0.444 & 0.389 & 0.417 \\
        PronouncedSpeech & & 0.29 & \textbf{0.325} & 0.23 & 0.245 & 0.25 & 0.31 & 0.32 & 0.3 \\
        RacketSports & 0.803 & 0.822 & 0.888 & 0.283 & 0.882 & 0.882 & 0.875 & 0.855 & 0.836 \\
        S2Agri-10pc-17 & & & & 0.609 & & & & & \\
        S2Agri-10pc-34 & & & & 0.609 & & & & & \\
        S2Agri-17 & & & & \textbf{0.596} & & & & & \\
        S2Agri-34 & & & & \textbf{0.596} & & & & & \\
        SPHERE-WUS & 0.75 & 0.75 & 0.75 & 0.5 & & & \textbf{1.0} & 0.75 & 0.75 \\
        STEW & 0.651 & 0.791 & 0.704 & 0.5 & 0.79 & 0.649 & 0.714 & 0.788 & \\
        SelfRegulationSCP1 & 0.775 & 0.771 & 0.884 & 0.502 & 0.819 & 0.805 & 0.85 & 0.833 & 0.833 \\
        SelfRegulationSCP2 & 0.539 & 0.55 & 0.55 & 0.5 & 0.506 & 0.511 & 0.528 & 0.489 & 0.539 \\
        ShortIntervalTask & & 0.644 & & 0.676 & 0.675 & \textbf{0.729} & & 0.665 & \\
        SitStand & & 0.364 & & 0.25 & 0.371 & 0.301 & 0.412 & 0.355 & \\
        Skoda & 0.954 & 0.939 & 0.945 & 0.236 & \textbf{0.969} & 0.963 & 0.942 & 0.845 & \\
        SongFamiliarity & & 0.507 & \textbf{0.537} & 0.519 & 0.513 & 0.525 & 0.528 & 0.52 & 0.48 \\
        SpokenArabicDigits\_eq & 0.972 & 0.963 & 0.98 & 0.1 & 0.992 & 0.993 & 0.977 & 0.955 & 0.925 \\
        StandWalkJump & 0.2 & 0.467 & 0.4 & 0.333 & 0.333 & 0.467 & 0.467 & 0.333 & 0.4 \\
        TactileTextureRecognition & 0.984 & 0.997 & 0.999 & 0.051 & & & \textbf{1.0} & 0.996 & 0.999 \\
        Tiselac & 0.766 & 0.825 & \textbf{0.836} & 0.063 & 0.812 & 0.805 & 0.832 & & \\
        UCDHE-MP & 0.5 & 0.785 & 0.812 & 0.724 & 0.818 & 0.876 & 0.837 & 0.865 & 0.815 \\
        UCDHE-MP-MC & 0.381 & 0.613 & 0.696 & 0.254 & & & 0.751 & 0.691 & 0.638 \\
        UCDHE-Rowing & 0.664 & 0.809 & \textbf{0.85} & 0.795 & 0.791 & 0.811 & 0.805 & 0.811 & 0.782 \\
        UCDHE-Rowing-MC & 0.491 & 0.73 & 0.755 & 0.205 & & & 0.702 & 0.745 & 0.57 \\
        UCIActivity & 0.975 & 0.951 & 0.959 & 0.192 & \textbf{0.998} & \textbf{0.998} & 0.986 & 0.823 & \\
        UIPRMD-DS-C & 0.75 & \textbf{1.0} & 0.833 & 0.5 & 0.667 & 0.861 & 0.833 & 0.944 & 0.889 \\
        UIPRMD-HS-C & 0.778 & \textbf{1.0} & \textbf{1.0} & 0.5 & \textbf{1.0} & \textbf{1.0} & \textbf{1.0} & \textbf{1.0} & \textbf{1.0} \\
        UIPRMD-IL-C & 0.5 & \textbf{0.944} & 0.5 & 0.5 & 0.5 & 0.5 & 0.444 & 0.722 & 0.611 \\
        UIPRMD-SASLR-C & \textbf{1.0} & 0.85 & 0.7 & 0.5 & 0.95 & 0.9 & \textbf{1.0} & \textbf{1.0} & 0.95 \\
        UIPRMD-SL-C & 0.778 & 0.833 & 0.833 & 0.5 & 0.5 & 0.5 & 0.944 & \textbf{1.0} & 0.778 \\
        UIPRMD-SSA-C & \textbf{1.0} & 0.944 & 0.944 & 0.5 & \textbf{1.0} & \textbf{1.0} & 0.944 & 0.944 & \textbf{1.0} \\
        UIPRMD-SSE-C & 0.667 & 0.944 & 0.833 & 0.5 & 0.667 & 0.778 & 0.833 & 0.944 & 0.833 \\
        UIPRMD-SSIER-C & 0.65 & 0.75 & 0.85 & 0.5 & 0.45 & 0.5 & 0.85 & 0.8 & \textbf{0.9} \\
        UIPRMD-SSS-C & 0.889 & \textbf{1.0} & \textbf{1.0} & 0.5 & \textbf{1.0} & \textbf{1.0} & \textbf{1.0} & 0.833 & 0.833 \\
        UIPRMD-STS-C & 0.806 & 0.944 & 0.944 & 0.5 & 0.694 & 0.694 & 0.944 & 0.972 & 0.917 \\
        USCActivity & 0.619 & 0.728 & 0.661 & 0.114 & 0.73 & \textbf{0.736} & 0.721 & 0.593 & \\
        UWaveGestureLibrary & 0.903 & 0.872 & 0.916 & 0.125 & 0.909 & 0.916 & 0.928 & 0.894 & 0.919 \\
        VisualSpeech & & 0.23 & 0.25 & 0.26 & 0.25 & 0.255 & 0.225 & 0.29 & 0.295 \\
        WISDM & 0.831 & 0.842 & 0.87 & 0.366 & 0.866 & 0.854 & 0.883 & 0.867 & \\
        WISDM2 & & 0.671 & 0.679 & 0.35 & 0.54 & 0.525 & 0.674 & & \\
        \bottomrule
    \end{longtable}
    \end{footnotesize}
    \end{landscape}

\end{document}